%% file: main.tex
\documentclass[runningheads]{llncs}

\usepackage{etoolbox}

\newtoggle{eccvfinal}

\usepackage{eccv}
\global \toggletrue{eccvfinal}

\usepackage{media9}

\usepackage{eccvabbrv}

\usepackage{graphicx}
\usepackage{booktabs}

\usepackage[accsupp]{axessibility}  %

\usepackage[colorlinks,citecolor=eccvblue]{hyperref}

\usepackage{orcidlink}

\input{preamble}

\usepackage{tikz}
\usepackage{algpseudocode}
\usepackage[font=small,labelfont=bf]{caption}
\usepackage{array}
\usepackage{multirow}
\usepackage{subcaption}
\usepackage[normalem]{ulem}
\usepackage{xparse}
\usepackage{pifont}
\usepackage[export]{adjustbox}
\usepackage{arydshln}

\usepackage[scaled=0.85]{DejaVuSansMono}
\definecolor{magenta}{rgb}{0.73, 0.31, 0.56}
\definecolor{mustard}{rgb}{0.99, 0.65, 0.0}
\definecolor{pastelblue}{rgb}{0.296875, 0.66796875, 0.8984375}
\definecolor{pastelorange}{rgb}{0.99609375, 0.26953125, 0.0}
\definecolor{pastelpurple}{rgb}{0.5, 0.1484375, 0.8671875}
\definecolor{pastelgreen}{rgb}{0.0234375, 0.57421875, 0.40234375}
\definecolor{pastelred}{rgb}{0.50390625, 0.046875, 0.0}
\usepackage{url}
\usepackage{bm}

\usepackage{makecell}
\usepackage{multirow}
\usepackage{multicol}
\usepackage{subcaption}
\usepackage{colortbl}
\usepackage{color}
\usepackage{calc}

\usepackage{titletoc}

\usepackage[capitalize]{cleveref}
\crefname{section}{\S}{\S\S}
\crefname{subsection}{\S}{\S\S}
\crefformat{table}{Table~#2#1#3}
\crefformat{figure}{Figure~#2#1#3}
\crefformat{equation}{Eq~#2#1#3}

\newlength\savewidth

\newlength\thinwidth

\definecolor{Gray}{gray}{0.92}
\definecolor{DarkGray}{gray}{0.5}
\newcolumntype{x}{>{\columncolor{Gray}}c}
\newcolumntype{H}{>{\setbox0=\hbox\bgroup}c<{\egroup}@{}}
\definecolor{LightCyan}{rgb}{0.88,1,1}
\definecolor{altRowColor}{gray}{0.92}
\definecolor{highlightRowColor}{rgb}{0.9, 0.9, 1}
\definecolor{snsBlueColor}{rgb}{0.2980392156862745, 0.4470588235294118, 0.6901960784313725}
\definecolor{snsGreenColor}{rgb}{0.3333333333333333, 0.6588235294117647, 0.40784313725490196}
\definecolor{RedColor}{rgb}{1.0, 0., 0.}
\definecolor{BlueColor}{rgb}{0.122, 0.467, 0.706}
\definecolor{OrangeColor}{rgb}{1.0, 0.5, 0.25}
\definecolor{snsTealColor}{rgb}{0.39215686274509803, 0.7098039215686275, 0.803921568627451}
\definecolor{snsVioletColor}{rgb}{0.5058823529411764, 0.4470588235294118, 0.7019607843137254}

\definecolor{GrayNumber}{gray}{0.5}
\definecolor{GrayXMark}{gray}{0.7}

\newcommand{\OURS}{\textsc{Emu Video}\xspace}

\newcommand{\Ours}{\OURS}
\newcommand{\TextToV}{Text-to-Video\xspace}
\newcommand{\textToV}{text-to-video\xspace}
\newcommand{\imageToV}{image-to-video\xspace}
\newcommand{\imageToVShort}{I2V\xspace}
\newcommand{\ImageToV}{Image-to-Video\xspace}

\newcommand{\textToVShort}{T2V\xspace}
\newcommand{\TextToI}{Text-to-Image\xspace}
\newcommand{\textToI}{text-to-image\xspace}
\newcommand{\textToIShort}{T2I\xspace}

\newcommand{\juice}{JUICE\xspace}

\newcommand{\unet}{U-Net\xspace}

\newcommand{\imagenvideo}{Imagen Video\xspace}
\newcommand{\mav}{Make-A-Video\xspace}
\newcommand{\ayol}{Align Your Latents\xspace}
\newcommand{\pyoco}{PYOCO\xspace}
\newcommand{\reusediffuse}{Reuse \& Diffuse\xspace}
\newcommand{\gen}{Gen2\xspace}
\newcommand{\sdvideo}{Stable Video Diffusion\xspace}
\newcommand{\itovgen}{I2VGen-XL\xspace}
\newcommand{\videocrafter}{VideoCrafter\xspace}
\newcommand{\cogvideo}{CogVideo\xspace}

\newcommand{\pika}{PikaLabs\xspace}
\newcommand{\vidcomp}{VideoComposer\xspace}

\newcommand{\mavdatasetshort}{MAV\xspace}
\newcommand{\ayldatasetshort}{AYL\xspace}

\newcommand{\ucf}{UCF101\xspace}

\newcommand{\quality}{Quality\xspace}
\newcommand{\faithfulness}{Faithfulness\xspace}
\newcommand{\qualityShort}{Q\xspace}
\newcommand{\faithfulnessShort}{F\xspace}
\newcommand{\motionShort}{M\xspace}

\newcommand{\beps}{\bm{\epsilon}}

\newcommand{\bx}{\mathbf{X}}
\newcommand{\finalDiffusionTime}{N}

\newcommand{\bv}{\mathbf{V}}
\newcommand{\bi}{\mathbf{I}}
\newcommand{\bc}{\mathbf{c}}
\newcommand{\bp}{\mathbf{p}}
\newcommand{\bmask}{\mathbf{m}}
\newcommand{\vTime}{T}
\newcommand{\iTime}{T_{p}}
\newcommand{\channels}{C}
\newcommand{\zHeight}{H}
\newcommand{\zWidth}{W}
\newcommand{\extrapolateModel}{\mathcal{F}}
\newcommand{\interpolateModel}{\mathcal{I}}

\newcommand{\supplementarysection}{%
  \setcounter{figure}{0}%
  \let\oldthefigure\thefigure%
  \setcounter{section}{0}
  \let\oldthesection\thesection%
  \setcounter{table}{0}
  \let\oldthetable\thetable%

}
\newcommand{\supplementarytitle}{
  \title{
    Appendix
}

\author{}
\institute{}

\authorrunning{R. Girdhar et al.}
\maketitle
\supplementarysection
}
\definecolor{GraphColor}{rgb}{0.11764705882352941, 0.5647058823529412, 1.0}
\definecolor{GraphColorT2V}{rgb}{.5,.0,.5}
\definecolor{magenta}{rgb}{0.73, 0.31, 0.56}
\definecolor{mustard}{rgb}{0.99, 0.65, 0.0}

\title{
  \OURS: Factorizing Text-to-Video Generation by Explicit Image Conditioning
}
\titlerunning{Factorizing Text-to-Video Generation by Explicit Image Conditioning}

\author{
  Rohit Girdhar$^{\dagger,*}$,
  Mannat Singh$^{\dagger,*}$,
  Andrew Brown$^*$,
  Quentin Duval$^*$,\\
  Samaneh Azadi$^*$,
  Sai Saketh Rambhatla,
  Akbar Shah,
  Xi Yin,
  Devi Parikh,
  Ishan Misra$^*$ \\
 {\small \texttt{\url{https://emu-video.metademolab.com/}}}
}
\institute{GenAI, Meta}

\authorrunning{R. Girdhar et al.}

\usepackage{pgfplots}
\pgfplotsset{compat=1.16}
\usepgfplotslibrary{groupplots}
\usepgfplotslibrary{fillbetween}
\usetikzlibrary{patterns}
\usepackage{xr-hyper}
\usepackage{hyperref}

\providecommand{\etal}[0]{\emph{et al.}}

\begin{document}
\input{figures/teaser}
\makeatletter{\renewcommand*{\@makefnmark}{}
\footnotetext{$^{\dagger}$Equal first authors $^*$Equal technical contribution}\makeatother}
\input{sections/abstract}
\input{sections/intro}

\input{sections/related}

\input{sections/approach}
\input{sections/ablations}

\input{sections/comparison_vs_prior}

\input{sections/applications}

\input{sections/conclusion}

{
    \small
	\bibliographystyle{splncs04}
    \bibliography{refs}
}
\newpage
\supplementarytitle

\input{sections/supplement}

\end{document}

%% file: preamble.tex
\usepackage[dvipsnames]{xcolor}

%% file: figures/teaser.tex
\maketitle
\begin{center}
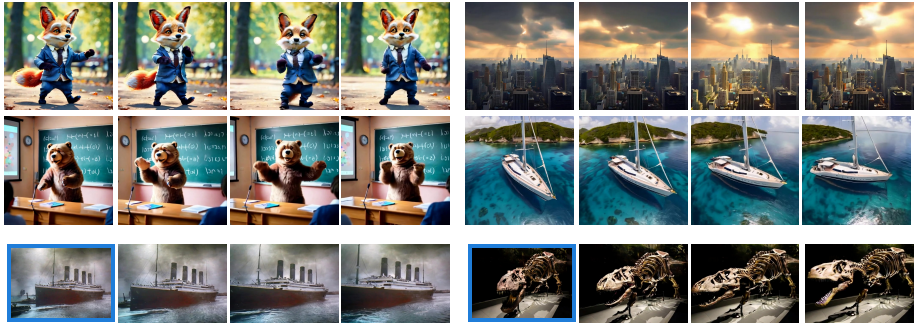

    \centering
    \captionsetup{type=figure}
    \input{figures/teaser/figure.tex}
    \captionof{figure}{
    {\bf \OURS} can generate high quality and temporally consistent videos while using a text prompt as input (top two rows), or an additional user-provided image (bottom row).
    Prompts: (top-left) A fox dressed in a suit dancing in a park,
    (top-right) The sun breaks through the clouds from the heights of a skyscraper,
    (middle-left): A bear is giving a presentation in the classroom,
    (middle-right): A 360 shot of a sleek yacht sailing gracefully through the crystal-clear waters of the Caribbean,
    (bottom-left): A ship driving off the harbor,
    (bottom-right): The dinosaur slowly comes to life.
    In the bottom two examples, a user-image is provided as an additional conditioning (shown in a \textcolor{eccvblue}{blue} border) and brought to life by \OURS . The first one is a historical picture of the RMS Titanic departing from Belfast, Northern Ireland; and the second is a picture of a Tyrannosaurus rex fossil.
    Please see the website linked above for videos.
    }
\label{fig:teaser}
\end{center}%

%% file: figures/teaser/figure.tex
\setlength{\tabcolsep}{1pt}
\resizebox{\linewidth}{!}{%
    \begin{tabular}{cccc@{\hskip 0.1in}cccc}
        \includegraphics[width=0.166\linewidth]{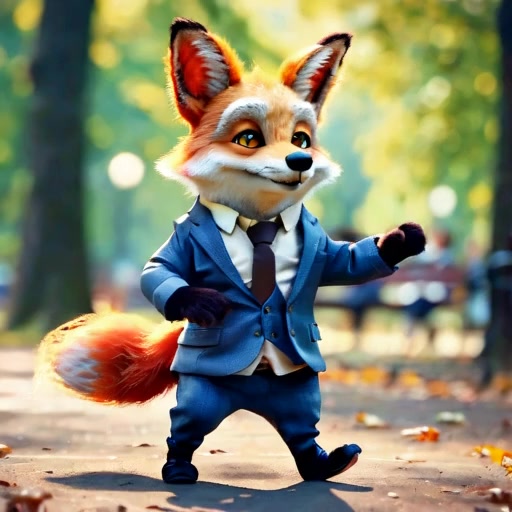} &
        \includegraphics[width=0.166\linewidth]{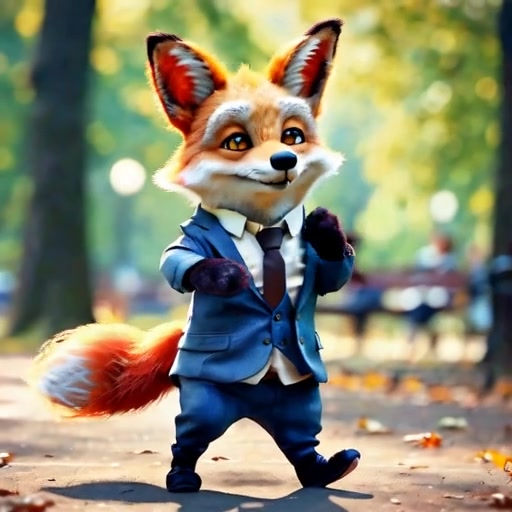} &
        \includegraphics[width=0.166\linewidth]{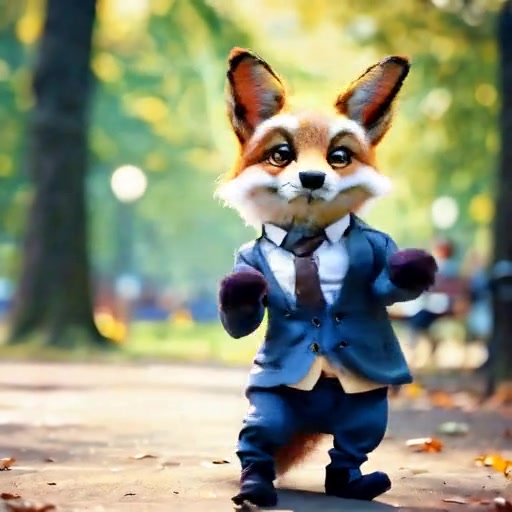} &
        \includegraphics[width=0.166\linewidth]{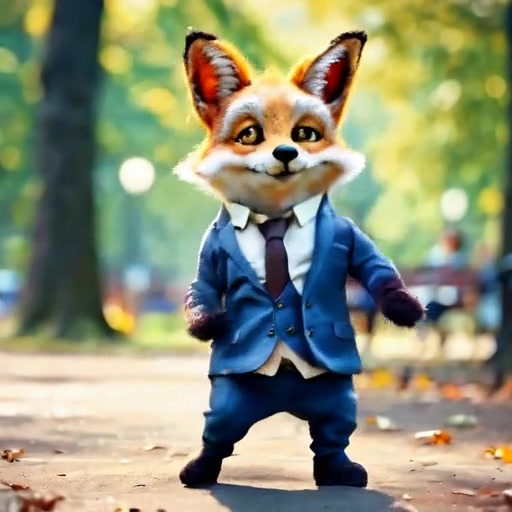} &
        \includegraphics[width=0.166\linewidth]{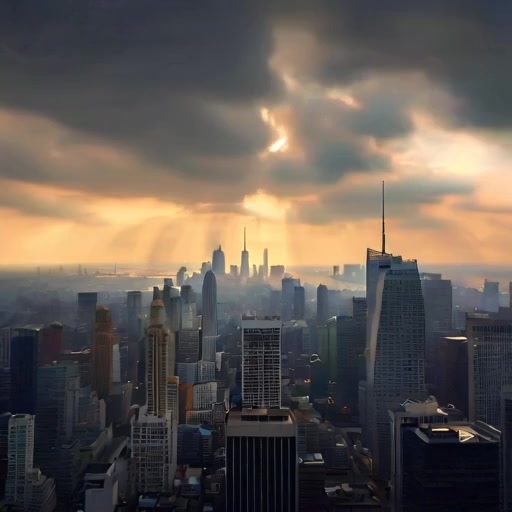} &
        \includegraphics[width=0.166\linewidth]{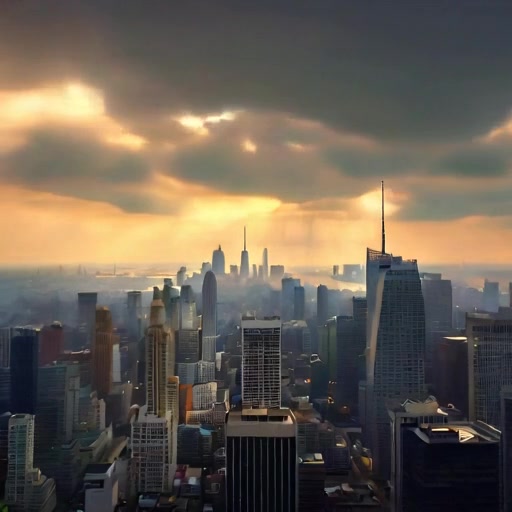} &
        \includegraphics[width=0.166\linewidth]{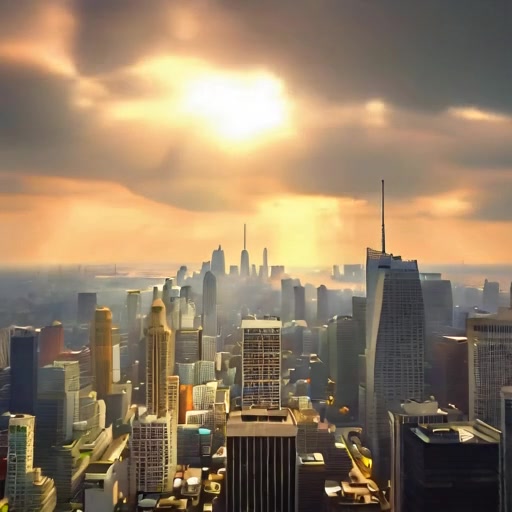} &
        \includegraphics[width=0.166\linewidth]{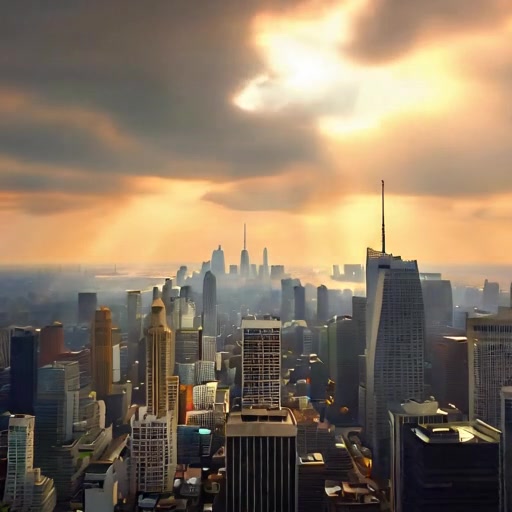} \\
        \includegraphics[width=0.166\linewidth]{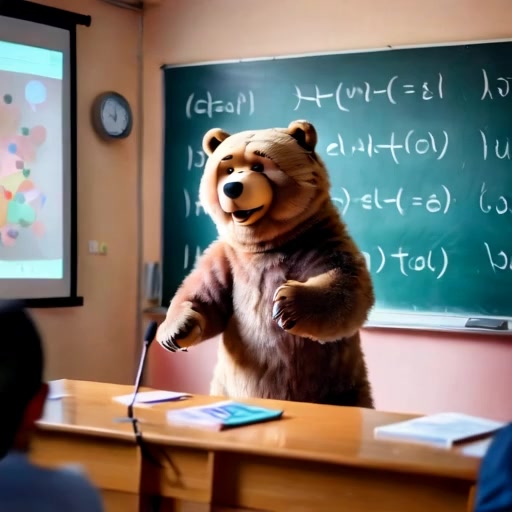} &
        \includegraphics[width=0.166\linewidth]{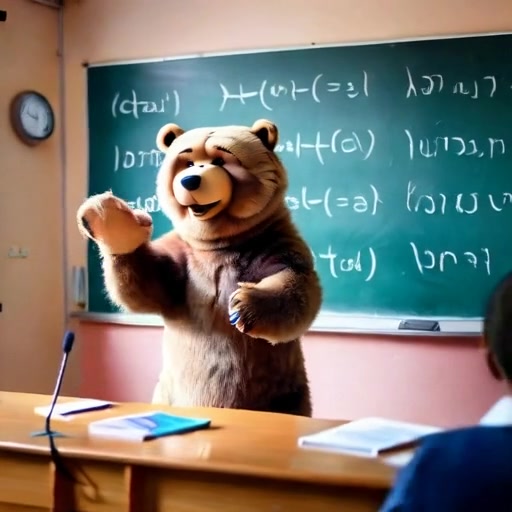} &
        \includegraphics[width=0.166\linewidth]{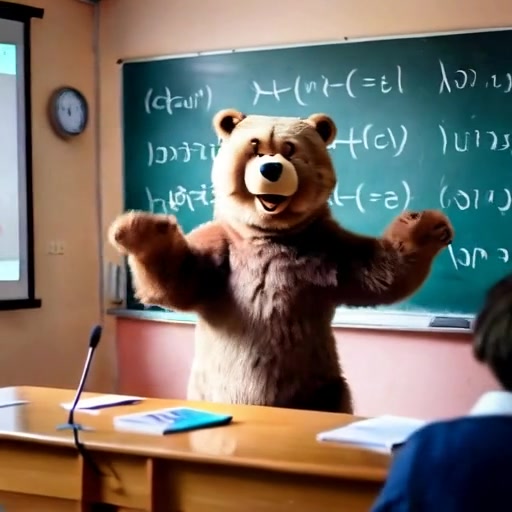} &
        \includegraphics[width=0.166\linewidth]{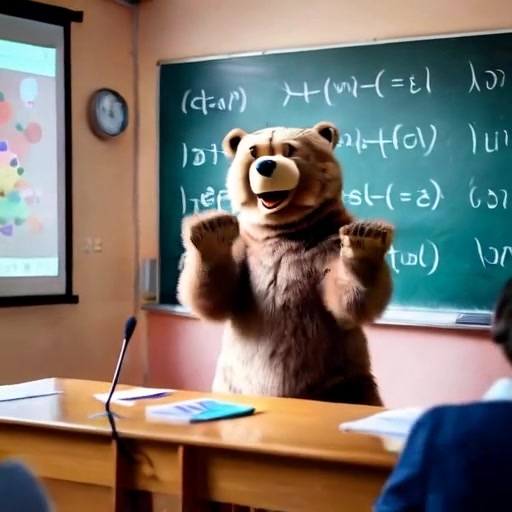} &
        \includegraphics[width=0.166\linewidth]{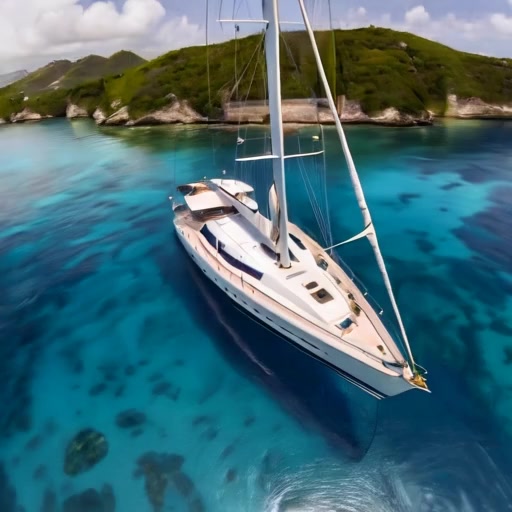} &
        \includegraphics[width=0.166\linewidth]{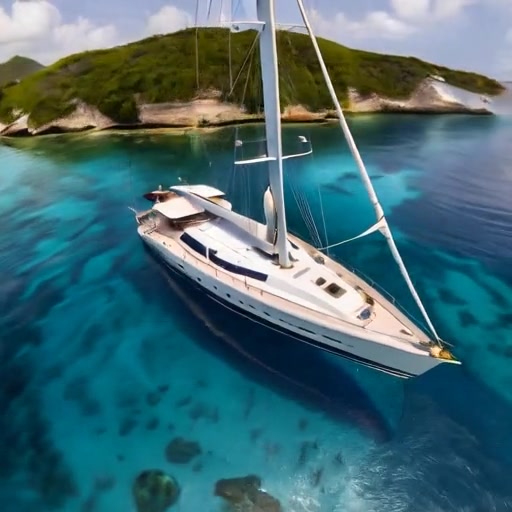} &
        \includegraphics[width=0.166\linewidth]{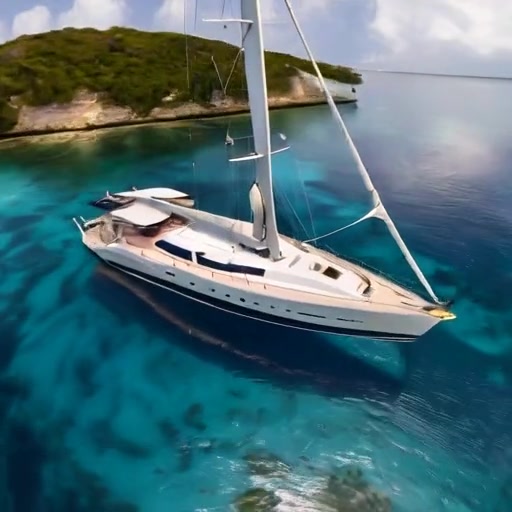} &
        \includegraphics[width=0.166\linewidth]{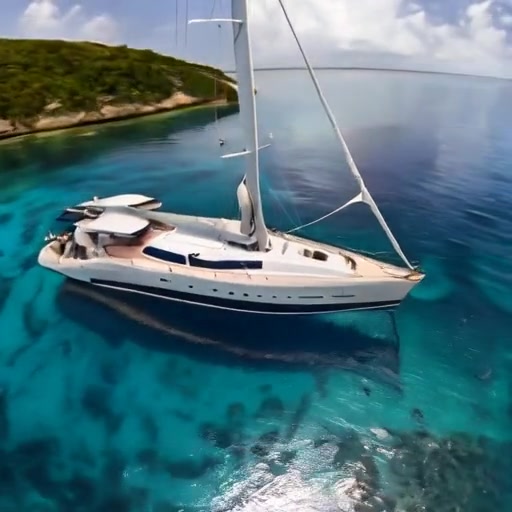} \\[0.1in]
        \setlength{\fboxrule}{2pt}
        \setlength{\fboxsep}{0pt}
        \hspace{-2\fboxrule-2\fboxsep}\raisebox{\fboxrule}{\fcolorbox{eccvblue}{white}{%
            \includegraphics[width=0.166\linewidth-2\fboxsep-2\fboxrule,height=0.12\linewidth-2\fboxsep-2\fboxrule]{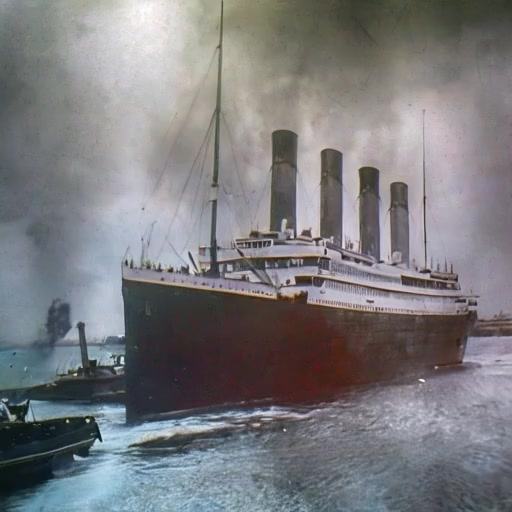}}%
        }
        &
        \includegraphics[width=0.166\linewidth,height=0.12\linewidth]{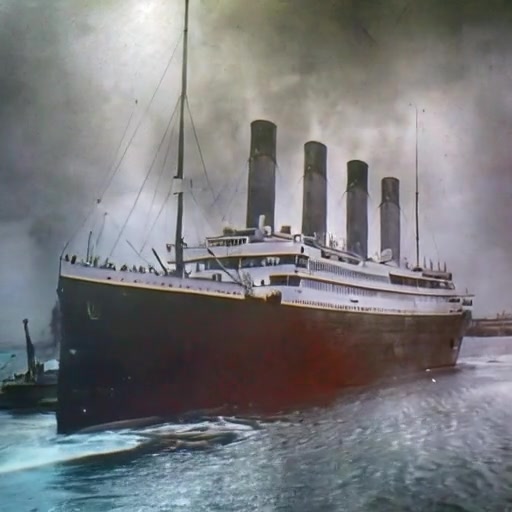} &
        \includegraphics[width=0.166\linewidth,height=0.12\linewidth]{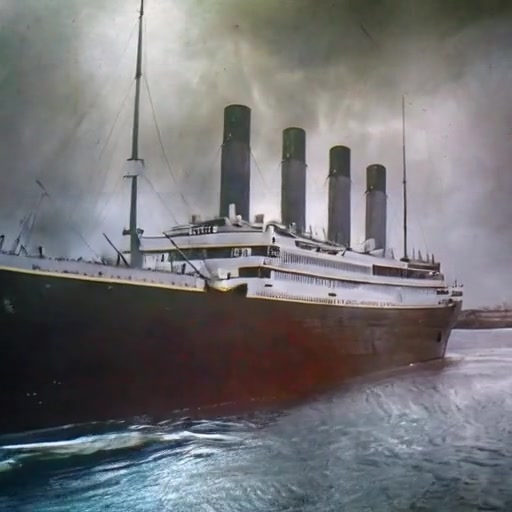} &
        \includegraphics[width=0.166\linewidth,height=0.12\linewidth]{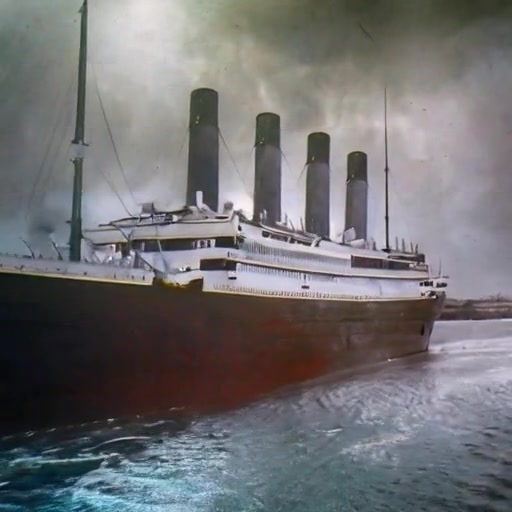} &
        \setlength{\fboxrule}{2pt}
        \setlength{\fboxsep}{0pt}
        \hspace{-2\fboxrule-2\fboxsep}\raisebox{\fboxrule}{\fcolorbox{eccvblue}{white}{%
            \includegraphics[width=0.166\linewidth-2\fboxsep-2\fboxrule,height=0.12\linewidth-2\fboxsep-2\fboxrule]{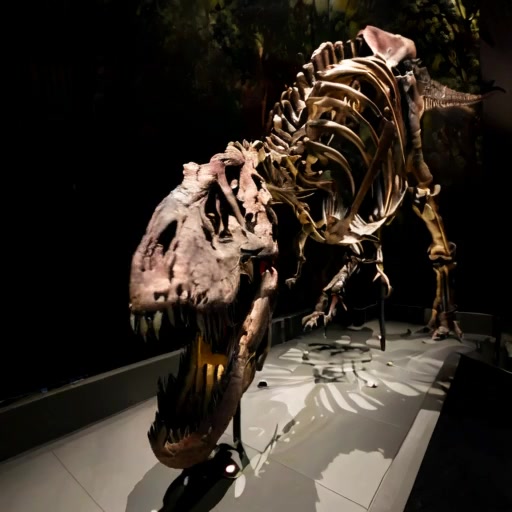}}%
        }
        &
        \includegraphics[width=0.166\linewidth,height=0.12\linewidth]{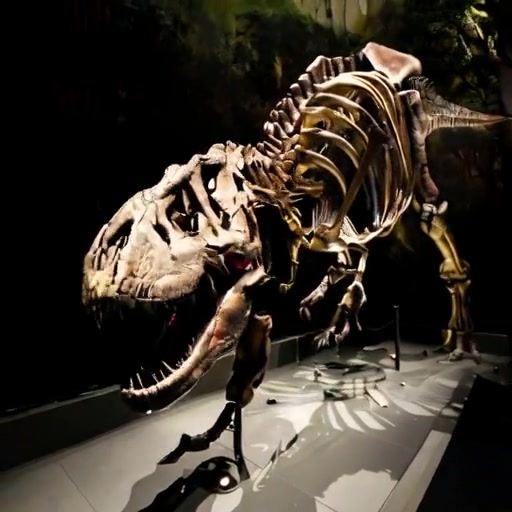} &
        \includegraphics[width=0.166\linewidth,height=0.12\linewidth]{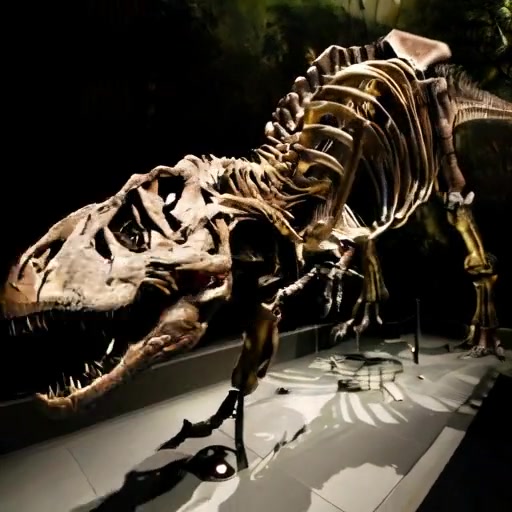} &
        \includegraphics[width=0.166\linewidth,height=0.12\linewidth]{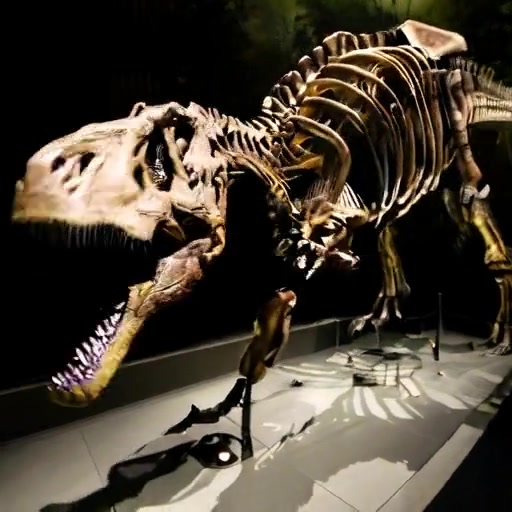} \\
    \end{tabular}%
}

%% file: sections/abstract.tex
\begin{abstract}
We present \OURS{},
a text-to-video generation model %
that factorizes the generation into two steps: first generating an image conditioned on the text, and then generating a video conditioned on the text and the generated image.
We identify critical design decisions--adjusted noise schedules for diffusion, and multi-stage training--that enable us to directly generate high quality and high resolution videos, without requiring a deep cascade of models as in prior work.
In human evaluations, our generated videos are strongly preferred in quality compared to all prior work--$81\%$ vs.\ Google's \imagenvideo, $90\%$ vs.\ Nvidia's PYOCO, and $96\%$ vs.\ Meta's Make-A-Video.
Our model outperforms commercial solutions such as RunwayML's Gen2 and Pika Labs.
Finally, our factorizing approach naturally lends itself to
animating images based on a user's text prompt, where our generations are preferred $96\%$ over prior work.
\end{abstract}

%% file: sections/intro.tex
\section{Introduction}
\label{sec:intro}

Large \textToI models~\cite{rombach2021highresolution,ho2022imagen,nichol2021glide,donahue2017adversarial,gafni2022make,dai2023emu} trained on web-scale image-text pairs generate diverse and high quality images.
While these models can be further adapted for \textToV (\textToVShort) generation~\cite{singer2023makeavideo,Blattmann2023AlignYL,ho2022imagen,ge2023preserve,hong2022cogvideo} by using video-text pairs, video generation still lags behind image generation in terms of quality and diversity.
Compared to image generation, video generation is more challenging as it requires modeling a higher dimensional spatiotemporal output space while still being conditioned only on a text prompt.
Moreover, video-text datasets are typically an order of magnitude smaller than image-text datasets~\cite{dai2023emu,singer2023makeavideo,ho2022imagen}.

The dominant paradigm in video generation uses diffusion models~\cite{singer2023makeavideo,ho2022imagen} to generate all video frames at once.
In stark contrast, in NLP, long sequence generation is formulated as an autoregressive problem~\cite{brown2020language}: predicting one word conditioned on previously predicted words. Thus, the conditioning signal for each subsequent prediction progressively gets stronger.
We hypothesize that strengthening the conditioning signal is also important for high quality video generation, which is inherently a time-series.
However, autoregressive decoding with diffusion models~\cite{voleti2022MCVD} is challenging since generating a single frame from such models itself requires many iterations. %

We propose \OURS to strengthen the conditioning for diffusion based \textToV generation with an explicit intermediate image generation step.
Specifically, we factorize \textToV generation into two subproblems: (1) generating an image from an input text prompt; (2) generating a video based on the stronger conditioning from the image \emph{and} the text.
Intuitively, giving the model a starting image and text makes video generation easier since the model only needs to predict how the image will evolve in the future.

Since video-text datasets are much smaller than image-text datasets, we initialize~\cite{singer2023makeavideo,Blattmann2023AlignYL} our \textToVShort model using a pretrained \textToI (\textToIShort) model whose weights are frozen.
Unlike direct \textToVShort methods~\cite{singer2023makeavideo,ho2022imagen}, at inference, our factorized approach explicitly generates an image, allowing us to easily retain the visual diversity, style, and quality of the \textToI model (see~\cref{fig:teaser}).
This allows \OURS to outperform direct \textToVShort methods, even when accounting for the same amount of training data, compute, and trainable parameters.

\input{figures/comparison_teaser.tex}

\par \noindent \textbf{Contributions.}
We show that \textToV (\textToVShort) generation quality can be greatly improved by factorizing the generation into first generating an image and using the generated image and text to generate a video.
We identify critical design decisions--changes to the diffusion noise schedule and multi-stage training--to efficiently generate videos at a high resolution of $512$px bypassing the need for a deep cascade of models used in prior work~\cite{singer2023makeavideo,ho2022imagen}.
We design a robust human evaluation scheme--JUICE--where we ask evaluators to justify their choices when making the selection in the pairwise comparisons of video generations.
~\cref{fig:teaser_compare} shows that \OURS significantly \emph{surpasses all prior work} including commercial solutions: an average win rate of $91.8\%$ for quality and $86.6\%$ for text faithfulness.
Beyond \textToVShort, \OURS can be used out-of-the-box for \imageToV where the model generates a video based on a user-supplied image and a text prompt.
In this setting, \OURS's generations are preferred $96\%$ of the times over VideoComposer~\cite{2023videocomposer}.

%% file: figures/comparison_teaser.tex
\begin{figure}

\centering
  \includegraphics[width=0.6\linewidth]{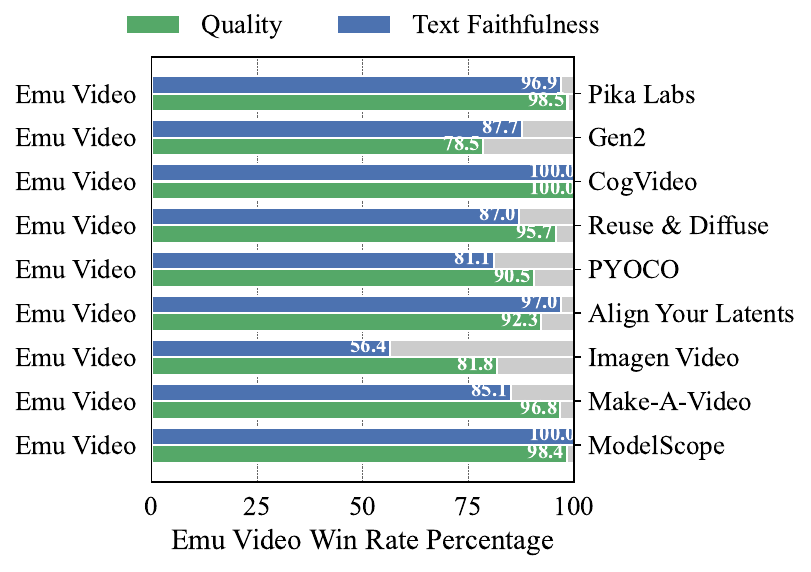}
  \caption{\textbf{\OURS \vs prior work} in \textToV in terms of video quality and text faithfulness win-rates evaluated by majority score of human evaluator preferences.
  Since most models from prior work are not accessible, we use the videos released by each method and their associated text prompt.
  The released videos are likely the \textit{best} generations and we compare without any cherry-picking of our own generations.
  We also compare to commercial solutions (\gen~\cite{gen} and \pika~\cite{pikalabs}) and the open source model CogVideo~\cite{hong2022cogvideo} and ModelScope~\cite{modelscope} using the prompt set from~\cite{Blattmann2023AlignYL}.
  \OURS significantly outperforms all prior work across both metrics.
}
  \label{fig:teaser_compare}
\end{figure}

%% file: sections/related.tex
\section{Related Work}
\label{sec:related}
\par \noindent \textbf{\TextToI (\textToIShort) diffusion models.}
Diffusion models~\cite{pmlr-v37-sohl-dickstein15} are a state-of-the-art approach for \textToIShort generation, and out-perform prior GAN~\cite{Sauer2023ICML,kang2023gigagan,brock2018large} or auto-regressive methods~\cite{ramesh2021zeroshot,Esser2021TamingTF,Aghajanyan2022CM3AC,make_a_scene}.
Diffusion models learn a data distribution by gradually denoising a normally distributed variable, often called `noise', to generate the output.
Prior work either denoises in the pixel space with pixel diffusion models~\cite{ramesh2022hierarchical,saharia2022photorealistic,ho2020denoising,ho2021cascaded,dhariwal2021diffusion,nichol2022glide}, or in a lower-dimensional latent space with latent diffusion models~\cite{rombach2021highresolution,dai2023emu}.
In this work, we leverage latent diffusion models for video generation.

\input{figures/approach_overview.tex}

\par \noindent \textbf{Video generation/prediction.}
Many prior works target the constrained settings of unconditional generation, or video prediction~\cite{10.5555/3454287.3455325,Kumar2020VideoFlow:,mathieu2016deep}.
These approaches include training VAEs~\cite{babaeizadeh2018stochastic,pmlr-v80-denton18a,babaeizadeh2021fitvid}, auto-regressive models~\cite{Ranzato2014VideoM,10.5555/3157096.3157104,pmlr-v70-kalchbrenner17a,hong2022cogvideo,yan2021videogpt}, masked prediction~\cite{gupta2023maskvit,Fu_2023_CVPR,52431}, LSTMs~\cite{Wichers2018HierarchicalLV,NIPS2015_07563a3f}, or GANs~\cite{10.1145/3487891,clark2019adversarial,DBLP:conf/nips/VondrickPT16,brooks2022generating}.
However, these approaches are trained/evaluated on limited domains.
In this work, we target the broad task of open-set \textToVShort generation.

\par \noindent \textbf{\TextToV(\textToVShort) generation.}
Most prior works tackle \textToVShort generation by leveraging \textToIShort models.
Several works take a training-free approach~\cite{zhang2023controlvideo,lee2023aadiff,hong2023large,text2video-zero} for \textit{zero-shot} \textToVShort generation by injecting motion information in the \textToIShort models.
Tune-A-Video~\cite{wu2023tune} targets \textit{one-shot} \textToVShort generation by fine-tuning a \textToIShort model with a single video.
While these methods require no or limited training, the quality and diversity of the generated videos is limited.

Many prior works instead improve \textToVShort generation by learning a \textit{direct mapping} from the text condition to the generated videos by introducing temporal parameters to a \textToIShort model~\cite{laptev2003space,Blattmann2023AlignYL,ge2023preserve,NEURIPS2022_b2fe1ee8,NEURIPS2022_39235c56,yang2022diffusion,yin2023nuwaxl,tang2023anytoany,hong2022cogvideo,villegas2023phenaki,Wu2021GODIVAGO,voleti2022MCVD}.
\mav~\cite{singer2023makeavideo} utilizes a pre-trained T2I model~\cite{ramesh2022hierarchical} and the prior network of~\cite{ramesh2022hierarchical} to train T2V generation without paired video-text data.
\imagenvideo~\cite{ho2022imagen} builds upon the Imagen T2I model~\cite{saharia2022photorealistic} with a cascade of diffusion models~\cite{NEURIPS2022_39235c56,ho2021cascaded}.
To address the challenges of modeling the high-dimensional spatiotemporal space, several works instead train \textToVShort diffusion models in a lower-dimensional latent space~\cite{Blattmann2023AlignYL,ge2023preserve,xing2023simda,gu2023reuse,fei2023empowering,he2023latent,an2023latentshift}, by adapting latent diffusion \textToIShort models.
Blattmann \etal~\cite{Blattmann2023AlignYL} freeze the parameters of a pre-trained T2I model and train new temporal layers, whilst Ge \etal~\cite{ge2023preserve} build on~\cite{Blattmann2023AlignYL} and design a noise prior tailored for \textToVShort generation.
The limitation of these approaches is that learning a direct mapping from text to the high dimensional video space is challenging.
We instead strengthen our conditioning signal by taking a factorization approach. %
Unlike prior work that enhancing the conditions for \textToVShort generation including leveraging large language models (LLMs) to improve textual description and understanding~\cite{fei2023empowering,hong2023large,lian2023llm}, or adding temporal information as conditions~\cite{chen2023control,2023videocomposer,zhang2023controlvideo,yin2023dragnuwa},
our method does not require any models to generate the conditions as we use the first frame of a video as the image condition.

\par \noindent \textbf{Factorized generation.}
The most similar works to \OURS, in terms of factorization, is CogVideo~\cite{hong2022cogvideo} and \mav~\cite{singer2023makeavideo}.
CogVideo builds upon the pretrained T2I model~\cite{ding2022cogview2} for \textToVShort generation using auto-regressive Transformer.
The auto-regressive nature is fundamentally different to our explicit image conditioning in both training and inference stages.
\mav~\cite{singer2023makeavideo} leverages the image embedding condition learnt from a shared image-text space.
Our factorization leverage the first frame as is, which is a stronger condition.
Moreover, \mav initializes from a pretrained \textToIShort model but finetunes all the parameters so it cannot retain the visual quality and diversity of the \textToIShort model as we do. \sdvideo~\cite{sdvideo} is a concurrent work that introduces similar factorization as ours for T2V generation.

%% file: figures/approach_overview.tex
\begin{figure*}[!t]
    \centering
    \includegraphics[width=\linewidth]{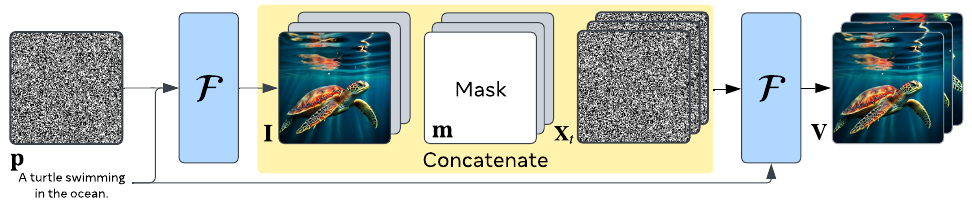}
    \caption{
    \textbf{Factorized \textToV generation}
    involves first generating
    an image $\bi$ conditioned on the text $\bp$, and then using stronger conditioning--the generated image \emph{and} text--to generate a video $\bv$.
    To condition our model $\extrapolateModel$ on the image, we zero-pad the image temporally and concatenate it with a binary mask $\bm$ indicating which frames are zero-padded, and the noised input. %
    }
    \label{fig:approach}
\end{figure*}

%% file: sections/approach.tex
\section{Approach}
\label{sec:approach}

The goal of \textToV (\textToVShort) generation is to construct a model
that takes as input a text prompt $\bp$ to generate a video $\bv$ consisting of $\vTime$ RGB frames.
Recent methods~\cite{singer2023makeavideo,Blattmann2023AlignYL,ge2023preserve,ho2022imagen} directly
generate the $\vTime$ video frames at once using text-only conditioning.
Our approach builds on the hypothesis that stronger conditioning by way of both text \emph{and} image can improve video
generation (\cf ~\cref{sec:approach-details}).

\subsection{Preliminaries}
Conditional Diffusion Models~\cite{pmlr-v37-sohl-dickstein15,ho2020denoising} are a class of generative models that are trained to generate the output using a conditional input $\bc$ by iteratively denoising from gaussian noise.
At training time, time-step $t \in [0,\finalDiffusionTime]$ dependent gaussian noise $\beps_t \sim \mathcal{N}(\mathbf{0}, \mathbf{1})$ is added to the original input signal $\bx$ to obtain a noisy input $\bx_t = \alpha_t \bx + \sqrt{1-\alpha_t} \beps_t$.
$\alpha_t$ defines the ``noise schedule'', \ie, noise added at timestep $t$ and $\finalDiffusionTime$ is the total number of diffusion steps.
The diffusion model is trained to denoise $\bx_t$ by predicting either $\beps_t$, $\bx$, or $v_t = \alpha_t \beps_t - \sqrt{1-\alpha_t} \bx$ (called v-prediction~\cite{salimans2022progressive}).
The signal-to-noise ratio (SNR) at timestep $t$ is given by $(\frac{\alpha_t}{1-\alpha_t})^2$ and decreases as $t \rightarrow \finalDiffusionTime$.
At inference, samples are generated by starting from pure noise $\bx_\finalDiffusionTime \sim \mathcal{N}(\mathbf{0}, \mathbf{1})$ and denoising it.
Note that at inference time $\bx_\finalDiffusionTime$ has no signal, \ie, zero SNR which has significant implications for video generation as we describe in~\cref{sec:approach-details}.

\subsection{{\bf \large \OURS}}
\label{sec:approach-details}

We factorize \textToV generation into two steps (1) generating the first frame (image) given the text prompt $\bp$ and (2) generating $\vTime$ frames of a video by leveraging the text prompt and the image conditioning.
We implement both steps using a latent diffusion model $\extrapolateModel$, illustrated in~\cref{fig:approach}.
We initialize $\extrapolateModel$ with a pre-trained text-to-image model to ensure that it is capable of generating images at initialization.
Thus, we only need to train $\extrapolateModel$ to solve the second step, \ie, extrapolate a video conditioned on a text prompt and a starting frame.
We train $\extrapolateModel$ using video-text pairs by sampling a starting frame $\bi$ and asking the model to predict the $\vTime$ frames using both the text prompt $\bp$ and the image $\bi$ conditioning.
We denote a video $\bv$ consisting of $\vTime$ RGB frames of spatial dimensions $H', W'$ as a $4$D tensor of shape $\vTime \times 3 \times H' \times W'$.
Since we use latent diffusion models, we first convert the video $\bv$ into a latent space $\bx \in \mathbb{R}^{\vTime \times \channels \times \zHeight \times \zWidth}$ using a image autoencoder applied frame-wise, which reduces the spatial dimensions.
The latent space can be converted back to the pixel space using the autoencoder's decoder.
The $T$ frames of the video are noised independently to produce the noised input $\bx_t$, which the diffusion model is trained to denoise.

\par \noindent \textbf{Image conditioning.}
We condition on the starting frame, $\bi$, by concatenating it with the noise.
Our design allows the model to use all the information in $\bi$ unlike~\cite{2023videocomposer,singer2023makeavideo} that lose image information by conditioning on a semantic embedding.
We represent $\bi$ as a single-frame video ($\vTime = 1$) and zero-pad it to obtain a $\vTime\times\channels\times\zHeight\times\zWidth$ tensor.
We use a binary mask $\bmask$ of shape $\vTime\times1\times\zHeight\times\zWidth$ that is set to $1$ at the first temporal position to indicate the position of the starting frame, and zero otherwise.
The mask $\bmask$, starting frame $\bi$, and the noised video $\bx_t$ are concatenated channel-wise as the input to the model.%

\par \noindent \textbf{Model.}
We initialize our latent diffusion model $\extrapolateModel$ using the pretrained \textToIShort model~\cite{dai2023emu}.
Like prior work~\cite{singer2023makeavideo}, we add new learnable temporal parameters: a $1$D temporal convolution after every spatial convolution,
and a $1$D temporal attention layer after every spatial attention layer. %
The original spatial convolution and attention layers are applied to each of the $\vTime$ frames independently and are kept frozen.
The pretrained \textToIShort model is text conditioned and combined with the image conditioning (above), $\extrapolateModel$ is conditioned on both text and image.

\par \noindent \textbf{Zero terminal-SNR noise schedule.}
We found that the diffusion noise schedules used in prior work~\cite{rombach2021highresolution,dai2023emu} have a train-test discrepancy which prevents high quality video generation (reported for images in~\cite{lin2023common,chen2023importance}).
At training, the noise schedule leaves some residual signal, \ie, has non-zero signal-to-noise (SNR) ratio even at the terminal diffusion timestep $\finalDiffusionTime$.
This prevents the diffusion model from generalizing at test time when we sample from random gaussian noise with no signal about real data.
The residual signal is higher for high resolution video frames, due to redundant pixels across both space and time.
We resolve this issue by scaling the noise schedule and setting the final $\alpha_{\finalDiffusionTime}=0$~\cite{lin2023common}, which leads to zero SNR at the terminal timestep $\finalDiffusionTime$ during training too.
We find that this design decision is \emph{critical} for high resolution video generation.

\par \noindent \textbf{Interpolation model.}
We use an interpolation model $\interpolateModel$, architecturally the same as $\extrapolateModel$, to convert a low frame-rate video of $\vTime$ frames into a high frame-rate video of $\iTime$ frames.
The interpolation model
operates on $\iTime\times\channels\times\zHeight\times\zWidth$ inputs/outputs.
For frame conditioning, the input $\vTime$ frames are zero-interleaved to produce $\iTime$ frames, and a binary mask $\bmask$ indicating the presence of the $\vTime$ frames are concatenated to the noised input (similar to the image conditioning for $\extrapolateModel$).
The model is trained on video clips of $\iTime$ frames of which $\vTime$ frames are fed as input.
For efficiency, we initialize $\interpolateModel$ from $\extrapolateModel$ and only train the temporal parameters of the model $\interpolateModel$ for the interpolation task.

\par \noindent \textbf{Simplicity in implementation.}
\OURS can be trained using standard video-text datasets, and does not require a deep cascade of models, \eg, $7$ models in~\cite{ho2022imagen}, for generating high resolution videos.
At inference, given a text prompt, we run $\extrapolateModel$ without the temporal layers to generate an image $\bi$.
We then use $\bi$ and the text prompt as input to $\extrapolateModel$ to generate $\vTime$ video frames, directly at high resolution.
We can increase the fps of the video using $\interpolateModel$.
Since the spatial layers are initialized from a pretrained \textToIShort model and kept frozen, our model retains the conceptual and stylistic diversity learned from large image-text datasets, and uses it to generate $\bi$.
This comes at no additional training cost unlike~\cite{ho2022imagen} that jointly finetune on image and video data to maintain such style.
Many direct \textToVShort approaches~\cite{singer2023makeavideo,Blattmann2023AlignYL} also initialize from a pretrained \textToIShort model and keep the spatial layers frozen.
However, they do not employ our image-based factorization failing to retain the quality and diversity in the \textToIShort model.

\par \noindent \textbf{Robust human evaluation (\juice).}
Similar to recent studies~\cite{dai2023emu,podell2023sdxl,singer2023makeavideo,ho2022imagen}, we find that the automatic evaluation metrics~\cite{unterthiner2019fvd} do not reflect improvements in quality.
We primarily use human evaluation to measure \textToVShort generation performance on two orthogonal aspects - (a) video generation quality denoted as \quality (\qualityShort) and (b) the alignment or `faithfulness' of the generated video to the text prompt, denoted as \faithfulness (\faithfulnessShort).
We found that asking human evaluators to JUstify their choICE (\juice)
when picking a generation over the other significantly improves the inter-annotator agreement (details in~\cref{appendix:human_eval}).
The annotators select one or more pre-defined reasons to justify their choice.
The reasons for picking one generation over the other for \quality are: pixel sharpness, motion smoothness, recognizable objects/scenes, frame consistency, and amount of motion.
For \faithfulness we use two reasons: spatial text alignment, and temporal text alignment.

\subsection{Implementation Details}
\label{sec:impl_details}

We provide complete implementation details in the Appendix~\cref{appendix:implementation_details} and highlight salient details next.

\par \noindent \textbf{Architecture and initialization.}
We adapt the \textToI \unet architecture from~\cite{dai2023emu} for our model and initialize all the spatial parameters with the pretrained model.
The pretrained model produces square $512$px images using an $8$ channel $64\times64$ latent as the autoencoder downsamples spatially by $8\times$.
The model uses both a frozen T5-XL~\cite{chung2022scaling}
and a frozen CLIP~\cite{radford2021learning} text encoder to extract features from the text prompt.
Separate cross-attention layers in the \unet attend to each of the text features.
After initialization, our model contains $2.7$B frozen spatial parameters, and $1.7$B trainable temporal parameters.

The temporal parameters are initialized as identity operations: identity kernels for convolution, and zeroing the final MLP layer of the temporal attention block.
In our preliminary experiments, the identity initialization improved the model convergence by $2\times$.
For the additional channels in the model input due to image conditioning, we add $C+1$ additional learnable channels (zero-initialized) to the kernel of the first spatial convolution layer.
Our model produces $512$px square videos of $\vTime=8$ or $16$ frames and is trained with square center-cropped video clips of $1$, $2$ or $4$ seconds sampled at $8$fps or $4$fps.
We train all our models with a batch size of $512$ and describe the details next.

\input{tables/ablation_keyframe_all.tex}

\par \noindent \textbf{Efficient multi-stage multi-resolution training.}
To reduce the computational complexity, we train in two stages - (1) for majority of the training iterations ($70$K) we train for a simpler task: $256$px $8$fps $1$s videos, which reduces per-iteration time by $3.5\times$ due to the reduction in spatial resolution; (2) we then train the model at the desired $512$px resolution on $4$fps $2$s videos for $15$K iterations.
The change in spatial resolution does not affect the $1$D temporal layers.
Although the frozen spatial layers were pretrained at $512$px, changing the spatial resolution at inference to $256$px led to no loss in generation quality.
We use the noise schedule from~\cite{rombach2021highresolution} for $256$px training, and with zero terminal-SNR for $512$px training using the v-prediction objective~\cite{salimans2022progressive} with $\finalDiffusionTime=1000$ steps for the diffusion training.
We sample from our models using $250$ steps of DDIM~\cite{song2020denoising}.
Optionally, to increase duration, we further train the model on $16$ frames from a $4$s video clip for $25$K iterations.

\par \noindent \textbf{Finetuning for higher quality.}
Similar to the observation in image generation~\cite{dai2023emu}, we find that the motion of the generated videos can be improved by finetuning the model on a small subset of high motion and high quality videos.
We automatically identify a small finetuning subset of $1.6$K videos from our training set which have high motion (computed using motion signals stored in \texttt{H.264} encoded videos).
We follow standard practice~\cite{rombach2021highresolution} and also apply filtering based on aesthetic scores~\cite{rombach2021highresolution} and CLIP~\cite{radford2021learning} similarity between the video's text and first frame. Specifically, we use a video with $N$ frames $\{f_j\}$ if $\text{CLIP}(f_1)>0.25$, $\text{aesthetic}(f1)>5.7$, $\min_{j=1}^{N-5}\sum_{i=j}^{j+5}(\text{motion score}(f_i))>0.5$.

\par \noindent \textbf{Interpolation model.}
We initialize the interpolation model from the video model $\extrapolateModel$.
Our interpolation model takes $8$ frames as input and outputs $\iTime\!=\!37$ frames at $16$fps.
During training, we use noise augmentation~\cite{ho2021cascaded} where we add noise to the frame conditioning by randomly sampling timesteps $t \in \{0,...250\}$.
At inference time, we noise augment the samples from $\extrapolateModel$ with $t = 100$.

%% file: tables/ablation_keyframe_all.tex
\begin{table*}[!t]
	\centering
	\resizebox{\linewidth}{!}{%
	\subfloat[
	\label{tab:ablate_keyframe_t2v_vs_t2i2v}
	]{
		\centering
                    \input{tabulars/ablation_t2v_vs_t2i2v.tex}
	}
    \subfloat[
	\label{tab:ablate_keyframe_zero_snr}
	]{
		\centering
                    \input{tabulars/ablation_zero_snr.tex}
	}
	\subfloat[
	\label{tab:ablate_keyframe_multi_stage}
	]{
		\centering
                    \input{tabulars/ablation_multi_stage_training.tex}
	}
	\subfloat[
	\label{tab:ablate_keyframe_finetuning}
	]{
		\centering
                    \input{tabulars/ablation_finetuning.tex}
	}
	\subfloat[
	\label{tab:ablate_keyframe_parameter_freezing}
	]{
		\centering
                    \input{tabulars/ablation_parameter_freeze_t2i2v.tex}
	}}
    \caption{\textbf{Key design decisions in \OURS}. Each table shows the preference, in terms of the \quality (\qualityShort) and \faithfulness (\faithfulnessShort), on adopting a design decision \vs a model that does not have it.
	Our results show clear preference to a) factorized generation that uses both image and text conditioning (against a direct video generation baseline that is only text conditioned),
	b) adopting zero terminal-SNR noise schedule for directly generating high resolution $512$px videos, c) adopting the multi-stage training setup compared to training directly at the high resolution, d) incorporating the high quality (HQ) finetuning, and e) freezing the spatial parameters. See~\cref{sec:ablations} for details.
    }
	\label{tab:ablate_keyframe_all}

\end{table*}

%% file: tabulars/ablation_t2v_vs_t2i2v.tex
\begin{tabular}{c | cc}
    \bf Method & \bf \qualityShort & \bf \faithfulnessShort \\
    \midrule
    Factorized & $70.5$ & $63.3$
    \\
\end{tabular}%

%% file: tabulars/ablation_zero_snr.tex
\begin{tabular}{c | cc}
    \bf Method & \bf \qualityShort & \bf \faithfulnessShort \\
    \midrule
    Zero SNR & $96.8$ & $88.3$
    \\
\end{tabular}%

%% file: tabulars/ablation_multi_stage_training.tex
\begin{tabular}{c | cc}
    \bf Method & \bf \qualityShort & \bf \faithfulnessShort \\
    \midrule
    Multi-stage & $81.8$ & $84.1$
    \\
\end{tabular}%

%% file: tabulars/ablation_finetuning.tex
\begin{tabular}{c | ccH}
    \bf Method & \bf \qualityShort & \bf \faithfulnessShort & \bf \motionShort \\
    \midrule
    HQ finetuned & $65.1$ & $79.6$ & -
    \\
\end{tabular}%

%% file: tabulars/ablation_parameter_freeze_t2i2v.tex
\begin{tabular}{c | cc}
    \bf Method & \bf \qualityShort & \bf \faithfulnessShort \\
    \midrule
    Frozen spatial & $55.0$ & $58.1$
    \\
\end{tabular}%

%% file: sections/ablations.tex
\section{Experiments}
\label{sec:experiments}

\par \noindent \textbf{Dataset.}
We train \OURS on a dataset of $34$M licensed video-text pairs
Our videos are $5$-$60$ seconds long %
 and cover a variety of natural world concepts.
The videos were not curated for a particular task and were \emph{not} filtered for text-frame similarity or aesthetics.
Unless noted, we train the model on the full set, and do not use the $1.6$K high motion quality finetuning subset described in~\cref{sec:impl_details}.

\par \noindent \textbf{Text prompt sets for human evaluation.}
We use the text prompt sets from prior work (\cf Appendix~\cref{tab:prompt_datasets}) to generate videos.
The prompts cover a wide variety of categories that can test our model's ability to generate natural and fantastical videos, and compose different visual concepts.
We use our proposed JUICE evaluation scheme (~\cref{sec:approach}) for reliable human evaluation and use the majority vote from $5$ evaluators for each comparison.

\input{figures/qualitative_zerosnr_explicitcondition.tex}
\subsection{Ablating design decisions}
\label{sec:ablations}

We study the effects of our design decisions using the $8$ frame generation setting and report human evaluation results in~\cref{tab:ablate_keyframe_all} using pairwise comparisons on the $307$ prompt set of~\cite{singer2023makeavideo}.

\par \noindent \textbf{Factorized \vs Direct generation.}
We compare our factorized generation to a direct \textToVShort generation model that generates videos from text condition only.
We ensure that the pretrained \textToIShort model, training data, number of training iterations, and trainable parameters are held constant for this comparison.
As shown in~\cref{tab:ablate_keyframe_t2v_vs_t2i2v}, the factorized generation model's results are strongly preferred both in \quality and \faithfulness.%
The strong preference in \quality is because the direct generation model does not retain the style and quality of the \textToI model despite frozen spatial parameters, while also being less temporally consistent (examples in~\cref{fig:qual_zerosnr_t2i2v}).

\par \noindent \textbf{Zero terminal-SNR noise schedule.}
We compare using zero terminal-SNR for the high resolution $512$px training against a model that is trained with the standard noise schedule.
\cref{tab:ablate_keyframe_zero_snr} shows that generations using zero terminal-SNR are \emph{strongly} preferred.
This suggests that the zero terminal-SNR noise schedule's effect of correcting the train-test discrepancy as described in~\cref{sec:approach-details} is critical for high resolution video generation.
We also found that zero terminal-SNR has a stronger benefit for our factorized generation compared to a direct \textToVShort model possibly.
Similar to images~\cite{lin2023common}, in the direct \textToVShort case, this decision primarily affects the color composition.
For our factorized approach, this design choice was critical for object consistency and high quality as our qualitative results in~\cref{fig:qual_zerosnr_t2i2v} show.

\par \noindent \textbf{Multi-stage multi-resolution training.}
We spend most training budget ($4\!\times\!$) on the $256$px $8$fps stage compared to the $3.5\!\times\!$ slower (due to increased resolution) $512$px $4$fps stage.
We compare to a baseline that trains only the $512$px stage with the same training budget.
~\cref{tab:ablate_keyframe_multi_stage} shows that our multi-stage training yields significantly better results.%

\par \noindent \textbf{High quality finetuning.}
We study the effect of finetuning our model on automatically identified high quality videos in~\cref{tab:ablate_keyframe_finetuning}.
We found that this finetuning improves on both metrics, %
particularly the model's ability to respect the motion specified in the text prompt as reflected by the strong gain in \faithfulness.%

\par \noindent \textbf{Parameter freezing.}
We test if freezing the spatial parameters of our model affects performance by
comparing it to a model where all parameters are finetuned during the second $512$px training stage.
For a fair comparison, the same conditioning images $\bi$ are used across both models.
~\cref{tab:ablate_keyframe_parameter_freezing} suggests that freezing the spatial parameters produces better videos, while reducing training cost.%

%% file: figures/qualitative_zerosnr_explicitcondition.tex
\begin{figure*}[!t]
    \centering

\input{figures/ablation/figure.tex}
    \caption{
    \textbf{Design choices in \OURS.}
    \textit{Top row:} Direct \textToV generation produces videos that have low visual quality and are inconsistent.
    \textit{Second row:} We use a factorized \textToV approach that produces high quality videos and improves consistency.
    \textit{Third row:} Not using a zero terminal-SNR noise schedule at $512$px generation leads to significant inconsistencies in the generations.
    \textit{Bottom row:} Finetuning our model (second row) with HQ data increases the motion in the generated videos.
    }
    \label{fig:qual_zerosnr_t2i2v}
\end{figure*}
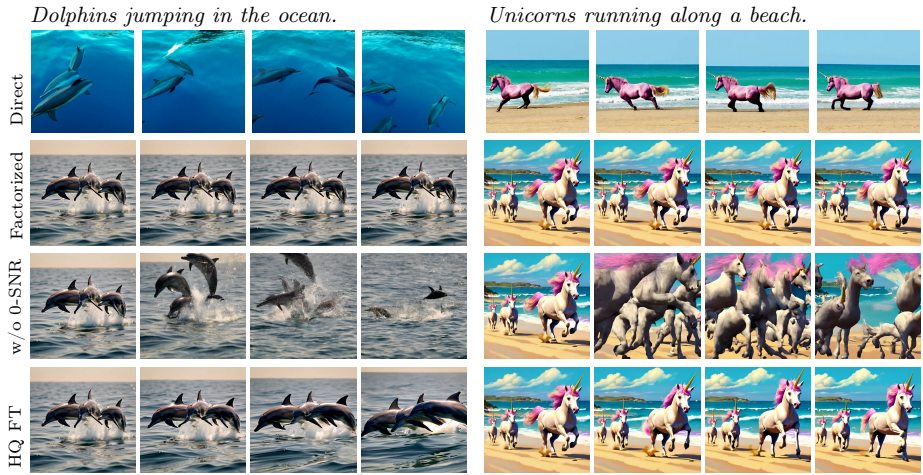

%% file: figures/ablation/figure.tex
\setlength{\tabcolsep}{1pt}
\resizebox{\linewidth}{!}{%
    \begin{tabular}{ccccc@{\hskip 0.1in}cccc}
        & \multicolumn{4}{l}{\footnotesize \it Dolphins jumping in the ocean.}
        &
        \multicolumn{4}{l}{\footnotesize  \it  Unicorns running along a beach.} \\
        \rotatebox[origin=l]{90}{\scriptsize Direct} &
        \includegraphics[width=0.12\linewidth]{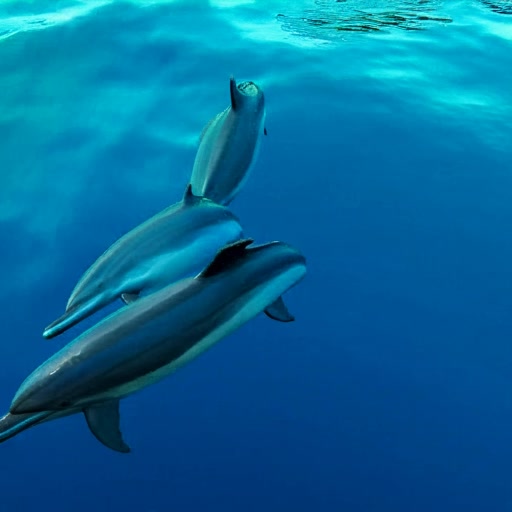} &
        \includegraphics[width=0.12\linewidth]{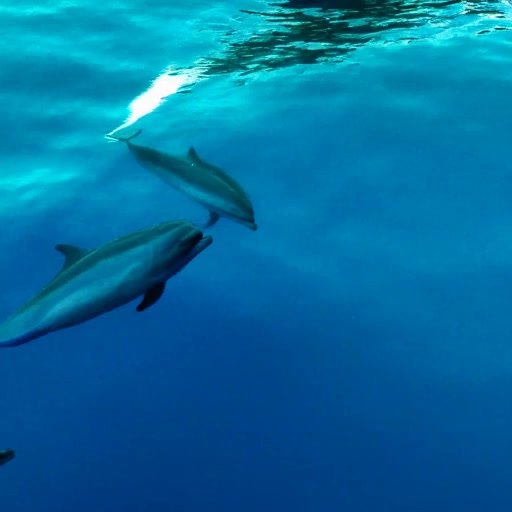} &
        \includegraphics[width=0.12\linewidth]{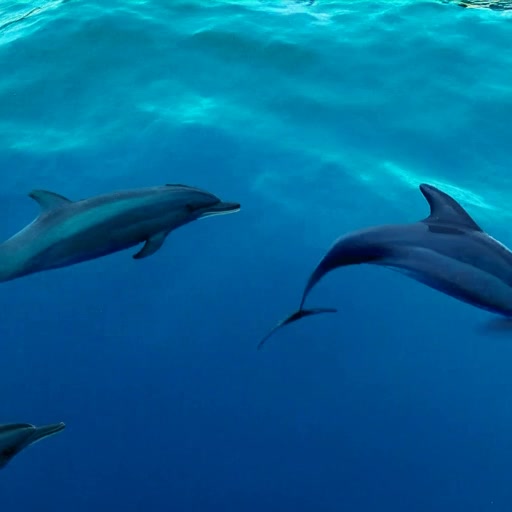} &
        \includegraphics[width=0.12\linewidth]{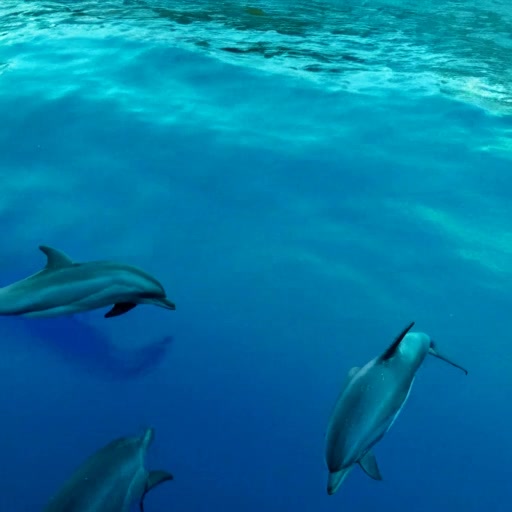} &
        \includegraphics[width=0.12\linewidth]{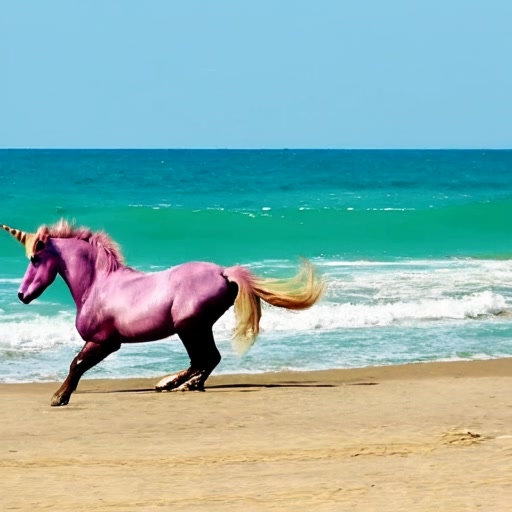} &
        \includegraphics[width=0.12\linewidth]{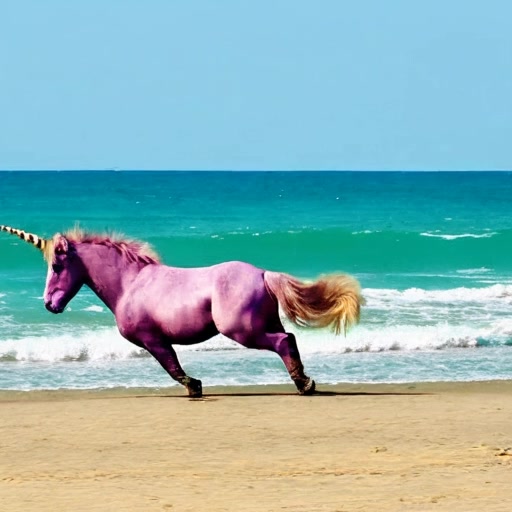} &
        \includegraphics[width=0.12\linewidth]{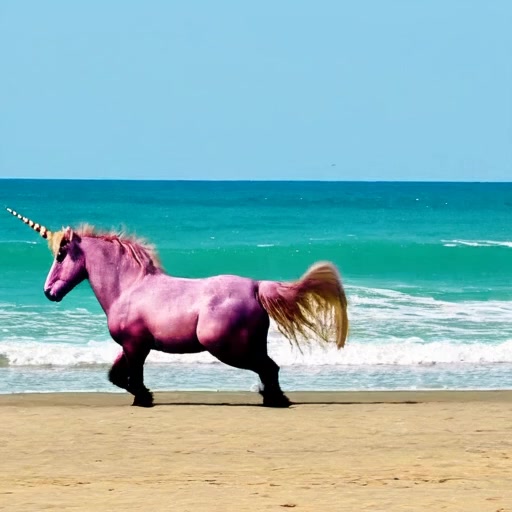} &
        \includegraphics[width=0.12\linewidth]{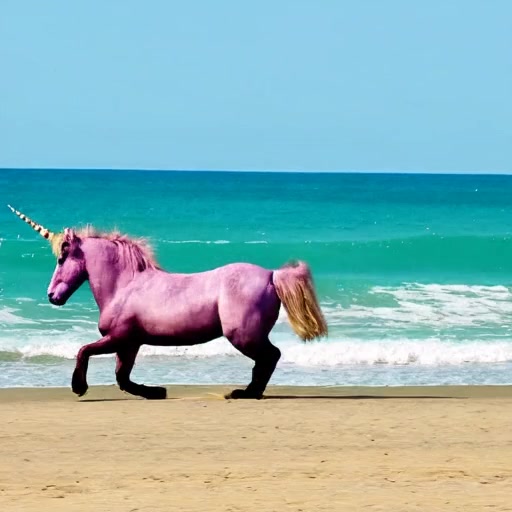} \\
        \rotatebox[origin=l]{90}{\scriptsize Factorized} &
        \includegraphics[width=0.124\linewidth]{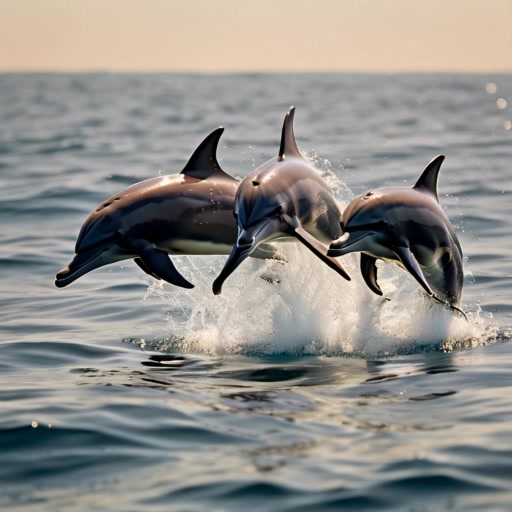} &
        \includegraphics[width=0.124\linewidth]{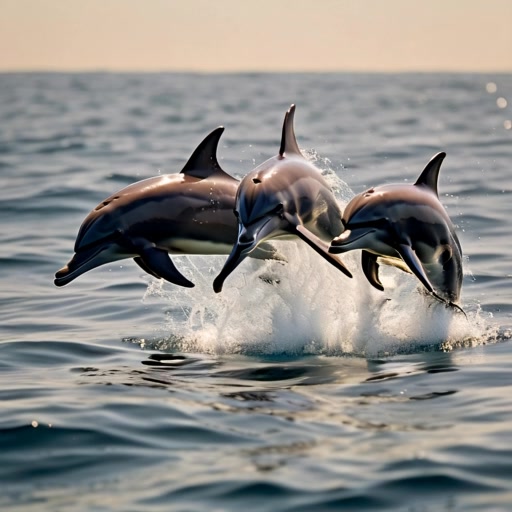} &
        \includegraphics[width=0.124\linewidth]{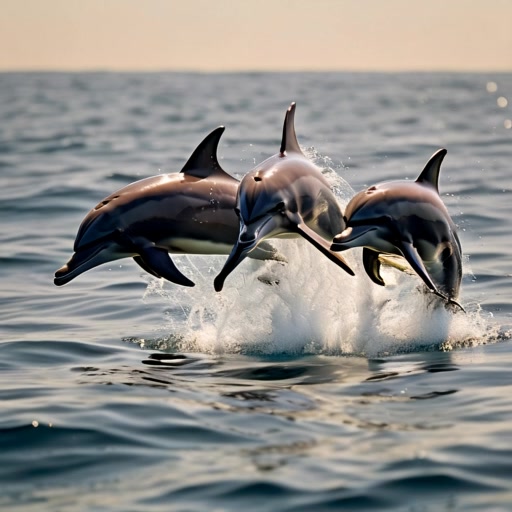} &
        \includegraphics[width=0.124\linewidth]{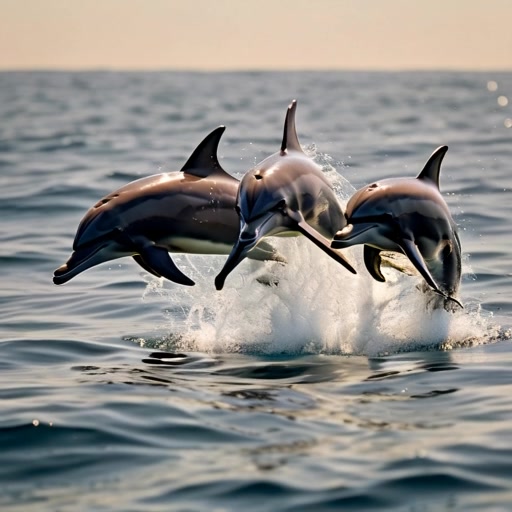} &
        \includegraphics[width=0.124\linewidth]{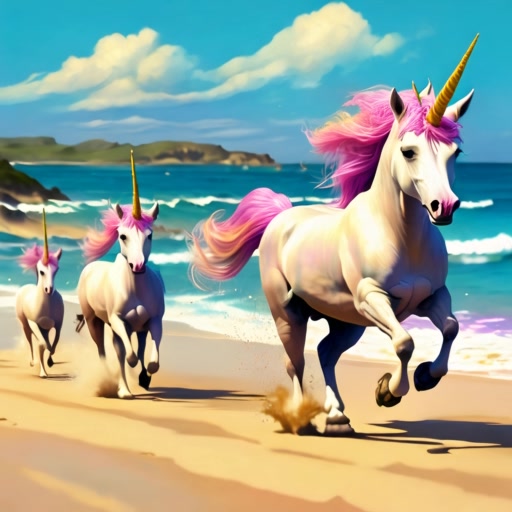} &
        \includegraphics[width=0.124\linewidth]{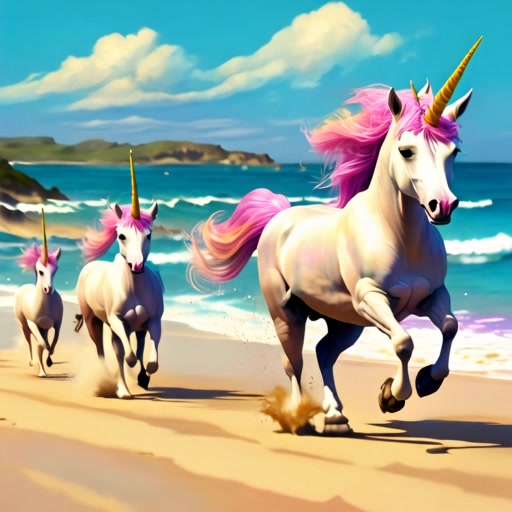} &
        \includegraphics[width=0.124\linewidth]{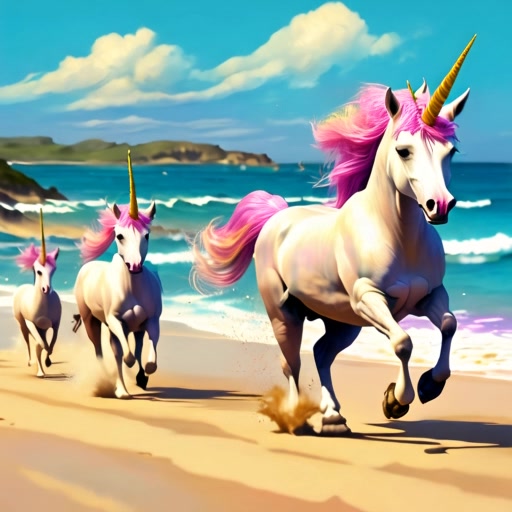} &
        \includegraphics[width=0.124\linewidth]{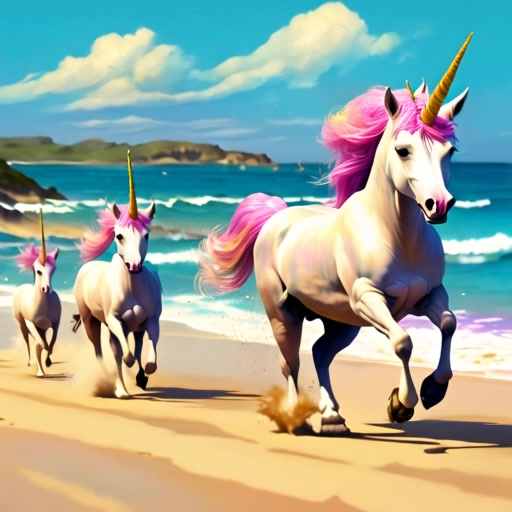} \\
        \rotatebox[origin=l]{90}{\scriptsize w/o 0-SNR} &
        \includegraphics[width=0.124\linewidth]{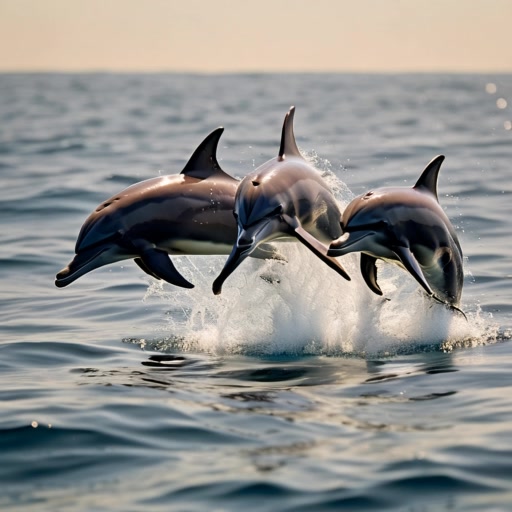} &
        \includegraphics[width=0.124\linewidth]{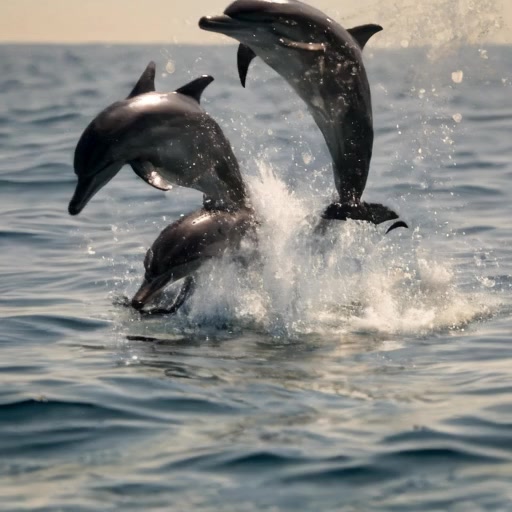} &
        \includegraphics[width=0.124\linewidth]{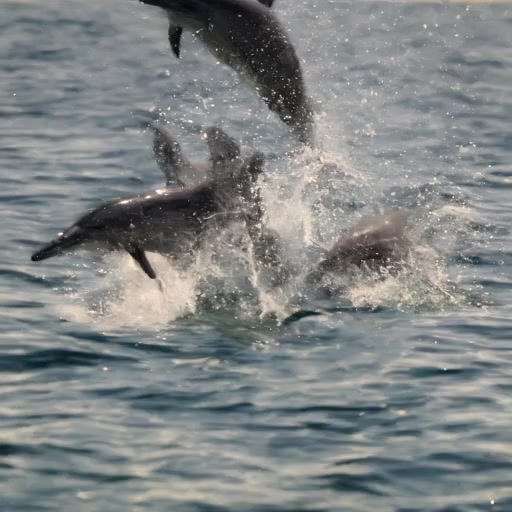} &
        \includegraphics[width=0.124\linewidth]{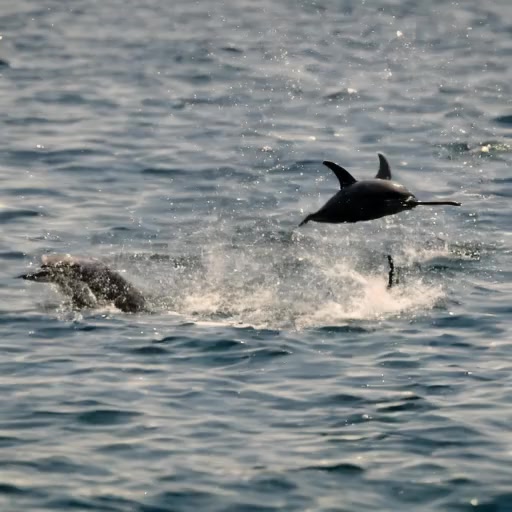} &
        \includegraphics[width=0.124\linewidth]{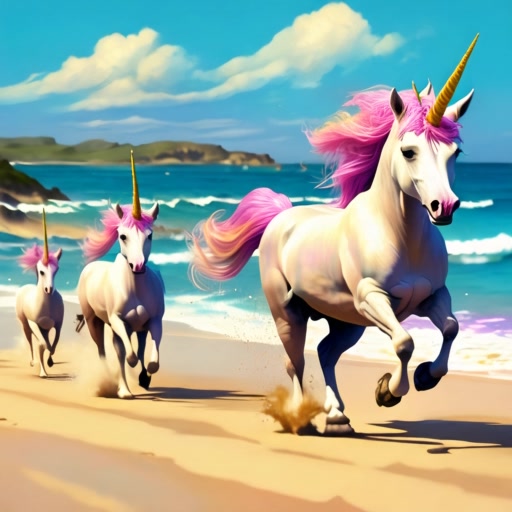} &
        \includegraphics[width=0.124\linewidth]{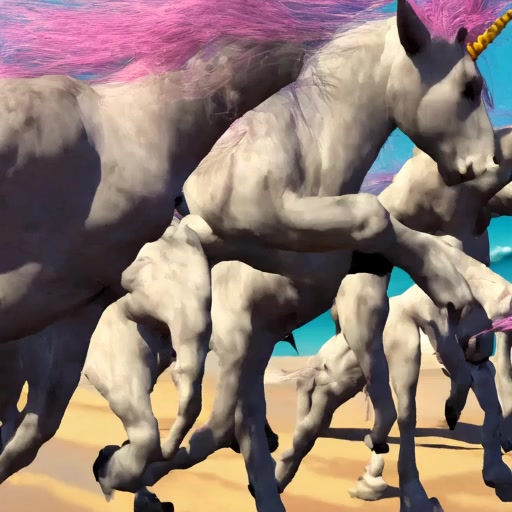} &
        \includegraphics[width=0.124\linewidth]{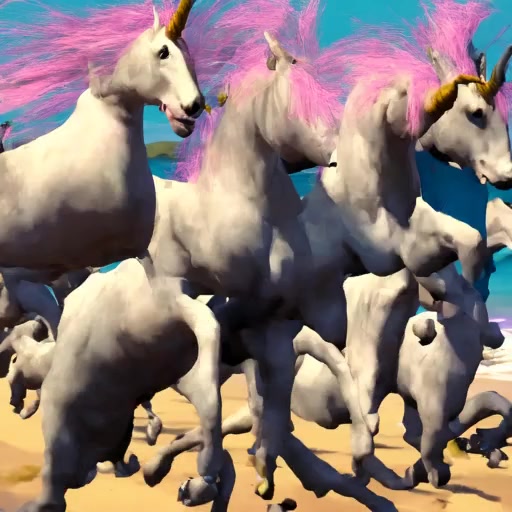} &
        \includegraphics[width=0.124\linewidth]{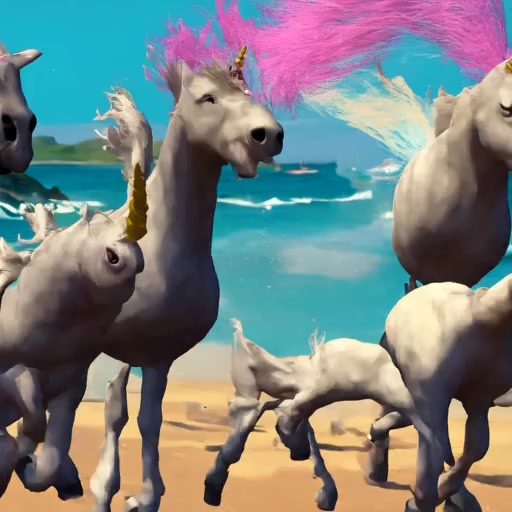} \\
        \rotatebox[origin=l]{90}{\scriptsize HQ FT} &
        \includegraphics[width=0.124\linewidth]{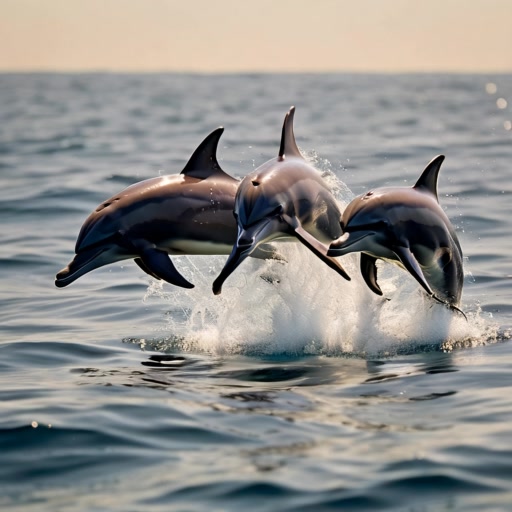} &
        \includegraphics[width=0.124\linewidth]{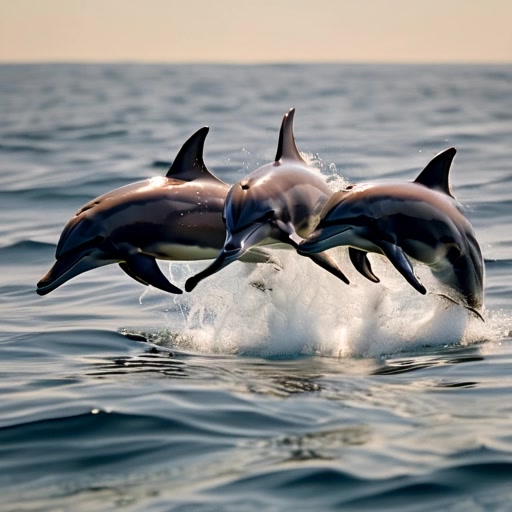} &
        \includegraphics[width=0.124\linewidth]{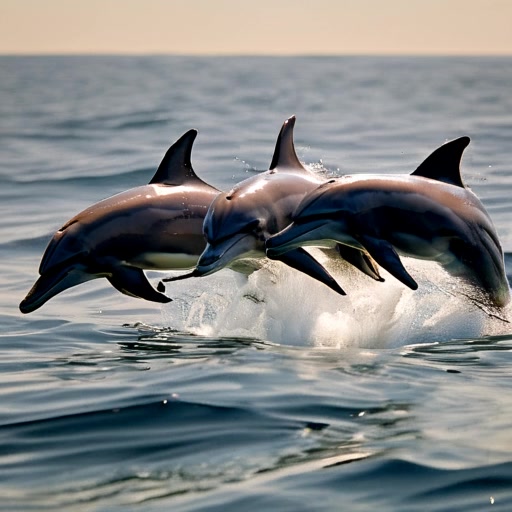} &
        \includegraphics[width=0.124\linewidth]{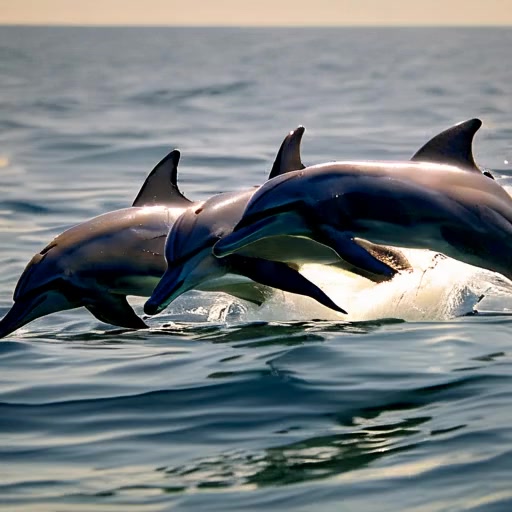} &
        \includegraphics[width=0.124\linewidth]{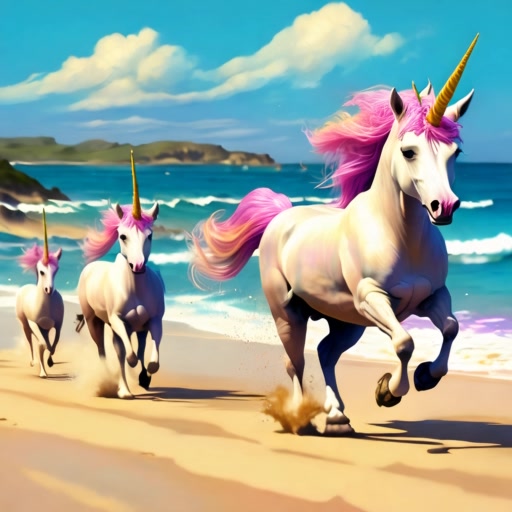} &
        \includegraphics[width=0.124\linewidth]{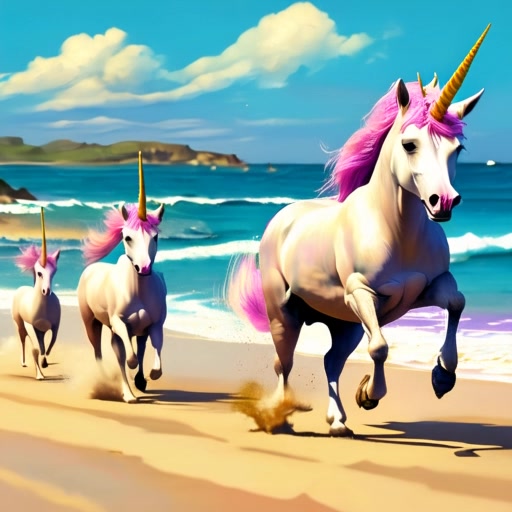} &
        \includegraphics[width=0.124\linewidth]{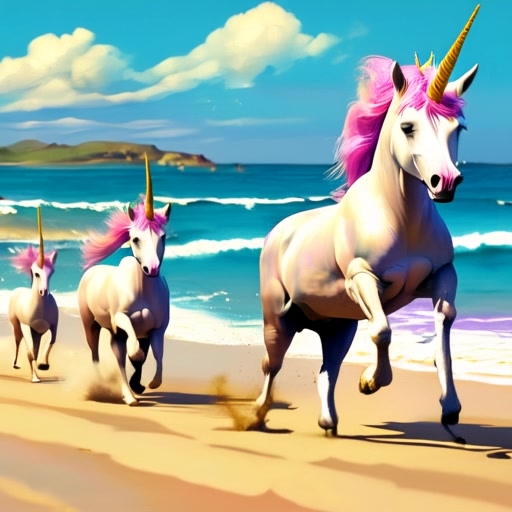} &
        \includegraphics[width=0.124\linewidth]{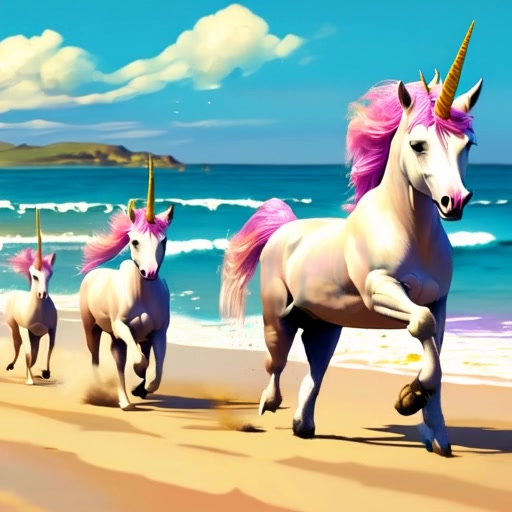} \\
    \end{tabular}%
}

%% file: sections/comparison_vs_prior.tex
\subsection{Comparison to prior work}
\label{sec:vs_prior_work}
\input{figures/comparison_prior_work.tex}

We evaluate \OURS against prior work and train $\extrapolateModel$ to produce $16$ frame $4$ second long videos and use the best design decisions from~\cref{sec:ablations}, including high quality finetuning.
We use the interpolation model $\interpolateModel$ on our generations to get $16$fps videos.
Please see~\cref{appendix:implementation_details} for details on how we interpolate 16-frame videos.

\par \noindent \textbf{Human evaluation of \textToV generation.}
Since many recent prior methods in \textToV generation are closed source~\cite{ho2022imagen,Blattmann2023AlignYL,ge2023preserve,gu2023reuse}, we use the publicly released examples from each of these methods.
Note that the released videos per method are likely to be the `best' representative samples from each method and may not capture their failure modes.
For \mav, we obtained non cherry-picked generations through personal communication with the authors.
For CogVideo~\cite{hong2022cogvideo}, we perform \textToVShort generation on the prompt set from~\cite{Blattmann2023AlignYL} using the open source models.
We also benchmark against commercially engineered black-box \textToV solutions, \gen~\cite{gen} and \pika~\cite{pikalabs}, obtaining generations through their respective websites using the prompts from~\cite{Blattmann2023AlignYL}.
We do not cherry-pick or contrastively rerank~\cite{ramesh2021zeroshot,yu2022scaling} our videos, and generate them using a deterministic random noise seed that is not optimized in any way.

Since each method generates videos at different resolutions, aspect-ratios, and frame-rates, we reduce annotator bias in human evaluations by postprocessing the videos for each comparison in~\cref{fig:teaser_compare} so that they match in these aspects.
Full details on this postprocessing and the text prompts used are in~\cref{appendix:comparisons_to_prior_work}.
As shown in~\cref{fig:teaser_compare}, \OURS's generations significantly outperform all prior work, including commercial solutions, both in terms of \quality (by an average of $91.8\%$) and \faithfulness (by an average of $86.6\%$).
We show some qualitative comparisons in~\cref{fig:qual_prior_work} and some additional generations in~\cref{fig:teaser}.
\OURS generates videos with significantly higher quality, and overall faithfulness to both the objects and motion specified in the text.
Since our factorized approach explicitly generates an image, we retain the visual diversity and styles of the \textToIShort model, leading to far better videos on fantastical and stylized prompts.
Additionally, \OURS generates videos with far greater temporal consistency than prior work.
We hypothesize that since we use stronger conditioning of image and text, our model is trained with a relatively easier task of predicting how an image evolves into the future, and thus is better able to model the temporal nature of videos.
Please see~\cref{appendix:qual_comparisons_to_prior_work} for more qualitative comparisons.
We include human evaluations where videos are not post-processed in the Appendix~\cref{appendix:comparisons_to_prior_work}, where again \OURS's generations significantly outperform all prior work.
The closest model in performance compared to ours is \imagenvideo when measured on \faithfulness, where we outperform \imagenvideo by $56\%$.
\imagenvideo's released prompts ask for generating text characters, a known failure mode~\cite{rombach2021highresolution,dai2023emu} of latent diffusion models used in \OURS.

\input{figures/juice_ours_vs_prior.tex}

We inspect the reasons that human evaluators prefer \OURS generations over the two strongest competitors in~\cref{fig:juice_ours_vs_prior}. A more detailed inspection is provided in~\cref{appendix:human_eval}.
\OURS generations are preferred due to their better pixel sharpness and motion smoothness.
While being state-of-the-art, \OURS is also simpler and has a two model cascade with a total of $6.0$B parameters ($2.7$B frozen parameters for spatial layers, and $1.7$B learnable temporal parameters each for $\extrapolateModel$ and $\interpolateModel$), which is much simpler than methods like \imagenvideo ($7$ model cascade, $11.6$B parameters), \mav ($5$ model cascade, $9.6$B parameters) trained using similar scale of data.

\input{tables/sota_ucf.tex}

\par \noindent \textbf{Automated metrics.}
In~\cref{tab:sota_ucf}, we compare against prior work using the zero-shot \textToVShort generation setting from~\cite{singer2023makeavideo} on the \ucf dataset~\cite{soomro2012ucf101}.
\OURS achieves a comptetitive IS score~\cite{salimans2016improved} and a lower FVD~\cite{unterthiner2019fvd}.
To confirm these automated scores, we also use human evaluations to compare our generations to \mav.
We use a subset of $303$ generated videos ($3$ random samples per \ucf class) and find that our generations are strongly preferred (\cref{tab:sota_ucf} right).
Qualitative comparisons can be found in~\cref{appendix:qual_comparisons_to_prior_work}.

\input{tables/i2v_quant_eval.tex}

\par \noindent \textbf{Animating images.}
A benefit of our factorized generation is that the same model can be used out-of-the-box to `animate' user-provided images by supplying them as the conditioning image $\bi$.
We compare \OURS's image animation with six methods, prior and concurrent work~\cite{chen2023videocrafter1,2023videocomposer, sdvideo, i2vgen} and commercial \imageToV (\imageToVShort) solutions~\cite{pikalabs,gen}, on the prompts from~\cite{singer2023makeavideo} and~\cite{Blattmann2023AlignYL}.
All the methods are shown the same image generated using a different \textToI model~\cite{podell2023sdxl} and expected to generate a video according to the text prompt\footnote{Due to lack of access to training data of SDXL~\cite{podell2023sdxl} and their underlying model, we leveraged their corresponding APIs for our comparison.}.
We report human evaluations in~\cref{tab:i2v_human_eval} and automated metrics in the Appendix~\cref{tab:i2v_automatic_metrics}.
Human evaluators strongly prefer \Ours's generations across all the baselines.
These results demonstrate the superior image animation capabilities of \Ours compared to methods specifically designed for the \imageToV task.%

%% file: figures/comparison_prior_work.tex
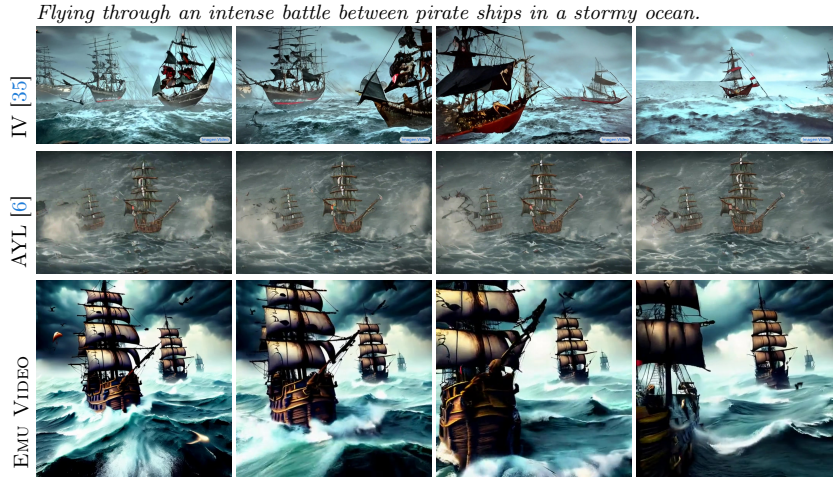
\begin{figure}
    \centering
    \input{figures/sota/figure.tex}
    \caption{
    \textbf{Qualitative comparison.}
    \OURS produces higher quality generations compared to \imagenvideo~\cite{ho2022imagen} and \ayol~\cite{Blattmann2023AlignYL} in terms of style and consistency.
    }
    \label{fig:qual_prior_work}
\end{figure}

%% file: figures/sota/figure.tex
\setlength{\tabcolsep}{1pt}
\resizebox{0.9\linewidth}{!}{%
    \begin{tabular}{ccccc}
        & \multicolumn{4}{l}{\footnotesize \it Flying through an intense battle between pirate ships in a stormy ocean.} \\
        \rotatebox[origin=l]{90}{\footnotesize IV~\cite{ho2022imagen}} &
        \includegraphics[width=0.245\linewidth]{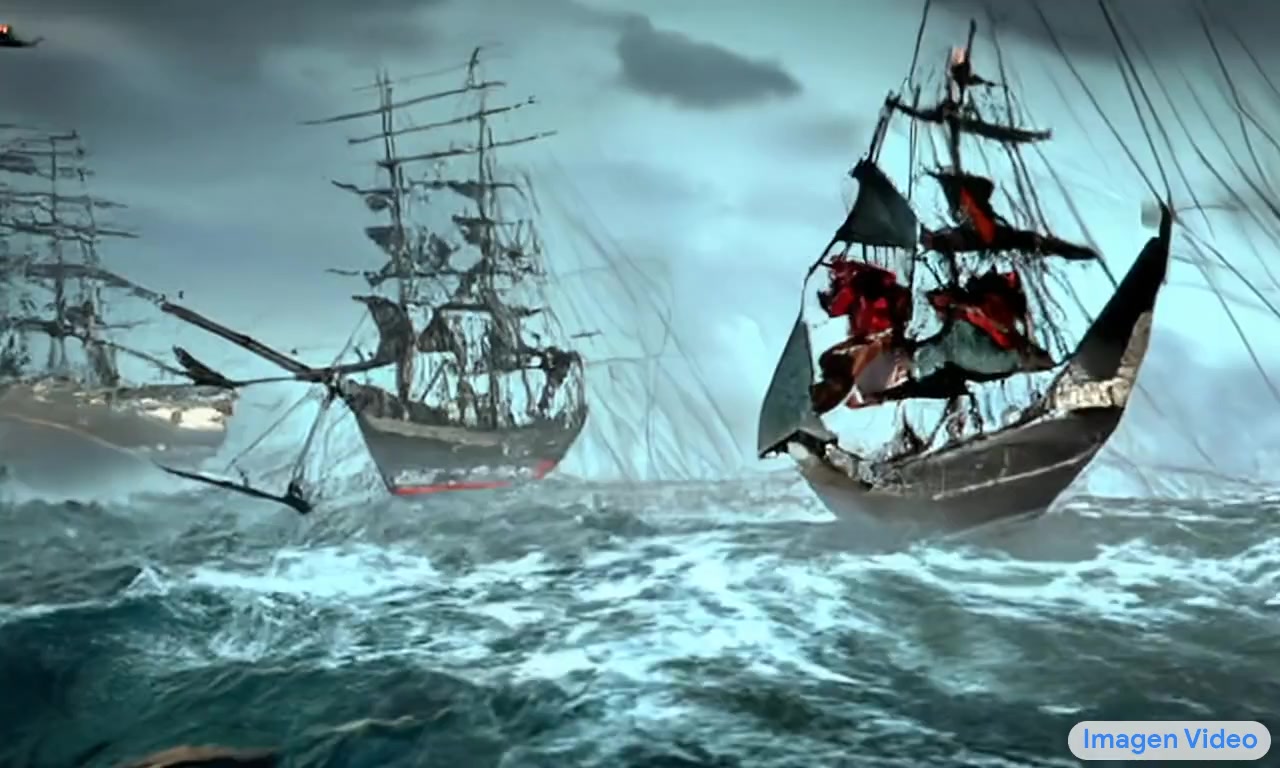} &
        \includegraphics[width=0.245\linewidth]{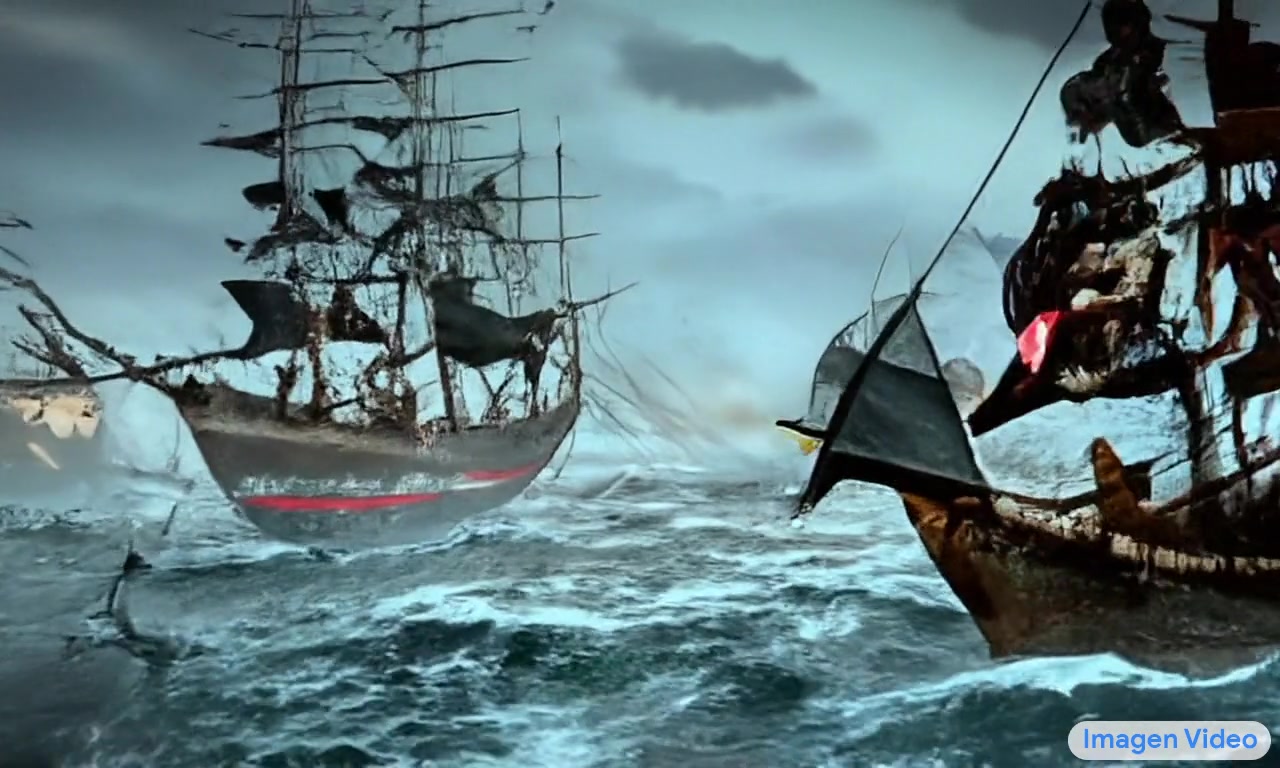} &
        \includegraphics[width=0.245\linewidth]{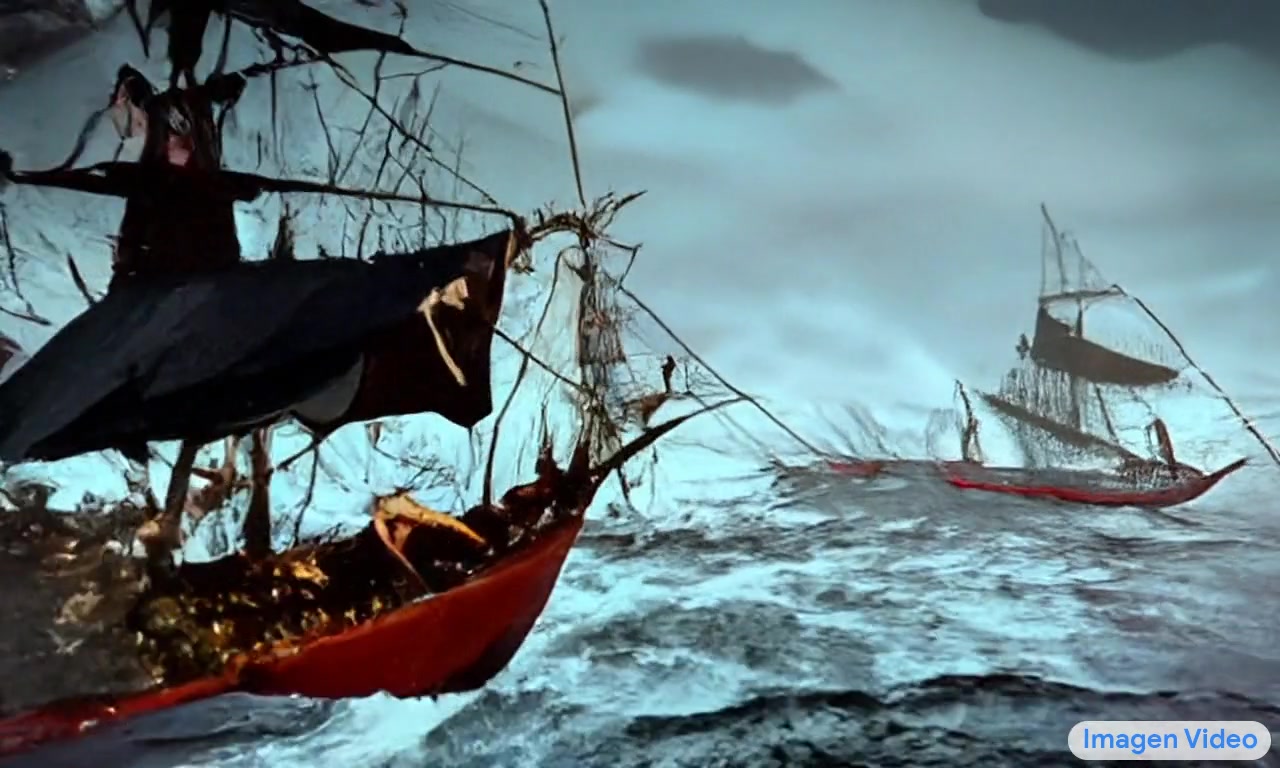} &
        \includegraphics[width=0.245\linewidth]{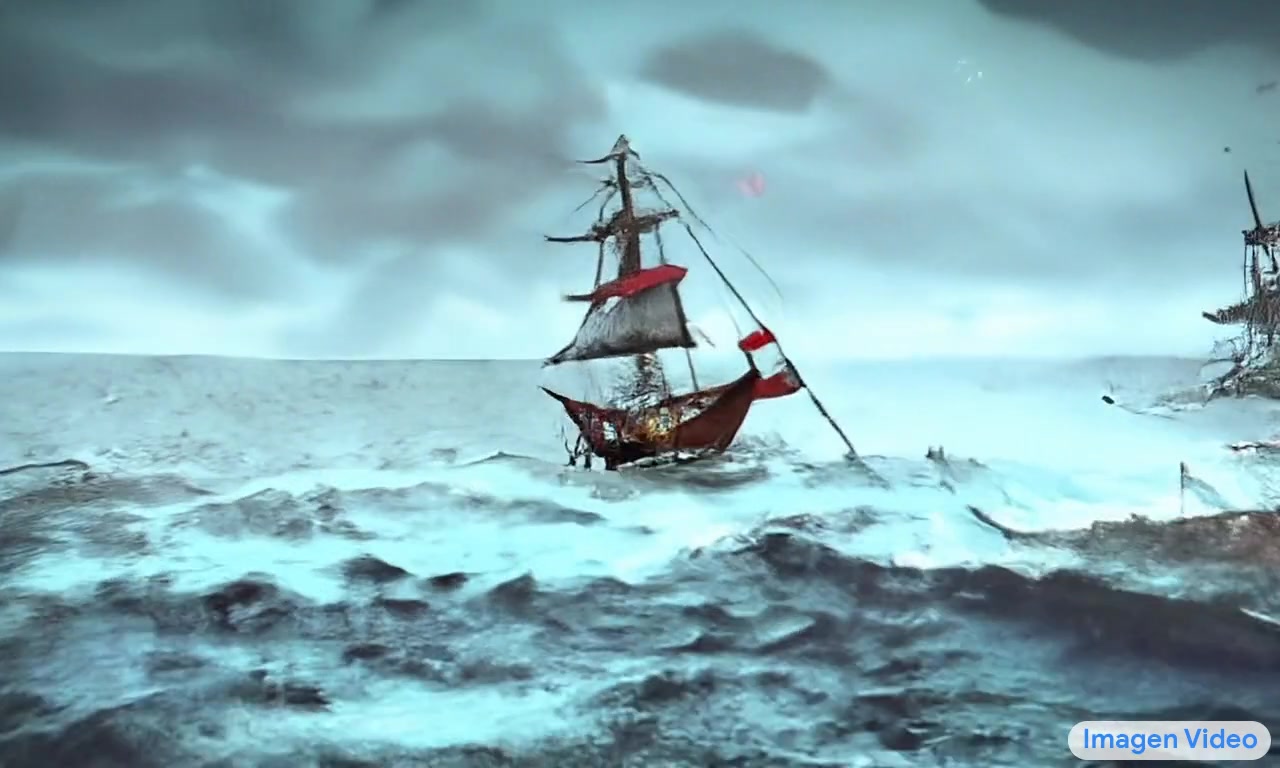} \\
        \rotatebox[origin=l]{90}{\footnotesize AYL~\cite{Blattmann2023AlignYL}} &
        \includegraphics[width=0.245\linewidth]{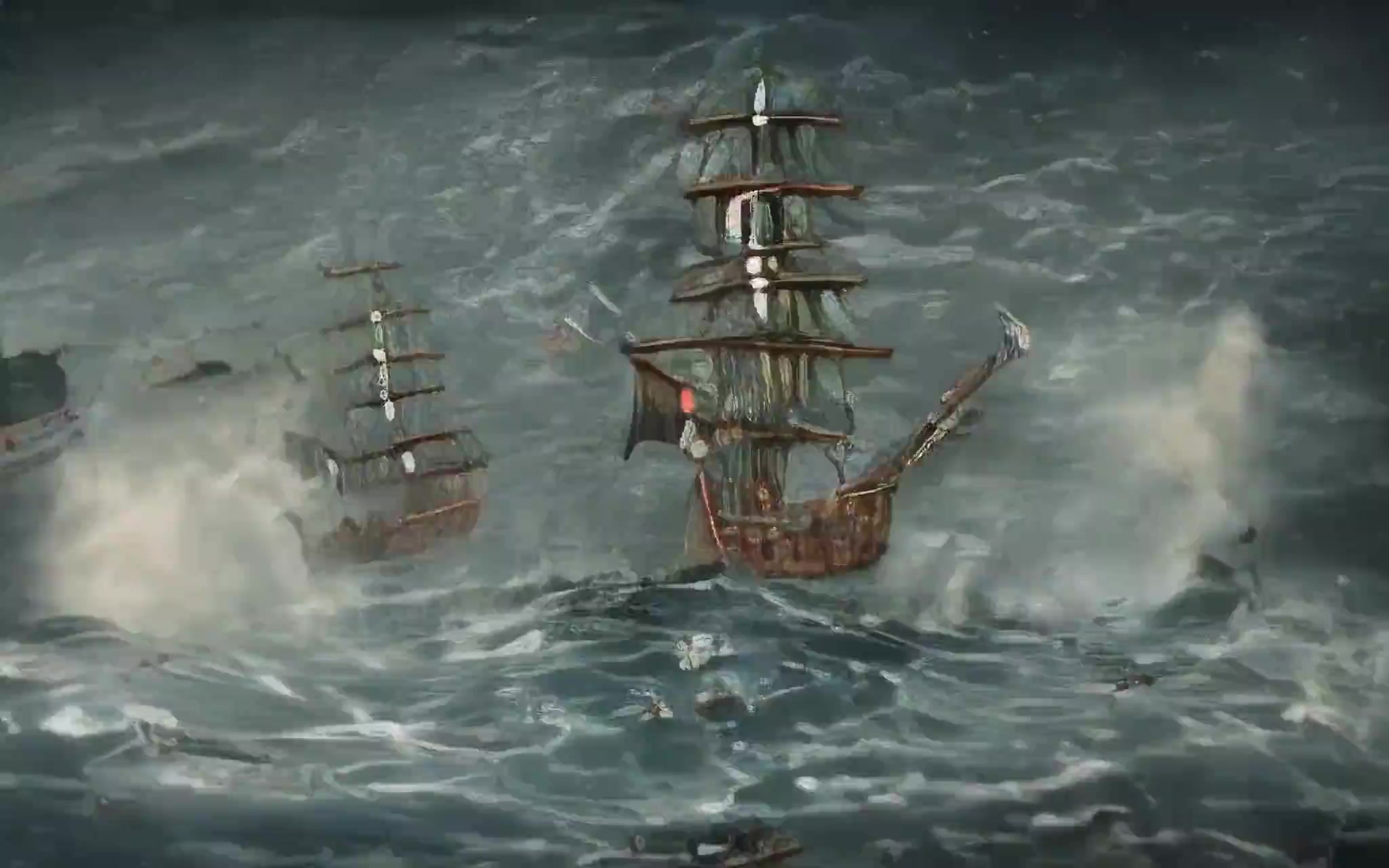} &
        \includegraphics[width=0.245\linewidth]{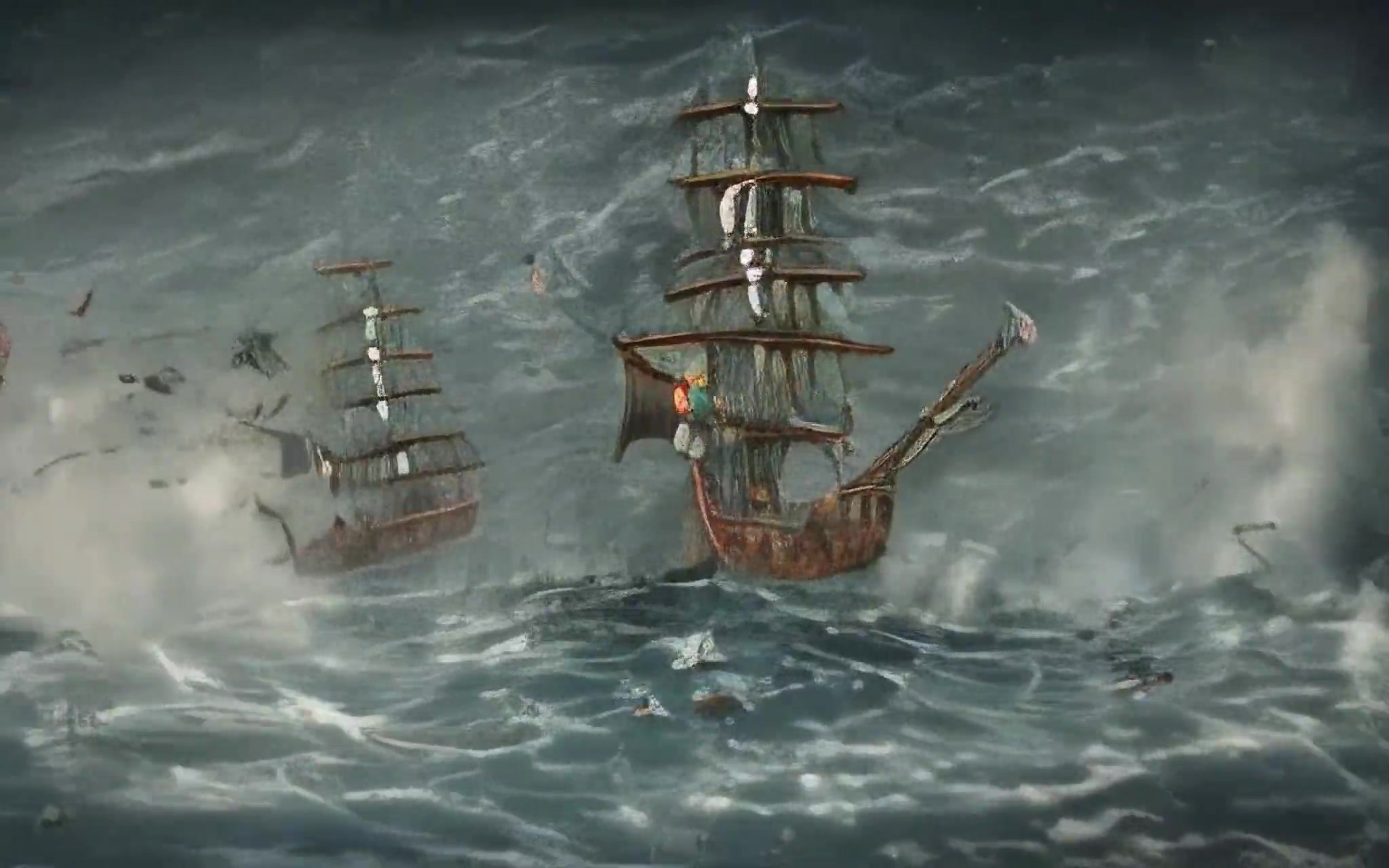} &
        \includegraphics[width=0.245\linewidth]{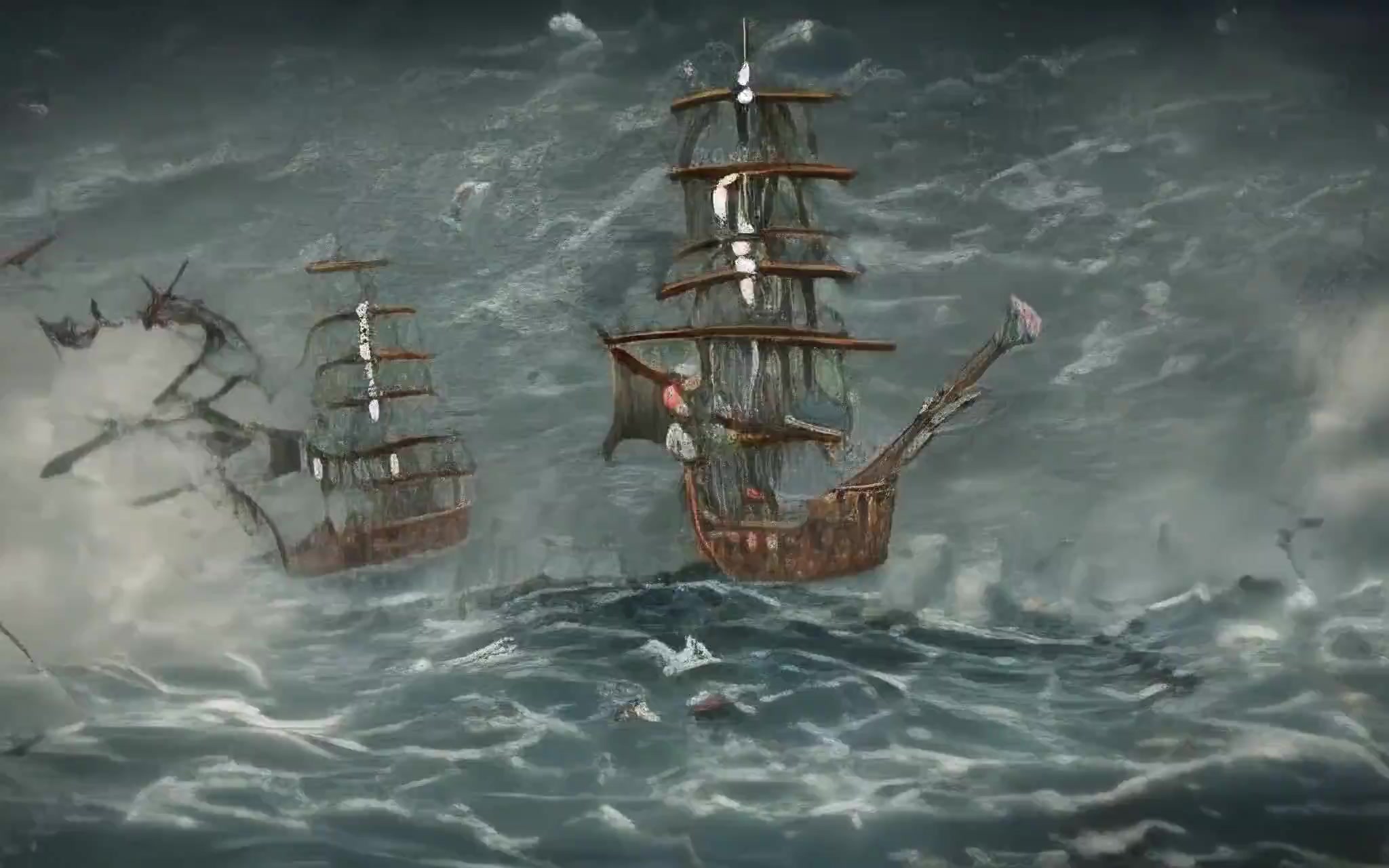} &
        \includegraphics[width=0.245\linewidth]{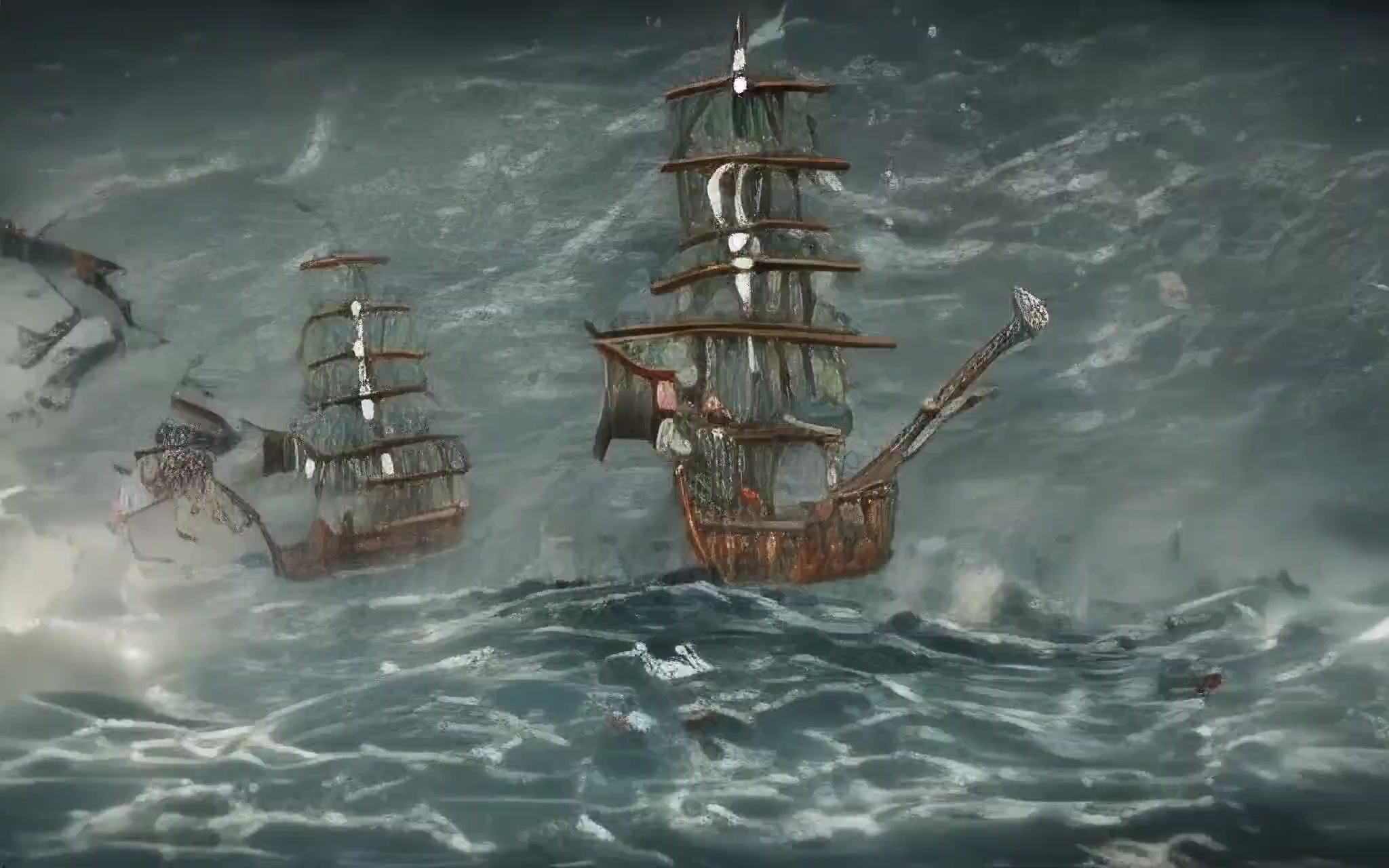} \\
        \rotatebox[origin=l]{90}{\footnotesize \OURS} &
        \includegraphics[width=0.245\linewidth]{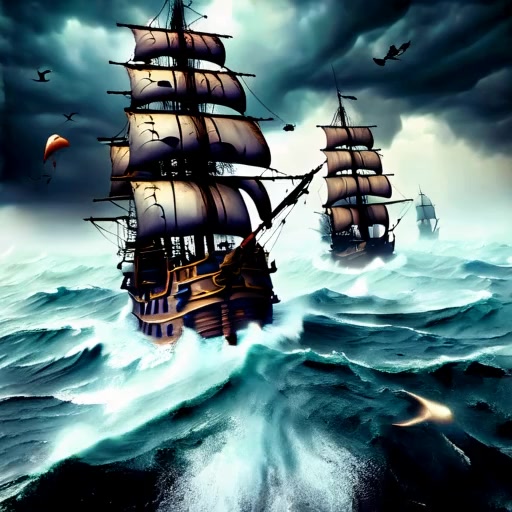} &
        \includegraphics[width=0.245\linewidth]{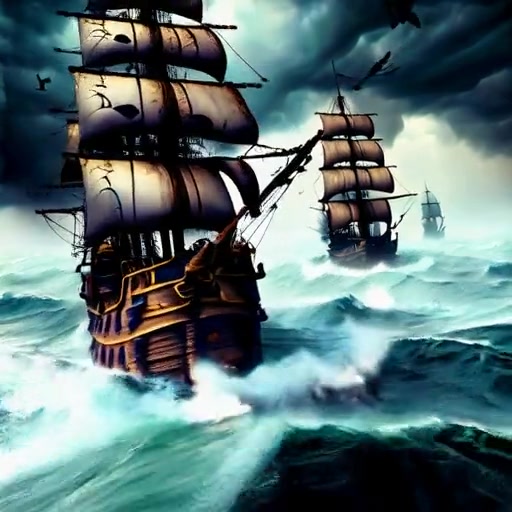} &
        \includegraphics[width=0.245\linewidth]{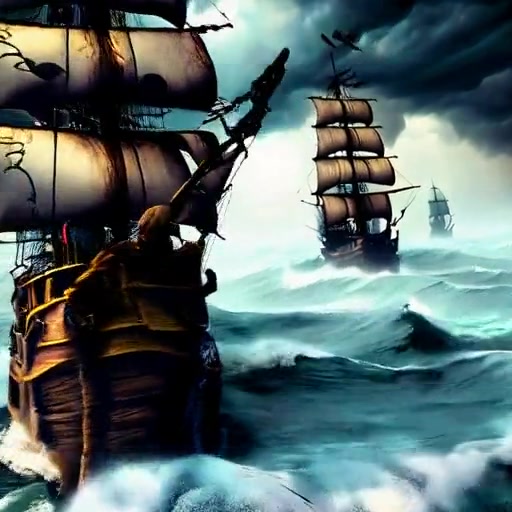} &
        \includegraphics[width=0.245\linewidth]{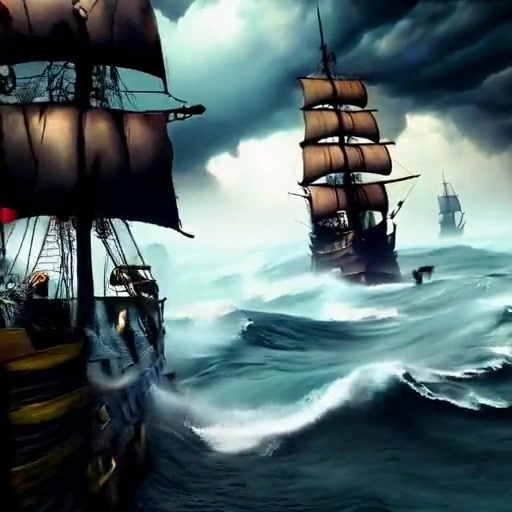} \\
    \end{tabular}%
}

%% file: figures/juice_ours_vs_prior.tex
\begin{figure}[!t]
    \centering
    \input{figures/juice.tex}
    \caption{\textbf{Percentage of each reason selected for samples where \OURS wins against \mav~\cite{singer2023makeavideo} or \imagenvideo~\cite{ho2022imagen} on \quality}.
    Human raters pick \OURS primarily due to their pixel sharpness and motion smoothness, with an overall preference of $96.8\%$ and $81.8\%$ to each baseline, respectively.
    }
    \label{fig:juice_ours_vs_prior}
\end{figure}
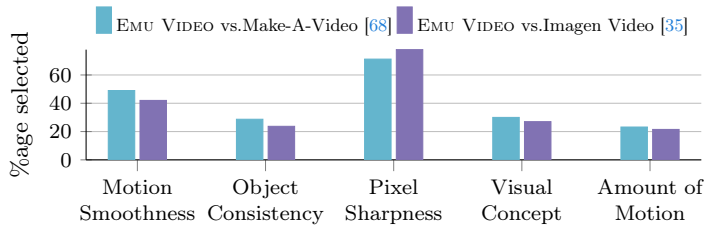

%% file: figures/juice.tex
\begin{tikzpicture}
    \begin{axis}
    [
    ybar,
    legend columns=2,
    legend style={at={(1.0,1.25)},anchor=east,fill=none,draw=none, legend columns=2,style={nodes={scale=0.75, transform shape}}},
    ylabel={\%age selected},
    legend style={font=\footnotesize},
    label style={font=\footnotesize},
    tick label style={font=\fontsize{8}{9.6}\selectfont},
    axis x line*=bottom,
    axis y line=left,
    y axis line style={-},
    ymajorgrids=true,
    ymin=0,
    symbolic x coords={ms, oc, ps, vc, am},
    xticklabels={None, Motion\\Smoothness, Object\\Consistency, Pixel\\Sharpness, Visual\\Concept, Amount of\\Motion},
    xticklabel style={align=center},
    table/col sep=semicolon,
    label style={font=\footnotesize},
    height=1.2in,
    width=0.8\linewidth,
    legend image code/.code={
        \draw [#1] (0cm,-0.1cm) rectangle (0.2cm,0.25cm); },
    ]

    \addplot [snsTealColor,fill=snsTealColor] coordinates {(ms, 49.0) (oc, 28.7) (ps, 71.2) (vc, 30.0) (am, 23.2)};
    \addlegendentry{\OURS vs.\mav~\cite{singer2023makeavideo}}
    \addplot [snsVioletColor,fill=snsVioletColor] coordinates {(ms, 42.0) (oc, 23.7) (ps, 77.9) (vc, 27.0) (am, 21.5)};
    \addlegendentry{\OURS vs.\imagenvideo~\cite{ho2022imagen}}
    \end{axis}
\end{tikzpicture}

%% file: tables/sota_ucf.tex
\vspace{-7mm}
\begin{figure}
    \centering
    \begin{minipage}[t]{.4\linewidth}
        \vspace{0pt}
        \resizebox{\linewidth}{!}{
        \setlength{\tabcolsep}{2pt}
        \begin{tabular}{l | cc}
            \bf \multirow{2}{*}{\bf Method} & \multicolumn{2}{c}{\bf Automated}\\
            & \bf FVD $\downarrow$ & \bf IS $\uparrow$ \\
            \midrule
            MagicVideo~\cite{zhou2023magicvideo} & $655.0$ & - \\
            Align Your Latents~\cite{Blattmann2023AlignYL} & $550.6$ & $33.5$ \\
            \mav~\cite{singer2023makeavideo} & $367.2$ & $33.0$\\
            \pyoco~\cite{ge2023preserve} & $355.2$ & $47.8$\\
            \midrule
            \Ours & $317.1$ & $42.7$\\
        \end{tabular}
    }
    \end{minipage}%
    \hspace{10pt}
    \begin{minipage}[t]{.4\linewidth}
        \vspace{0pt}
    \input{tikz_graph_sources/ucf_human_eval.tex}
    \end{minipage}%
    \captionof{table}{\textbf{Automated metrics for zero-shot \textToV evaluation on \ucf.}
    (Left) We present automated metrics and observe that \OURS achieves competitive IS and outperforms all prior work on FVD.
    (Right) We conduct human evaluations to compare \OURS and \mav where \OURS significantly outperforms \mav both in \quality ($90.1\%$) and \faithfulness ($80.5\%$).
    }
    \label{tab:sota_ucf}
    \vspace*{-0.5cm}
\end{figure}

%% file: tikz_graph_sources/ucf_human_eval.tex
\begin{tikzpicture}
    \begin{axis}
    [
    ybar,
    ylabel={\% Win Rate},
    label style={font=\scriptsize},
    tick label style={font=\scriptsize},
    axis x line*=bottom,
    axis y line=left,
    ymajorgrids=true,
    ymin=0,
    ytick={0,25,50,75,100},
    symbolic x coords={Q, F},
    xtick={Q, F},
    xticklabel style={align=center},
    xtick distance=0.1,
    table/col sep=semicolon,
    label style={font=\scriptsize},
    bar shift=0pt,
    height=1.1in,
    width=\linewidth,
    title style={font=\scriptsize, align=center},
    title={\bf Human Evaluation\\\vs \mav},
    enlarge x limits=0.8,
    bar width=10pt,
    ]
    \addplot [snsTealColor,fill=snsGreenColor] coordinates {(Q, 90.1)};
    \addplot [snsVioletColor,fill=snsBlueColor] coordinates {(F, 80.5)};
    \end{axis}
\end{tikzpicture}

%% file: tables/i2v_quant_eval.tex
\begin{table}
    \centering
    \resizebox{0.7\linewidth}{!}{
    \input{tabulars/i2v_human_eval.tex}
    }
    \caption{\textbf{Human evaluation of \Ours \vs prior$^*$ and concurrent$^{**}$ work in text-conditioned image animation.} 
    We compare \Ours against six methods across two prompt sets using generations from~\cite{podell2023sdxl} as the starting images.
    \OURS's animated videos are strongly preferred over all baselines. 
    }
    \label{tab:i2v_human_eval}
   \vspace*{-0.5cm}
\end{table}

%% file: tabulars/i2v_human_eval.tex
\begin{tabular}{cc | cc}
     \bf Method & \#Prompts &  \bf \qualityShort & \bf \faithfulnessShort \\
    \hline
     \OURS \vs \vidcomp \imageToVShort$^*$~\cite{2023videocomposer} & \multirow{6}{*}{$65$ \cite{Blattmann2023AlignYL}}  & $96.9$  & $96.9$ \\
     \OURS \vs \pika \imageToVShort$^*$~\cite{pikalabs} & & $84.6$ & $84.6$ \\
     \OURS \vs \gen \imageToVShort$^*$~\cite{gen} & & $70.8$ & $76.9$ \\
     \OURS \vs \videocrafter \imageToVShort$^*$~\cite{chen2023videocrafter1} & & $81.5$ & $80.0$ \\
     \OURS \vs \sdvideo \imageToVShort$^{**}$~\cite{sdvideo} &  & $72.3$ & $73.9$ \\
     \OURS \vs \itovgen \imageToVShort$^{**}$~\cite{i2vgen} & & $69.2$ & $66.1$ \\
    \hline
    \OURS \vs \vidcomp \imageToVShort$^*$~\cite{2023videocomposer} & $307$ \cite{singer2023makeavideo}& $97.4$  & $91.2$  \\
\end{tabular}%

%% file: sections/applications.tex
\subsection{Analysis}

\par \noindent \textbf{Nearest neighbor baseline.}
We expect good and useful generative models to outperform a nearest neighbor retrieval baseline and create videos not in the training set.
We construct a strong nearest neighbor baseline that retrieves videos from the full training set ($34$M videos) by using the text prompt's CLIP feature similarity to the training prompts.
When using the evaluation prompts from~\cite{singer2023makeavideo}, human evaluators prefer \OURS's generations $81.1\%$ in \faithfulness over real videos confirming that \OURS outperforms the strong retrieval baseline.
We manually inspected and confirmed that \OURS outperforms the baseline for prompts not in the training set.

\par \noindent \textbf{Extending video length with longer text.}
Recall that our model conditions on the text prompt and a starting frame to generate a video.
With a small architectural modification, we can also condition the model on $\vTime$ frames and \textit{extend} the video.
Thus, we train a variant of \Ours to generate the future $16$ frames conditioned on the `past' $16$ frames.
While extending the video, we use a \textit{future} text prompt different from the one used for the original video and visualize results in~\cref{fig:longer_video}.
We find that the extended videos respect the original video as well as the future text prompt.
\input{figures/longer_video_w_longer_text.tex}

%% file: figures/longer_video_w_longer_text.tex
\begin{figure}[t]
    \centering
    \input{figures/2x_extraoplate_v1_beer/figure.tex}
    \caption{\textbf{Extending to longer videos.}
    We test a variant of \OURS that is conditioned on all the frames from the original video, and generates new videos conditioned on a future prompt.
    For two different future prompts, our model generates plausible extended videos that respect the original video and the future text.
    }
    \label{fig:longer_video}
    \vspace*{-0.2cm}
\end{figure}
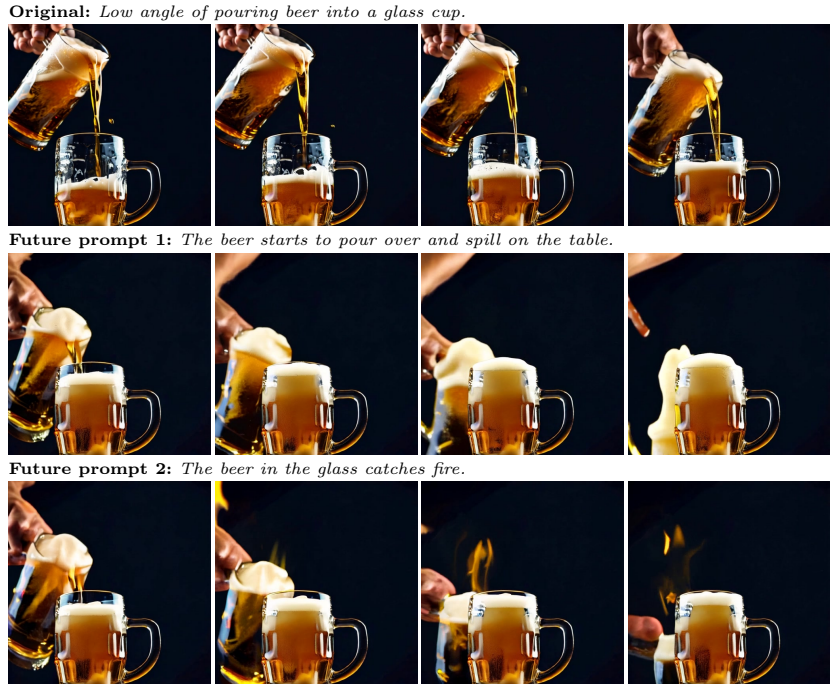

%% file: figures/2x_extraoplate_v1_beer/figure.tex
\setlength{\tabcolsep}{1pt}
\setlength\extrarowheight{-4pt}
\resizebox{0.9\linewidth}{!}{%
    \begin{tabular}{cccc}
        \multicolumn{4}{l}{\scriptsize \it {\bf Original:} Low angle of pouring beer into a glass cup.} \\
        \includegraphics[width=0.245\linewidth]{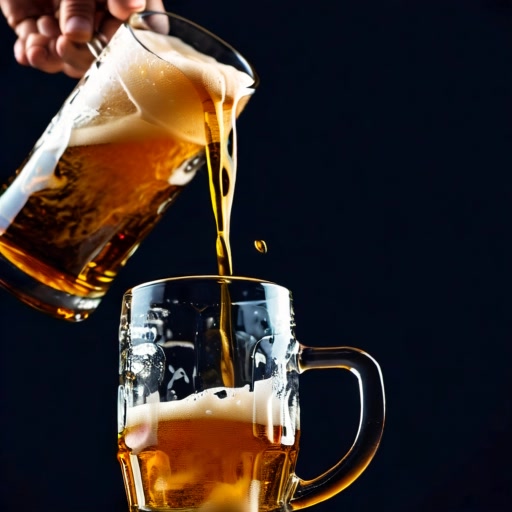} &
        \includegraphics[width=0.245\linewidth]{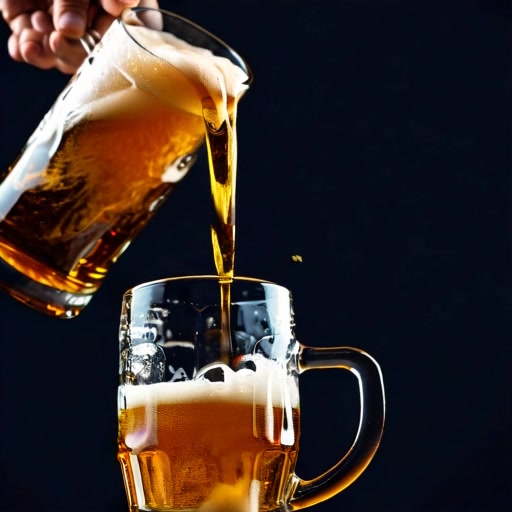} &
        \includegraphics[width=0.245\linewidth]{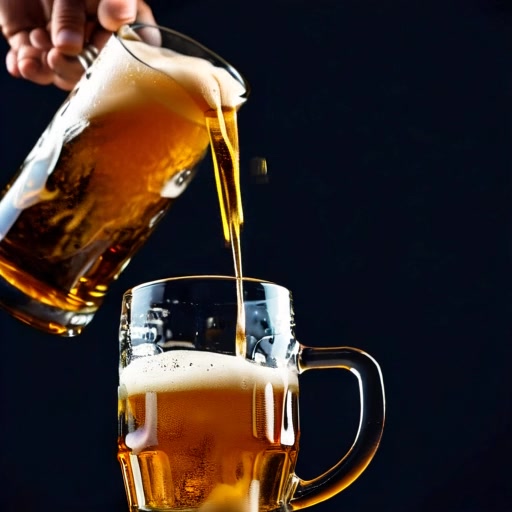} &
        \includegraphics[width=0.245\linewidth]{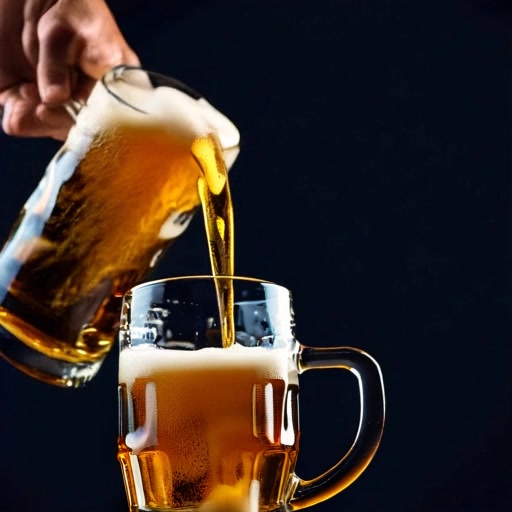} \\
        \multicolumn{4}{l}{\scriptsize \it {\bf Future prompt 1:}  The beer starts to pour over and spill on the table.} \\
        \includegraphics[width=0.245\linewidth]{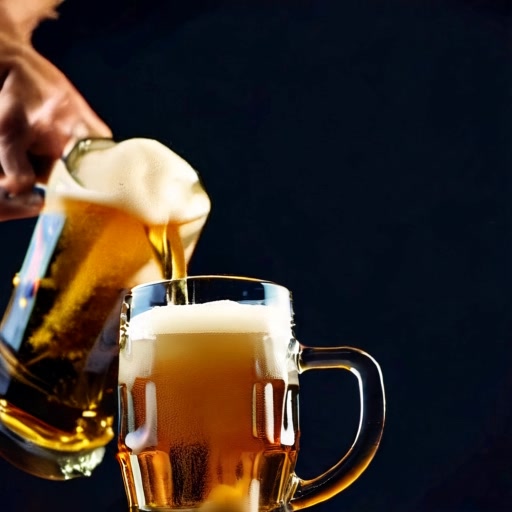} &
        \includegraphics[width=0.245\linewidth]{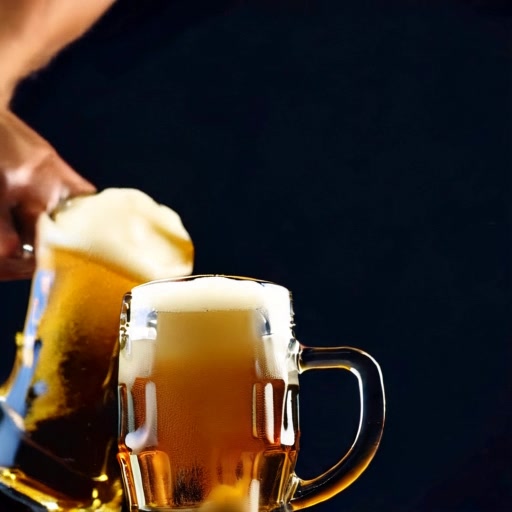} &
        \includegraphics[width=0.245\linewidth]{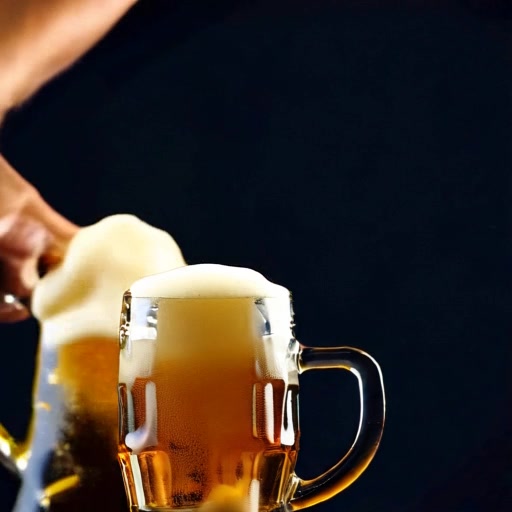} &
        \includegraphics[width=0.245\linewidth]{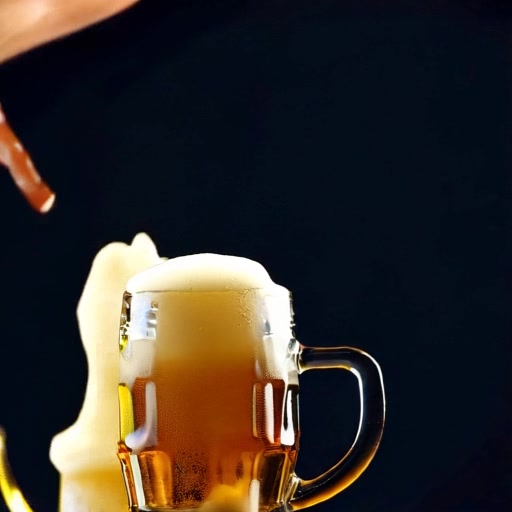} \\
        \multicolumn{4}{l}{\scriptsize \it {\bf Future prompt 2:} The beer in the glass catches fire.} \\
        \includegraphics[width=0.245\linewidth]{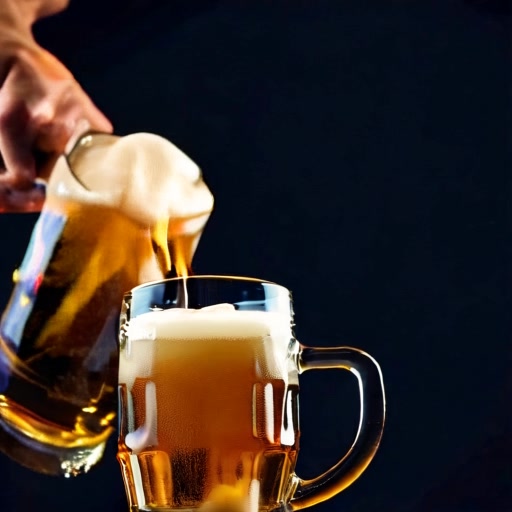} &
        \includegraphics[width=0.245\linewidth]{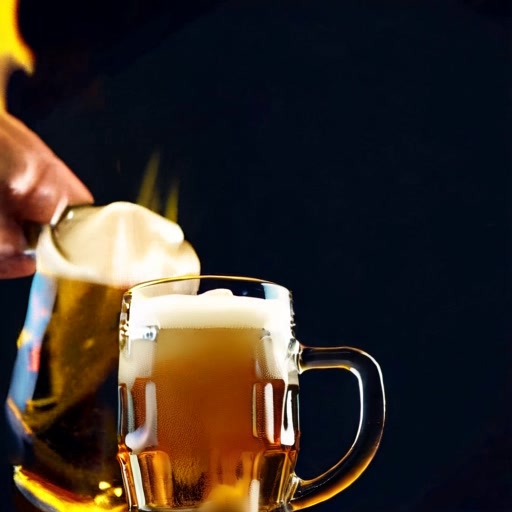} &
        \includegraphics[width=0.245\linewidth]{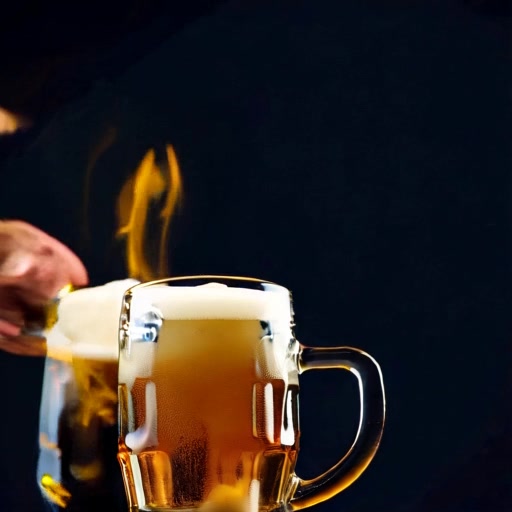} &
        \includegraphics[width=0.245\linewidth]{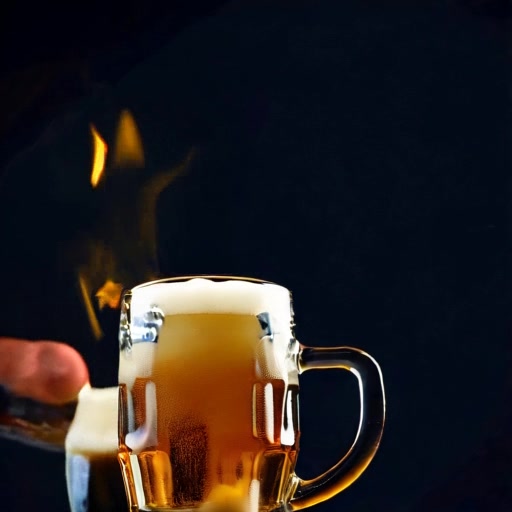} \\
    \end{tabular}%
}

%% file: sections/conclusion.tex
\section{Limitations and ethical considerations}
\label{sec:conclusion}

We presented \OURS, a factorized approach to \textToV generation that leverages strong image and text conditioning.
\OURS significantly outperforms all prior work including commercial solutions.
Although our model has been a step change in video generation and shares valuable insights into the modeling and evaluation challenges, there are limitations.
\OURS can be improved in the following aspects as future research directions:
the realism of the presented content, fine-grained details such as hand and face artifacts, modeling physics, and maintaining quality and consistency
for long video durations. These factors have been considered in the JUICE metric where the raters are asked to consider object/scene consistency and
pixel quality in their evaluations. Another direction for future research is to improve \OURS's ability to recover from conditioning
frames that are not representative of the prompt.
Strengthening the conditioning for video models using pure autoregressive decoding with diffusion models is not currently computationally attractive.
However, further research may provide benefits for longer video generation.

\vspace{-0.2cm}
\paragraph{\textbf{Ethical considerations.}}
We propose advancements in generative methods specifically to improve the generation of high dimensional video outputs.
Generative methods can be applied to a large variety of different usecases which are beyond the scope of this work.
A careful study of the data, model, its intended applications, safety, risk, bias, and societal impact is necessary before any real world application.

\newpage
\par \noindent \textbf{Acknowledgments.}
We are grateful for the support of multiple collaborators at Meta who helped us in this work.
Baixue Zheng, Baishan Guo, Jeremy Teboul, Milan Zhou, Shenghao Lin, Kunal Pradhan, Jort Gemmeke, Jacob Xu, Dingkang Wang, Samyak Datta, Guan Pang, Symon Perriman, Vivek Pai, Shubho Sengupta for their help with the data and infra.
We would like to thank Uriel Singer, Adam Polyak, Shelly Sheynin, Yaniv Taigman, Licheng Yu, Luxin Zhang, Yinan Zhao, David Yan, Yaqiao Luo, Xiaoliang Dai, Zijian He, Peizhao Zhang, Peter Vajda, Roshan Sumbaly, Armen Aghajanyan, Michael Rabbat, and Michal Drozdzal for helpful discussions.
We are also grateful to the help from Lauren Cohen, Mo Metanat, Lydia Baillergeau, Amanda Felix, Ana Paula Kirschner Mofarrej, Kelly Freed, Somya Jain.
We thank Ahmad Al-Dahle and Manohar Paluri for their support.

%% file: sections/supplement.tex
\section{Implementation Details}
\label{appendix:implementation_details}

In this section we include details on the architectures and hyper-parameters used for training the models in the main paper, and on the use of multiple conditionings for classifier-free guidance.
For both our text-to-video ($\extrapolateModel$) and interpolation ($\interpolateModel$)
models we train with the same U-Net architecture. We share the exact model configuration for our U-Net in \cref{tab:app_unet_arch},
and the configuration for our 8-channel autoencoder in \cref{tab:app_vae_arch}.

\input{tables/appendix_unet_arch}
\input{tables/appendix_vae_arch}

\cref{tab:app_training_settings} shares the training hyperparameters we used for various stages of our training -- 256px training,
512px training, High Quality finetuning, and frame interpolation. For inference, we use the DDIM sampler~\cite{song2020denoising}
with 250 diffusion steps. We use Classifier Free Guidance (CFG)~\cite{ho2022classifier} with $w_{img}$ of 7.5 for image generation, and
$w_{img}$ of 2.0 and $w_{txt}$ of 7.5 for both video generation and frame interpolation. We share more
details about handling multiple conditionings for Classifier Free Guidance next.

\input{tables/appendix_training_hyperparams}

\noindent \textbf{Multiple Conditionings for CFG.}
For video generation, our model receives two conditioning signals (image $\bi$, text prompt $\bp$), which we use in conjunction for Classifier Free Guidance~\cite{ho2022classifier}.
\cref{equation:app_cfg} lists the combined CFG equation we use.
\begin{equation}
    \begin{aligned}
        \tilde{\bx} = \bx + w_i (\bx(\bi) - \bx(\varnothing)) + w_p (\bx(\bi, \bp) - \bx(\bi))
    \end{aligned}
\label{equation:app_cfg}
\end{equation}
\cref{equation:app_cfg} was chosen such that:
(1) if the CFG scales for image $w_i$ and text prompt $w_p$ are both equal to $1$, the resulting vector $\tilde{\bx}$ should be equal to the prediction $\bx(\bi, \bp)$ conditioned on the image and text, without Classifier Free Guidance.
(2) if the CFG scales for image $w_i$ and text $w_p$ are both equal to $0$, the resulting vector $\tilde{\bx}$ should be equal to the un-conditioned prediction $\bx(\varnothing)$.

In \cref{equation:app_cfg} there is an ordering on the conditionings.
We also considered alternate orderings in which we start with the text conditioning first instead of the image conditioning:
\begin{equation}
    \begin{aligned}
        \tilde{\bx} = \bx + w_p (\bx(\bp) - \bx(\varnothing)) + w_i (\bx(\bi, \bp) - \bx(\bp))
    \end{aligned}
\label{equation:app_cfg_reversed}
\end{equation}
\cref{equation:app_cfg_reversed} did not lead to improvement over \cref{equation:app_cfg}, but required significantly different values for $w_i$ and $w_p$  to work equally well.
We also considered formulas without ordering between the two conditionings, for instance:
\begin{gather*}
    \tilde{\bx} = \bx + w_i (\bx(\bi) - x(\varnothing)) + w_p(\bx(\bp) - x(\varnothing)) \\
    \text{ and } \\
    \tilde{\bx} = \bx(\bi, \bp) + w'_i (\bx(\bi, \bp) - \bx(\bp)) + w'_p (\bx(\bi, \bp) - x(\bi)) \\
    \text{where } w'_i = (w_i - 1) \text{ and } w'_p = (w_p - 1)
\end{gather*}
Similar to \cref{equation:app_cfg_reversed}, those formulas did not improve over \cref{equation:app_cfg},
and in addition miss the useful properties listed above.

\noindent \textbf{Selecting CFG scales.}
\cref{equation:app_cfg} requires to find the guidance factor $w_i$ for image and $w_p$ for text.
We found that these factors influence the motion in the generated videos.
To quantify this, we measure a `motion score' on the generated videos by computing the mean energy of the motion vectors in the resulting \texttt{H.264} encoding.
We found that the motion score was a good proxy for the amount of motion, but did not provide signal into consistency of the motion.
Higher motion as computed through motion vectors does not necessarily translate to interesting movement, as it could be undesirable jitter, or reflect poor object consistency.
\cref{tab:cfg_scales_and_motion} shows how the CFG scales directly influence the amount of motion in the generated videos.

After narrowing down a few CFG value combinations by looking at the resulting motion score, we identified the best values by visual inspection and human studies.
Qualitatively, we found that the (1) higher $w_i$ for a fixed $w_p$, the more the model stays close to the initial image and favors camera motion; and (2) the higher $w_p$ for a fixed $w_i$, the more the model favors movement at the expense of object consistency.

\input{tables/cfg_scales_motion.tex}

\noindent
\textbf{Frame Interpolation Model.}
Here, we include extra details on the frame interpolation model, $\interpolateModel$.
First we explain our masked zero-interleaving strategy.
Second we explain how we interpolate 16-frame 4fps videos from $\extrapolateModel$.
\cref{sec:impl_details} in the main paper details how $\interpolateModel$ is trained to take 8 zero-interleaved frames (generated from $\extrapolateModel$ at 4fps) as conditioning input and generate 37 frames at 16fps.
One option for training an interpolation model that increases the fps by 4-fold is to generate 3 new frames between every pair of input frames (as in~\cite{Blattmann2023AlignYL}).
However, the downside to this approach is that the resulting interpolated video has a slightly shorter duration than the input video (since every input frame has 3 new generated frames after it, except the last input frame).
We instead take the approach of using $\interpolateModel$ to increase the duration of the input video, and we design a zero-interleaving scheme accordingly.
Our interpolation model is trained to generate 3 new frames between every pair of frames, and also 4 new frames either side of the input video.
As a result, during training $\interpolateModel$ takes as conditioning input a 2s video, and generates a 2.3s video.

For interpolating 16-frame input videos from $\extrapolateModel$ (as described in~\cref{sec:vs_prior_work} in the main paper), we simply split the videos into two 8-frame videos and run interpolation on both independendly.
In order to construct our final interpolated video, we discard the \textit{overlapping} frames (the last 5 frames of the first interpolated video, and the first 4 of the second), and concatenate the two videos frame-wise.
The resulting interpolated video is 65 frames long at 16fps (4.06 seconds in duration -- we refer to these videos as 4 seconds long in the main paper for brevity).

\section{Additional experiments}

We detail additional experiments, viz. (i) an investigation into the effect of the initial image on our video generations,
(ii) a quantitative comparison to prior work in image animation with automated metrics,
(iii) a joint investigation into the effect of the number of training steps and data,
and finally (iv) an analysis into the effect of the amount of training data.

\input{tables/appendix_gen_evals.tex}
\par \noindent \textbf{Image conditioning for commercial \textToVShort systems.}
We study the effect of image conditioning on the commercial \textToVShort solution from \gen~\cite{gen} in~\cref{tab:i2v_gen2}.
The \gen API has two video generation variants: (1) A pure \textToVShort API that accepts a text prompt as input and generates a video; and (2) an "image + text" API, denoted as \gen \imageToVShort, that takes an image and a text prompt as input to generate a video.
We use images generated from~\cite{podell2023sdxl} for the \gen \imageToVShort variant.

We observe that the \gen \imageToVShort variant outperforms the \gen API that only accepts a text prompt as input.
We benchmark \OURS against both variants of the API and observe that it outperforms \gen and the stronger \gen \imageToVShort API.
In~\cref{tab:i2v_human_eval}, we also compare \OURS using the same images as \gen \imageToVShort for ``image animation'' and observe that \OURS outperforms \gen \imageToVShort in that setting as well.

\par \noindent \textbf{Automated metrics for image animation.}
We follow the setting from~\cref{tab:i2v_human_eval} and report automated metrics for comparison in~\cref{tab:i2v_automatic_metrics}.
Following~\cite{esser2023structure, 2023videocomposer}, we report Frame consistency (FC) and Text consistency (TC).
We also report CLIP Image similarity~\cite{brooks2022instructpix2pix} (IC) to measure the fidelity of
generated frames to the conditioned image.
We use CLIP ViT-B/32 model for all the metrics.
Compared to \vidcomp~\cite{2023videocomposer}, \Ours generates smoother motion, as measure by frame consistency, maintains a higher faithfulness to the conditioned image, as measured by the image score, while adhering to the text on both the prompt sets.
\Ours fares slightly lower compared to \pika and \gen on all three metrics. Upon further inspection, \Ours (motion score of 4.98) generates more motion compared to \pika and \gen (motion scores of 0.63 and 3.29 respectively).
Frame and image consistency favour static videos resulting in the lower scores of \Ours on these metrics.

\begin{table}[htb]
    \centering
    \resizebox{0.6\linewidth}{!}{
    \setlength{\tabcolsep}{2pt}
    \input{tabulars/i2v_automatic_eval.tex}
    }
    \caption{\textbf{Automatic evaluation of \Ours \vs prior work in text-conditioned image animation.} We compare \Ours against three contemporary methods following the settings from~\ref{tab:i2v_human_eval}
    using Frame consistency (FC), Image similarity (IC), and Text consistency (TC).
    \Ours outperforms \vidcomp across both the prompt sets and all three metrics.
    Automatic metrics favor static videos to ones with motion, resulting in lower scores for \Ours compared to \pika and \gen.
    }
    \label{tab:i2v_automatic_metrics}
\end{table}

\begin{figure}
	\centering
    \resizebox{0.8\linewidth}{!}{

    \input{tikz_graph_sources/iters_vs_perf.tex}
    }
    \vspace{-0.1in}
    \caption{{\bf Performance \vs training iterations.} On training the $256$px stage for fewer or more iterations, we compare the generations after the same $512$px finetuning to the 100\% trained model via human evaluations, both before and after HQ finetuning. We observe a drop in performance with fewer or more iterations, indicating that around $70$K steps of low-resolution high-FPS pretraining stage is optimal.
    }
    \label{pgf:iters_vs_perf}
\end{figure}
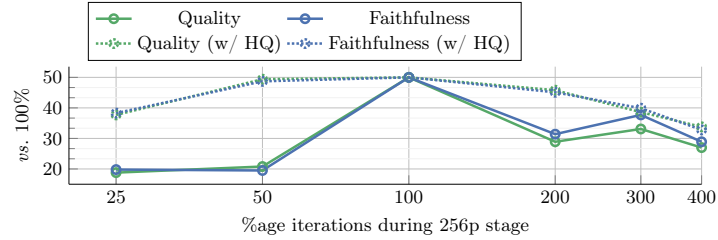

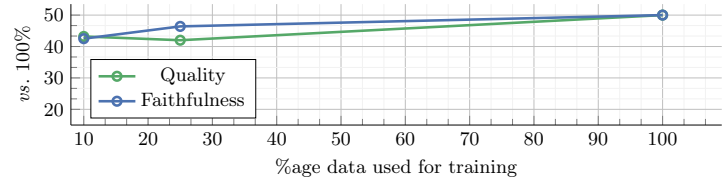
\begin{figure}
	\centering
    \resizebox{0.8\linewidth}{!}{

    \input{tikz_graph_sources/data_vs_perf.tex}
    }
    \vspace{-0.1in}
    \caption{{\bf Performance \vs training data.} We train our model with less data (for both $256$px and $512$px stages) while keeping the training steps constant,
    and compare the generations with the the 100\% data model via human evaluations.
    We observe that even with 10\% data, we only see a slight degradation in performance ($\sim 43$\% on both Quality and Faithfulness), showcasing that
    our method works well even with a fraction of the data.}
    \label{pgf:data_vs_perf}
\end{figure}

\par \noindent \textbf{Effect of the number of training steps and data.}
In~\cref{pgf:iters_vs_perf}, we vary the number of training steps in the initial low-resolution high-FPS pretraining stage. Note that since we run one full epoch through the data during this training stage, reducing the steps correspondingly also reduces the amount of training data seen. We finetune each of these models at higher resolution/low FPS ($512$px, $4$fps) for the same (small) number of steps --
$15$K. We compare the model trained with $100\%$ low-resolution pretraining with models with less low-resolution pretraining using human evaluations. We observe a gradual drop in performance as we reduce the low-resolution pretraining iterations to $75\%$, $50\%$ and $25\%$, indicating the importance of that stage.

\par \noindent \textbf{Effect of the amount of training data.}
In~\cref{pgf:data_vs_perf}, we vary the amount of training data, while keeping the training iterations fixed for both the training stages, and perform a similar comparison as in~\cref{pgf:iters_vs_perf}.
Here we find a much smaller drop in performance as we reduce the amount of data. This suggests that \OURS can be trained effectively with relatively much smaller datasets, as long as the model is trained long enough (in terms of training steps).

\input{tables/prompt_datasets.tex}
\input{figures/juice_AMT.tex}

\section{Human evaluations}
\label{appendix:human_eval}

We rely on human evaluations for making quantitative comparisons to prior work.
In~\cref{sec:experiments} in the main paper, we introduce our method for robust human evaluations.
We now give extra details on this method, termed JUICE, and analyse how it improves robustness, and explain how we ensure fairness in the evaluations.
Additionally, in~\cref{tab:prompt_datasets} we summarize the prompt datasets used for evaluations.

\begin{figure}
    \centering
    \resizebox{0.8\linewidth}{!}{
        \input{tikz_graph_sources/juice_bar_plot.tex}
    }
    \caption{\textbf{Human agreement in \OURS \vs \mav}. Distribution of samples with `split' ($2|3$ or $3|2$ votes), `partial' ($4|1$ or $1|4$ votes), or `complete' ($5|0$ or $0|5$ votes) agreement when using a naive evaluation \vs JUICE.
    Our JUICE evaluation reduces ambiguity in the task and results in a $28\%$  reduction in the number of samples with `split' agreement and a $24\%$ increase in the number of samples with `complete' agreement.
    This improves Fleiss' kappa from 0.004 to 0.31.}
    \label{pgf:juice_agreement}
\end{figure}

\subsection{Robust Human Evaluations with JUICE}
When comparing to prior work, we use human evaluations to compare the generations from pairs of models.
Unlike the naive approach, where evaluators simply pick their choice from a pair of generations, we ask the evaluators to select a reason when making their choice.
We call this approach JUICE, where evaluators are asked to `justify your choice'.
We show an example of the templates used for human evaluations for both video quality and text faithfulness in Figure~\ref{fig:amt_template}, where the different possible justifying reasons are shown.
One challenge faced when asking evaluators to justify their choice is that human evaluators who are not experts in video generation may not understand what is meant by terms such as ``Object/scene consistency'' or ``Temporal text alignment'' or may have subjective interpretations, which would reduce the robustness of the evaluations.
To alleviate this challenge, for each justifying option we show the human evaluators examples of generated video comparisons where each of the factors could be used is used in determining a winner.
It is important that when giving human evaluators training examples such as these that we do not bias them towards \OURS's generations over those of prior work.
Thus, to ensure fairness in the comparisons, we make sure that these training examples include cases where generated videos from different prior works are superior to \OURS and vice-versa.
As detailed in the main paper, for each comparison between two videos from two different models, we use the majority vote from 5 different human evaluators.
To further reduce annotator bias we make sure that the relative positioning of the generated videos being shown to the human evaluators is randomized.
For details on how we ensure fairness in human evaluations when comparing videos with different resolutions, see~\cref{appendix:comparisons_to_prior_work}.

Next, we analyze quantitatively how JUICE improves human evaluation reliability and robustness.
To identify unbiased JUICE factors differentiating any two video generation models on \quality and \faithfulness, we made an initial pool of random video samples generated by a few models, and asked internal human raters to explicitly explain their reasoning for picking one model over another. We then categorized them into five reasons for \quality and two for \faithfulness as mentioned in Section~\ref{sec:approach-details}.

\noindent \textbf{Effect of JUICE on improving evaluation reliability and robustness of human evaluations.} We measure the reliability of our human evaluations when evaluators are required
to justify their choice.
For each pair of videos which are compared, we look at the votes for model A \vs model B and call the agreement between annotators either `split' ($2|3$ or $3|2$ votes), `partial' ($4|1$ or $1|4$ votes), or `complete' ($5|0$ or $0|5$ votes).
We run human evaluations comparing our generations \vs \mav, first using a naive evaluation template and then with JUICE, and show the results in \cref{pgf:juice_agreement}.
We observe that the number of samples with `split' agreement is decreased significantly by $28\%$, and the number of `complete' agreements is increased by $24\%$.

\begin{figure}
    \centering
    \resizebox{0.8\linewidth}{!}{
        \input{tikz_graph_sources/fleiss_kappa_plot.tex}
    }
    \caption{{\bf Analysis of Fleiss' kappa} for a simulated two-class five-raters evaluation task. The blue dot shows the kappa value when we have a complete agreement among evaluators on all the samples.
    We progressively replace samples with $5|0$ or $0|5$ votes (complete agreeement) with either $1|4$ or $4|1$ or $3|2$ or $2|3$ votes and compute the Fleiss' kappa (shown in green and red). The shaded region shows the kappa value for different proportions of samples with complete, partial or split agreements.
}
    \label{pgf:fleiss_kapp}
\end{figure}

Next, we use Fleiss' kappa~\cite{fleiss1973equivalence} as a statistical measure for inter-rater reliability for a fixed number of raters. This metric stands for the amount by which the observed agreement exceeds the agreement by chance, \ie, when the evaluators made their choices completely randomly.
Fleiss' kappa works for any number of evaluators giving categorical ratings and we show the values in~\cref{pgf:fleiss_kapp}.
The value of kappa is always in the range of $[-1,1]$, with positive kappa values representing an agreement. To better understand its behavior and range of scores in our evaluation setup, we perform an experiment on a simulated data representing our specific case of 304 tasks with two classes, model A-vs-B, and five evaluators per task.
We begin with  computing the kappa value when we have a `complete' agreement among evaluators on all tasks, i.e. when all five evaluators choose either model A or model B in each task.
This run receives a kappa value of 1 (blue dot in \cref{pgf:fleiss_kapp}).
We gradually decrease the number of samples with complete agreement by introducing samples with `partial' agreement when four out of five evaluators picked model A or model B (green line in~\cref{pgf:fleiss_kapp})
Similarly, we decrease the number of samples with complete agreement by replacing them with samples where three out of the five evaluators picked model A or model B, illustrated with a red line. As shown in the plot, the kappa value ranges from $-0.2$ (ratings always being `split') to $1.0$ (ratings always having `complete' agreement).
Different proportions of samples with `complete', `partial' or `split' agreements result in a kappa value in the shaded region.
We compute and compare kappa values for the naive evaluation and JUICE evaluation--0.004 and 0.31, respectively--confirming the improvement in the inter-rater reliability of JUICE.

\input{figures/upset.tex}
\input{figures/juice_ours_vs_prior_supp.tex}
\input{figures/juice_ours_vs_prior_lose_supp.tex}

\noindent \textbf{Analyzing human evaluations.}
To clearly understand the strengths of each model in our evaluations, we find the most contributing factors when \OURS generations are preferred to each baseline in Figures~\ref{fig:juice_ours_vs_prior},~\ref{fig:juice_ours_vs_prior_supp}. A more detailed distribution of each reason and its co-occurrence with other factors is illustrated in Figure~\ref{fig:upset}. We similarly, plot the percentage of each reason picked for the best three baseline generations preferred to \OURS in Figure~\ref{fig:juice_ours_vs_prior_supp_lose}.

\section{Comparisons to Prior Work}
\label{appendix:comparisons_to_prior_work}
In~\cref{sec:vs_prior_work} in the main paper, we conduct human evaluations comparing \OURS to prior work.
Here, we share further details and include human evaluation results using a different setup.
Specifically, in~\cref{appendix:datasets_for_comparison_to_prior_work} we outline the prompt datasets that are used in comparisons to prior work.
In~\cref{appendix:sampling_from_commercial} we detail how we sampled from the commercial models that we compare to in the main paper.
In~\cref{appendix:postprocessing_details} we give details on the postprocessing done for the human evaluations in~\cref{fig:teaser_compare} in the main paper.
In~\cref{appendix:original_resolution_comparisons} we include further human evaluations conducted without postprocessing the videos from \OURS or prior work.

\subsection{Datasets used for Prior Work Comparisons}
\label{appendix:datasets_for_comparison_to_prior_work}
Since many of the methods that we compare to in~\cref{fig:teaser_compare} are closed source, we cannot generate samples from all of them with one unified prompt dataset, and instead must construct different datasets via each method's repsective publicly released example generated videos.
In total, we use 5 different prompt datasets.
The human evaluations in~\cref{fig:teaser_compare} for Make-A-Video, Imagen Video, Align Your Latents, PYOCO, and Reuse \& Diffuse were conducted using the prompt datasets from the respective papers (see~\cref{tab:prompt_datasets} for details).
Certain methods that we compare to are either open-source (CogVideo) or can be sampled from through an online interface (Gen2 and Pika Labs).
For these, human evaluations are conducted using the prompt set from Align Your Latents.

\input{tables/appendix_video_dimensions.tex}
\subsection{Sampling from Commercial Models}
\label{appendix:sampling_from_commercial}
The commercially engineered black-box text-to-video models that we compare to (Pika Labs and Gen2) can be sampled from through an online interface.
Here we include details for how we sampled from these models.
In both cases, these interfaces allow for certain hyper-parameters to be chosen which guide the generations.

We selected optimal parameters for each of the models by varying the parameters over multiple generations and choosing those that consistently resulted in the best generations.
For Pika Labs, we use the arguments ``-ar 1:1 -motion 2'' for specifying the aspect ratio and motion.
For Gen2, we use the ``interpolate'' and ``upscale'' arguments and a ``General Motion'' score of 5.
All samples were generated on October 24th 2023.
\subsection{Postprocessing Videos for Comparison}
\label{appendix:postprocessing_details}
\input{tables/appendix_downsampling.tex}
Our goal with our main human evaluations in~\cref{fig:teaser_compare} is to ensure fairness and reduce any human evaluator bias.
To ensure this fairness, we postprocess the videos from each model being compared (as outlined in~\cref{sec:vs_prior_work} in the main paper).
Here, we give further details on the motivation behind this decision, and explain how this postprocessing is done.
Results for human evaluations conducted without any postprocessing are discussed in~\cref{appendix:original_resolution_comparisons}.

\input{tables/appendix_sota_human_eval.tex}

As outlined in~\cref{appendix:human_eval}, our human evaluations are conducted by showing evaluators repeated comparisons of videos generated by two different models for the same prompt, and asking them which model they prefer in terms of the metric being evaluated.
It is key for the fairness of the human evaluation that the evaluator treats each comparison independently.
It is hence important that the evaluator does not know which model generated which video, otherwise they can become biased towards one model over the other.
Since each method generates videos at different dimensions (see~\cref{tab:app_video_dimensions}), conducting the human evaluations without postprocessing the videos would lead to this annotator bias.
Hence we decide to postprocess the videos being compared such that they have the same aspect-ratios, dimensions and frame rates so that they are indistinguishable aside from their generated content.
For each pair of models being compared, we downsample these dimensions to the minimum value between the two models (see~\cref{tab:app_downsampling_settings} for details).
Next, we detail how we postprocess the videos.

\noindent \textbf{Aspect Ratio.}
Since \OURS generates videos at a 1:1 aspect ratio, all videos are postprocessed to a 1:1 aspect ratio by centre cropping.

\noindent \textbf{Spatial Dimension.}
The height and width of videos are adjusted using bilinear interpolation.

\noindent \textbf{Video Duration.}
The duration of videos is reduced via temporal centre cropping.

\noindent \textbf{Frame rate.}
The frame rate is adjusted using torchvision.
The number of frames is selected according to the desired frame rate and video duration.

Next we discuss human evaluation results where videos are compared without any postprocessing.

\subsection{Prior Work at Original Dimensions}
\label{appendix:original_resolution_comparisons}
In this Section, we include further human evaluation results between \OURS and prior work where we do not perform any postprocessing on the videos and conduct the evaluations with the original dimensions (as detailed in~\cref{tab:app_video_dimensions}).
In this system-level comparison, human evaluators are comparing between videos that may have very different aspect ratios, durations, and frame rates, and in turn may become biased towards one model over another after seeing repeated comparisons.
We note that since the dimensions of the videos here are so large, we must scale the height of each video so that both compared videos can fit on one screen for human evaluators.
All other dimensions remain as in the original sampled videos.
The results are in~\cref{tab:appendix_sota_human_eval}.
Similar to the human evaluations conducted with postprocessed videos in~\cref{fig:teaser_compare} in the main paper, \OURS significantly outperforms prior work in terms of both text faithfulness and video quality.
Even when comparing \OURS's generated videos to generated videos with longer durations (including PYOCO, Imagen Video), wider aspect ratios (incliding Gen2, Align Your Latents), or higher frame rates (including Pika, Gen2), human evaluators still prefer \OURS's generated videos in both metrics.
We hypothesize that the vastly improved frame quality and temporal consistency of \OURS still outweighs any benefits that come from any larger dimensions in the prior work's videos.

Interestingly, \OURS wins by larger margins here than in the postprocessed setting (an average win rate of 93.8\% in quality and 93.1\% in faithfulness here, vs.  $91.8\%$ and $86.6\%$ in the postprocessed comparison).
We conjecture that this improvement in win rates for \OURS may be due to the potential evaluator bias introduced in this evaluation setting.
This introduced bias tends to favor \OURS since our video generations are on average superior in terms of quality and faithfulness than those of prior work.
Hence in this paper we primarily report and refer to the human evaluation scores from the fairer postprocessed setting.

\section{Qualitative Results}
\label{appendix:qual_comparisons_to_prior_work}
In this Section, we include additional qualitative results from \OURS (in~\cref{appendix:qual_results_ours}), and further qualitative comparisons between \OURS and prior work (in~\cref{appendix:qual_results_compare})

\subsection{Further \OURS qualitative Results} \label{appendix:qual_results_ours}
Examples of \OURS's \textToVShort generations are shown in ~\cref{fig:our_gens_t2v}, and \OURS's \imageToVShort generations are shown in~\cref{fig:our_gens_i2v}.
As shown, \OURS generates high quality video generations that are faithful to the text in \textToVShort and to both the image and the text in \imageToVShort.
The videos have high pixel sharpness, motion smoothness and object consistency, and are visually compelling.
\OURS generates high quality videos for both natural prompts and fantastical prompts.
We hypothesize that this is because \OURS is effectively able to retain the wide range of styles and diversity of the \textToIShort model due to the factorized approach.

\subsection{Qualitative Comparisons to Prior Work} \label{appendix:qual_results_compare}
We include further qualitative comparisons to prior work in~\cref{fig:compare_to_prior_1,,fig:compare_to_prior_2,,,fig:compare_to_prior_3,,,fig:compare_to_prior_4,,,fig:compare_to_prior_5,,fig:compare_to_prior_6}.
This Section complements~\cref{sec:vs_prior_work} in the main paper where we quantatively demonstrate via human evaluation that \OURS significantly outperforms the prior work in both video quality and text faithfulness.
\OURS consistently generates videos that are significantly more text faithful (see~\cref{fig:compare_to_prior_2,,fig:compare_to_prior_4}), with greater motion smoothness and consistency (see~\cref{fig:compare_to_prior_3,,fig:compare_to_prior_5}), far higher pixel sharpess (see~\cref{fig:compare_to_prior_6}), and that are overall more visually compelling (see~\cref{fig:compare_to_prior_1}) than the prior work.

\input{figures/Appendix_Qual_Compare_Figures/Our_generations_T2V/our_gen_figure.tex}

\input{figures/Appendix_Qual_Compare_Figures/Our_generations_I2V/our_gen_figure.tex}

\input{figures/Appendix_Qual_Compare_Figures/Our_generations_prior_work_compare/our_gen_figure.tex}
\input{figures/Appendix_Qual_Compare_Figures/Our_generations_prior_work_compare/our_gen_figure2.tex}

\input{figures/Appendix_Qual_Compare_Figures/Our_generations_prior_work_compare2/our_gen_figure.tex}

\input{figures/Appendix_Qual_Compare_Figures/Our_generations_prior_work_compare3/our_gen_figure.tex}

\input{figures/Appendix_Qual_Compare_Figures/Our_generations_prior_work_compare4/our_gen_figure.tex}

\input{figures/Appendix_Qual_Compare_Figures/Our_generations_prior_work_compare5/our_gen_figure.tex}

\input{figures/ucf.tex}

%% file: tables/appendix_unet_arch.tex
\begin{table}[!htb]
    \centering
    \resizebox{0.5\linewidth}{!}{
        \begin{tabular}{l|c}
            \bf Setting & \bf Value  \\
            \midrule
            {\tt input\_shape} & [17, $\vTime$, 64, 64] \\
            {\tt output\_shape} & [8, $\vTime$, 64, 64] \\
            {\tt model\_channels} & 384 \\
            {\tt attention\_resolutions} & [4, 2, 1] \\
            {\tt num\_res\_blocks} & [3, 4, 4, 4] \\
            {\tt channel\_multipliers} & [1, 2, 4, 4] \\
            {\tt use\_spatial\_attention} & True \\
            {\tt use\_temporal\_attention} & True \\
            {\tt transformer\_config:} &  \\
            \quad {\tt d\_head} & 64 \\
            \quad {\tt num\_layers} & 2 \\
            \quad {\tt context\_dim\_layer\_1} & 768 \\
            \quad {\tt context\_dim\_layer\_2} & 2048 \\
        \end{tabular}
    }
    \caption{
        \textbf{U-Net architecture details.} Our U-Net contains $4.3$B total parameters, out of which $2.7$B are
        initialized from our pretrained text-to-image model and kept frozen, resulting in $1.7$B trainable parameters.
        $\vTime$ is the total frames produced by the model.
    }
    \label{tab:app_unet_arch}
\end{table}

%% file: tables/appendix_vae_arch.tex
\begin{table}[!htb]
    \centering
    \resizebox{0.5\linewidth}{!}{
        \begin{tabular}{l|c}
            \bf Setting & \bf Value  \\
            \midrule
            {\tt type} & AutoencoderKL~\cite{rombach2021highresolution} \\
            {\tt z\_channels} & 8 \\
            {\tt in\_channels} & 3 \\
            {\tt out\_channels} & 3 \\
            {\tt base\_channels} & 128 \\
            {\tt channel\_multipliers} & [1, 2, 4, 4] \\
            {\tt num\_res\_blocks} & 2 \\
        \end{tabular}
    }
    \caption{
        \textbf{VAE architecture details.} We use an image based VAE and apply it to videos frame-by-frame. Our VAE
        encoder downsamples videos spatially by $8\times8$ and produces 8 channel latents.
    }
    \label{tab:app_vae_arch}
\end{table}

%% file: tables/appendix_training_hyperparams.tex
\begin{table}[!htb]
    \begin{center}
        \centering
        \resizebox{0.7\linewidth}{!}{
            \begin{tabular}{l|c|c|c|c}
                \bf \multirow{2}{*}{\bf Setting} & \multicolumn{4}{c}{\bf Training stage} \\
                 & \bf \bf 256px & \bf 512px & \bf HQ FT & \bf FI \\
                 & $\extrapolateModel$ & $\extrapolateModel$ & $\extrapolateModel$ & $\interpolateModel$ \\
                \midrule
                Diffusion settings: \\
                \quad {Loss} & \multicolumn{4}{c}{Mean Squared Error} \\
                \quad {Timesteps} & \multicolumn{4}{c}{1000} \\
                \quad {Noise Schedule} & \multicolumn{1}{c|}{quad} & \multicolumn{3}{c}{quad$^*$} \\
                \quad {Beta start} & \multicolumn{1}{c|}{$8.5\times 10^{-4}$} & \multicolumn{3}{c}{$8.5\times 10^{-4^*}$} \\
                \quad {Beta end} & \multicolumn{1}{c|}{$1.2\times 10^{-2}$} & \multicolumn{3}{c}{$1.2\times 10^{-2^*}$} \\
                \quad {Var type} & \multicolumn{4}{c}{Fixed small} \\
                \quad {Prediction mode} & {eps-pred} & \multicolumn{3}{|c}{v-pred} \\
                \quad {0-term-SNR rescale} & False & \multicolumn{3}{|c}{True~\cite{lin2023common}} \\
                Optimizer & \multicolumn{4}{c}{AdamW~\cite{loshchilov2017decoupled}} \\
                Optimizer Momentum & \multicolumn{4}{c}{$\beta_1=0.9,\beta_2=0.999$} \\
                Learning rate: \\
                \quad Schedule & \multicolumn{4}{c}{Constant} \\
                \quad Warmup Schedule & \multicolumn{4}{c}{Linear} \\
                \quad Peak & \multicolumn{2}{c|}{1e-4} & 2.5e-5 & 1.5e-4\\
                \quad Warmup Steps & \multicolumn{2}{c|}{1K} & 10K & 1.5K \\
                Weight decay & \multicolumn{2}{c|}{0.0} & 1e-4 & 0.0\\
                Dataset size & \multicolumn{2}{c|}{34M} & 1.6K & 34M\\
                Batch size & \multicolumn{2}{c|}{512} & 64 & 384 \\
                Transforms: \\
                \quad {Clip Sampler} & \multicolumn{4}{c}{Uniform} \\
                \quad {Frame Sampler} & \multicolumn{4}{c}{Uniform} \\
                \quad {Resize} \\
                \qquad {interpolation} & \multicolumn{4}{c}{Box + Bicubic} \\
                \qquad {size} & 256px & \multicolumn{3}{|c}{512px} \\
                \quad {Center Crop} & 256px & \multicolumn{3}{|c}{512px} \\
                \quad {Normalize Range} & \multicolumn{4}{c}{[-1, 1]} \\
            \end{tabular}
        }
    \end{center}
    \caption{
        \textbf{Training hyperparameters} for various stages in our pipeline:
        256px training, 512px training, High Quality finetuning (HQ FT), and frame interpolation (FI).
        $^*$: noise schedules are changed afterwards with zero terminal-SNR rescaling~\cite{lin2023common}.
    }
    \label{tab:app_training_settings}
\end{table}

%% file: tables/cfg_scales_motion.tex
\begin{table}[]
    \centering
    \resizebox{0.5\linewidth}{!}{
    \begin{tabular}{c|c|c|c}
        Model & $w_p$ & $w_i$ & Motion Score \\
        \hline
        w/o HQ finetuning & 2.0  & 1.0 & 1.87 \\
        w/o HQ finetuning & 8.0  & 1.0 & 2.87 \\
        w/o HQ finetuning & 16.0 & 1.0 & 3.86 \\
        \hline
        w/o HQ finetuning & 8.0 & 1.0 & 2.87 \\
        w/o HQ finetuning & 8.0 & 2.0 & 0.61 \\
        w/o HQ finetuning & 8.0 & 3.0 & 0.25 \\
        \hline
        HQ finetuned & 2.0 & 2.0  & 11.1 \\
        HQ finetuned & 8.0 & 2.0  & 12.7 \\
        HQ finetuned & 16.0 & 2.0  & 13.5 \\
        \hline
        HQ finetuned & 8.0 & 1.0  & 14.9 \\
        HQ finetuned & 8.0 & 2.0  & 12.7 \\
        HQ finetuned & 8.0 & 3.0  & 11.3 \\
        \hline
    \end{tabular}
    }
    \caption{
        We measure the amount of motion in the generated videos using an automated motion score where a higher value reflects more motion.
        We use the prompts from~\cite{singer2023makeavideo}.
        The ratio of text CFG scale $w_p$ to image CFG scale $w_i$ influences the amount of motion in the video.
        We also observe that, w/o HQ fine-tuning, motion is much less and that the relative effect of CFG scales is even more pronounced.
    }
    \label{tab:cfg_scales_and_motion}
\end{table}

%% file: tables/appendix_gen_evals.tex
\begin{table}
    \centering
    \resizebox{0.5\linewidth}{!}{
    \begin{tabular}{cc | cc}
        \bf Method & \#Prompts &  \bf \qualityShort & \bf \faithfulnessShort \\
       \hline
        \gen \vs \gen \imageToVShort & \multirow{3}{*}{$65$ \cite{Blattmann2023AlignYL}}  & $41.5$  & $44.6$ \\
        \OURS \vs \gen \imageToVShort & & $72.3$ & $78.4$ \\
        \OURS \vs \gen & & $78.5$ & $87.7$ \\
   \end{tabular}%
    }
    \caption{\textbf{Image conditioning for commercial \textToVShort}
        We compare \OURS against two video generation variants of \gen API: (1) \gen which accepts only a text prompt as input and (2) \gen \imageToVShort which accepts an input image (generated using~\cite{podell2023sdxl}) and a text prompt.
        We observe that the second variant (\gen \imageToVShort) outperforms the \textToV \gen variant.
        \OURS's generations are strongly preferred to both the variants of the \gen API.
    }
    \label{tab:i2v_gen2}
\end{table}

%% file: tabulars/i2v_automatic_eval.tex
\begin{tabular}{cc | ccc}
    \bf Method &  \bf Dataset & \bf FC ($\uparrow$) & \bf IC ($\uparrow$) &  \bf TC($\uparrow$) \\
    \hline
    \vidcomp~\cite{2023videocomposer}& \multirow{4}{*}{\ayldatasetshort~\cite{Blattmann2023AlignYL}} & 96.8 & 86.4 & 33.3\\
    \pika \imageToVShort & &99.9& 95.0& 34.6 \\
    \gen \imageToVShort & &99.9& 96.8& 34.3 \\
    \Ours & &99.3 & 94.2 & 34.2\\
    \hline
    \vidcomp~\cite{2023videocomposer}& \multirow{2}{*}{\mavdatasetshort~\cite{singer2023makeavideo}} & 95.2 & 82.6 & 31.3\\
    \Ours & &98.9 & 91.3 & 32.1\\
\end{tabular}%

%% file: tikz_graph_sources/iters_vs_perf.tex
\begin{tikzpicture}
    \begin{axis}[
        xmode=log,
        xmin=20,
        xmax=400,
        xtick={25, 50, 100, 200, 300, 400},
        xticklabels={25, 50, 100, 200, 300, 400},
        ytick={20, 30, 40, 50},
        ymin=15,
        xlabel={\%age iterations during $256$p stage},
        grid=both,
        grid style={line width=.1pt, draw=gray!10},
        major grid style={line width=.2pt,draw=gray!50},
        minor tick num=2,
        axis x line*=bottom,
        axis y line*=left,
        height=1.4in,
        width=\linewidth,
        ylabel style= {align=center},
        ylabel={\vs 100\%},
        yticklabel style = {font=\small},
        xticklabel style = {font=\small},
        legend columns=2,
        legend style={cells={align=left}, font=\small, at={(0.03,1.3)}, anchor=west},
    ]
    \addplot[mark=o, very thick, snsGreenColor] plot coordinates {
        (25, 18.8)  %
        (50, 20.8)  %
        (100, 50.0)
        (200, 28.9)  %
        (300, 33.1)  %
        (400, 27.0) %
    };
    \addlegendentry{Quality}
    \addplot[mark=o, very thick, snsBlueColor] plot coordinates {
        (25, 19.8)
        (50, 19.5)
        (100, 50.0)
        (200, 31.4)
        (300, 37.7)
        (400, 28.9)
    };
    \addlegendentry{Faithfulness}

    \addplot[mark=o, very thick, densely dotted, snsGreenColor] plot coordinates {
        (25, 37.7)  %
        (50, 49.4)  %
        (100, 50.0)  %
        (200, 45.8)  %
        (300, 38.6)  %
        (400, 33.8) %
    };
    \addlegendentry{Quality (w/ HQ)}
    \addplot[mark=o, very thick, densely dotted, snsBlueColor] plot coordinates {
        (25, 38.3)
        (50, 48.7)
        (100, 50.0)
        (200, 45.1)
        (300, 39.9)
        (400, 32.8)
    };
    \addlegendentry{Faithfulness (w/ HQ)}
    \end{axis}
\end{tikzpicture}

%% file: tikz_graph_sources/data_vs_perf.tex
\begin{tikzpicture}
    \begin{axis}[
        ytick={20, 30, 40, 50},
        ymin=15,
        xmin=8,
        xlabel={\%age data used for training},
        grid=both,
        grid style={line width=.1pt, draw=gray!10},
        major grid style={line width=.2pt,draw=gray!50},
        minor tick num=2,
        axis x line*=bottom,
        axis y line*=left,
        height=1.4in,
        width=\linewidth,
        ylabel style= {align=center},
        ylabel={\vs 100\%},
        yticklabel style = {font=\small},
        xticklabel style = {font=\small},
        legend style={cells={align=left}, font=\small, at={(0.03,0.3)}, anchor=west},
    ]
    \addplot[mark=o, very thick, snsGreenColor] plot coordinates {
        (10, 43.2)
        (25, 42.0)
        (100, 50.0)
    };
    \addlegendentry{Quality}
    \addplot[mark=o, very thick, snsBlueColor] plot coordinates {
        (10, 42.5)
        (25, 46.4)
        (100, 50.0)
    };
    \addlegendentry{Faithfulness}
    \end{axis}
\end{tikzpicture}

%% file: tables/prompt_datasets.tex
\begin{table}[htb]
    \centering
    \resizebox{0.4\linewidth}{!}{
    \setlength{\tabcolsep}{2pt}
    \begin{tabular}{c|c}
        \bf Source & \bf \#prompts\\
        \hline
        \mav~\cite{singer2023makeavideo} & 307 \\
        \imagenvideo~\cite{ho2022imagen} & 55 \\
        \ayol~\cite{Blattmann2023AlignYL} & 65 \\
        \pyoco~\cite{ge2023preserve} & 74 \\
        \reusediffuse~\cite{gu2023reuse} & 23 \\
    \end{tabular}
    }
    \caption{\textbf{Text prompt sets} used for evaluation in our work. We use the text prompt sets from prior work to generate videos.
    }
    \label{tab:prompt_datasets}
\end{table}

%% file: figures/juice_AMT.tex
\begin{figure*}[htb]
    \centering
    \captionsetup{type=figure}
    \includegraphics[width=\linewidth]{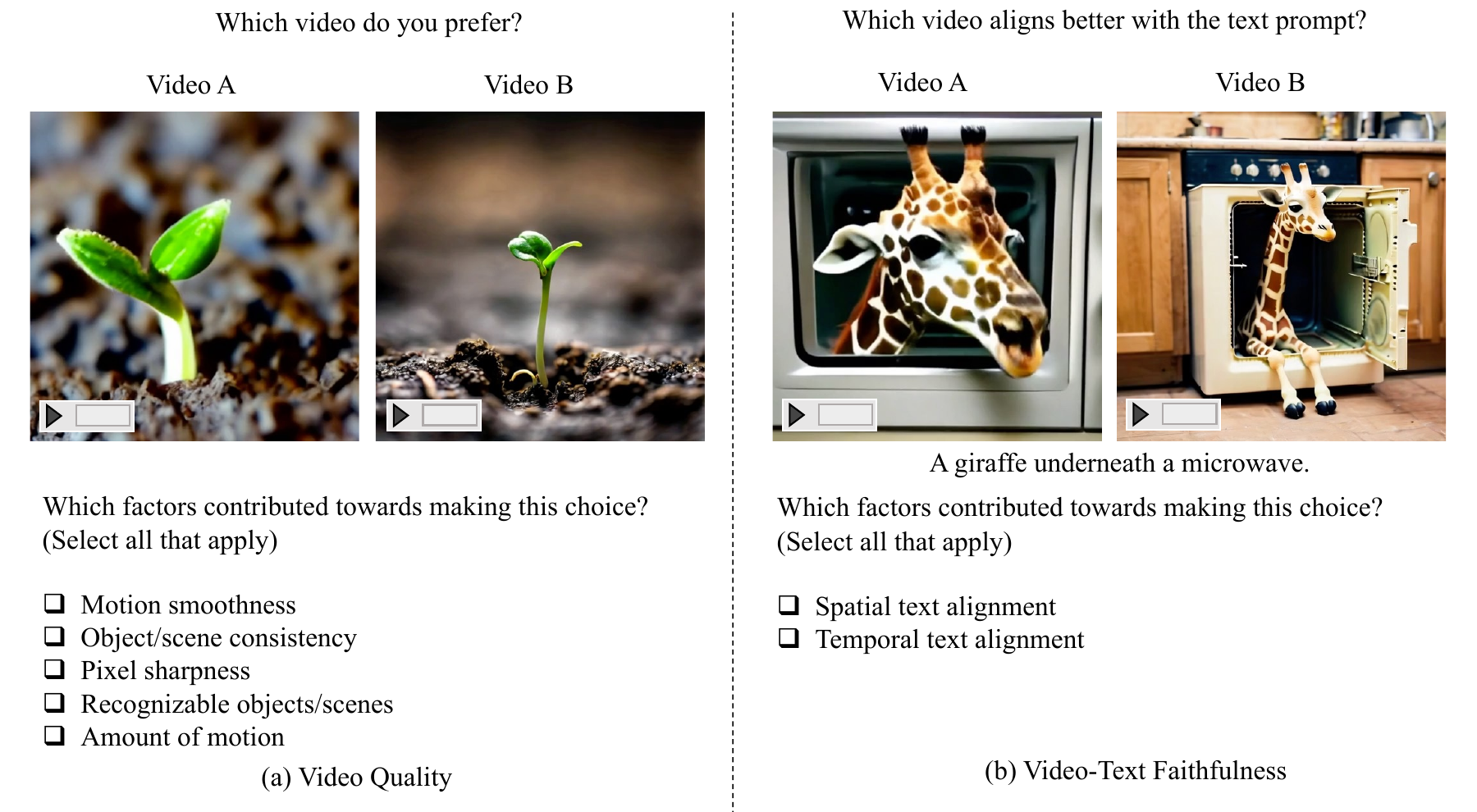}
    \captionof{figure}{
     The JUICE template to compare two models in terms of (a) video quality and (b) video-text alignment.
     Here, human evaluators must justify their choice of which generated video is superior through the selection of one or more contributing factors, shown here.
     To ensure that human evaluators have the same understanding of what these factors mean, we additionally provide training examples of video comparisons where each of the justifying factors could be used in selecting a winner.}
\label{fig:amt_template}
\end{figure*}

%% file: tikz_graph_sources/juice_bar_plot.tex
\pgfplotstableread[row sep=\\,col sep=&]{
    interval & naive & juice  \\
    split     & 184  & 101    \\
    partial     & 106 & 89    \\
    complete    & 17  & 90  \\
    }\mydata

\begin{tikzpicture}
    \begin{axis}[
            title= Distribution of samples with different levels of agreement,
            height=2.5in,
            width=\linewidth,
            bar width=0.4cm,
            xtick=data,
            xtick style={draw=none},
            xtick pos=bottom,
            ytick pos=left,
            enlarge x limits=0.2,
            ybar,
            ylabel = {Number of samples},
            xlabel = {Types of agreement},
            ymin=0,
            y tick label style={font=\small},
            legend style={cells={align=left}, font=\footnotesize, at={(0.98,0.88)}, column sep=5pt, anchor=east}, %
            legend cell align={left},%
            legend image code/.code={
            \draw [#1] (0cm,-0.1cm) rectangle (0.5cm,0.1cm); }, %
            symbolic x coords={split, partial,complete},
        ]
        \addplot[BlueColor, fill=BlueColor, bar width=15pt] table[x=interval,y=naive]{\mydata};
        \addlegendentry{Naive template}
        \addplot[OrangeColor, fill=OrangeColor, bar width=15pt] table[x=interval,y=juice]{\mydata};
        \addlegendentry{Juice}

        \draw[thick,-latex, xshift=8pt, draw=ForestGreen] (axis cs:split, 184) -- (axis cs:split, 101) node[midway, right, font=\small, ForestGreen] {28\% };
        \draw[thick,-latex, xshift=-8pt, draw=ForestGreen] (axis cs:complete, 17) -- (axis cs:complete, 90) node[midway, left, font=\small, ForestGreen] {24\% };
    \end{axis}
\end{tikzpicture}

%% file: tikz_graph_sources/fleiss_kappa_plot.tex
\begin{tikzpicture}
    \begin{axis}[
        ytick={-0.2, 0.2, 0.6, 1.0},
        ymin=-0.3,
        xmin=-0.05,
        xlabel={Proportion of samples with complete agreement},
        grid=both,
        grid style={line width=.1pt, draw=gray!10},
        major grid style={line width=.2pt,draw=gray!50},
        minor tick num=2,
        axis x line*=bottom,
        axis y line*=left,
        height=2.8in,
        width=\linewidth,
        ylabel style= {align=center},
        ylabel={Kappa},
        yticklabel style = {font=\small},
        xticklabel style = {font=\small},
        legend style={cells={align=left}, font=\footnotesize, at={(0.05,0.9)}, anchor=west},
    ]
    \addplot[name path=A, mark=none, line width=3pt, snsGreenColor] plot coordinates {
        (1,1)
        (0.9, 0.917)
        (0.8, 0.838)
        (0.7, 0.757)
        (0.6,0.678)
        (0.5, 0.6)
        (0.4, 0.517)
        (0.3, 0.438)
        (0.2, 0.357)
        (0.1, 0.278)
        (0.0, 0.2)

    };
    \addlegendentry{$1|4$ or $4|1$ ratings}
    \addplot[name path=B, mark=none, line width=3pt, RedColor] plot coordinates {
        (1,1)
        (0.9, 0.878)
        (0.8, 0.759)
        (0.7, 0.637)
        (0.6,0.518)
        (0.5, 0.4)
        (0.4, 0.278)
        (0.3, 0.159)
        (0.2, 0.037)
        (0.1, -0.082)
        (0.0, -0.2)
    };
    \addlegendentry{$2|3$ or $3|2$ ratings}
    \addplot[mark=*, BlueColor, mark size=4pt] coordinates {(1,1)};
    \addplot [gray!30] fill between [
        of=A and B,soft clip={domain=0:1},
    ];
    \end{axis}
\end{tikzpicture}

%% file: figures/upset.tex
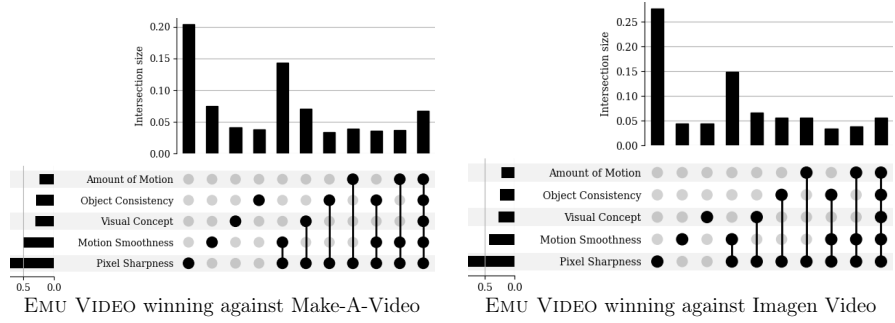
\begin{figure}
    \centering
    \captionsetup{type=figure}
    \vspace{-1cm}
    \input{figures/upset_plot/figure.tex}
    \captionof{figure}{
    Vertical bars show percentage of each reason and its co-occurrence with other reasons picked for \OURS against \mav (left) and \imagenvideo (right). Horizontal bars depict the overall percentage of each reason, similar to Figure~\ref{fig:juice_ours_vs_prior}. Pixel sharpness and motion smoothness are the two most contributing factors in the \OURS win against both baselines.  
    }
\label{fig:upset}
\end{figure}%

%% file: figures/upset_plot/figure.tex
\setlength{\tabcolsep}{10pt}
\resizebox{\linewidth}{!}{%
    \begin{tabular}{cc}
        \includegraphics[width=0.7\linewidth]{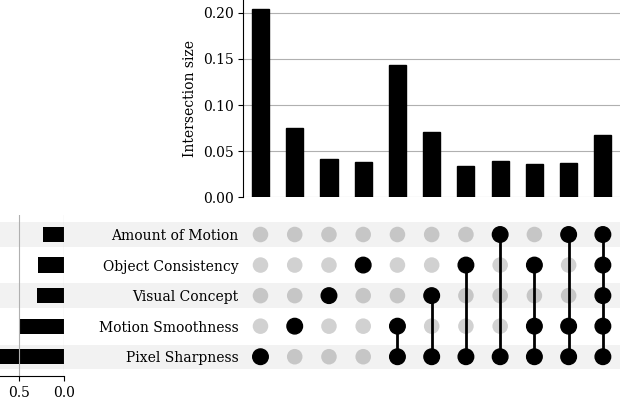} &
        \includegraphics[width=0.7\linewidth]{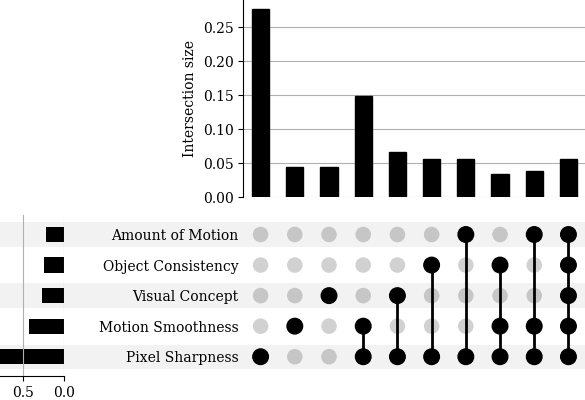} \\
        \large \OURS winning against \mav & 
        \large \OURS winning against \imagenvideo \\
        \setlength{\fboxrule}{2pt}
        \setlength{\fboxsep}{0pt}
    \end{tabular}%
}

%% file: figures/juice_ours_vs_prior_supp.tex
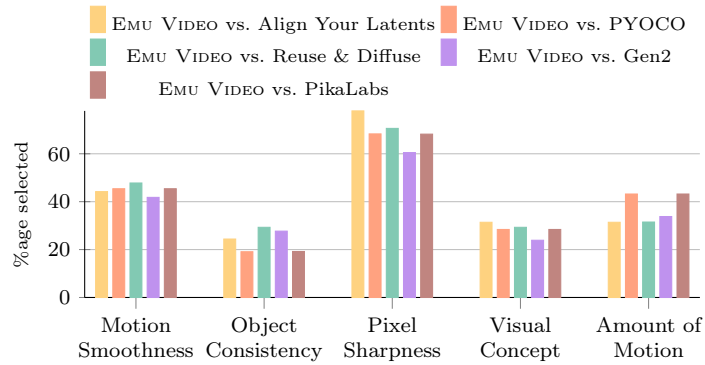
\begin{figure}
    \centering
    \input{figures/juice_supp_all.tex}
    \caption{\textbf{Percentage of each reason selected for samples where \OURS wins against each baseline model on \quality.} Reasons that human evaluators pick \OURS generations over the baseline models from Figure~\ref{fig:teaser_compare} are primarily pixel sharpness and motion smoothness of our videos for most models. Amount of motion in \OURS generations is also an impactful winning factor against \pyoco and \pika. 
    }
    \label{fig:juice_ours_vs_prior_supp}
\end{figure}

%% file: figures/juice_supp_all.tex
\begin{tikzpicture}
    \begin{axis}
    [
    ybar,
    legend columns=2,
    legend style={at={(1.0,1.3)},anchor=east,fill=none,draw=none, legend columns=2},
    ylabel={\%age selected},
    legend style={font=\scriptsize },
    label style={font=\scriptsize},
    tick label style={font=\fontsize{8}{9.6}\selectfont},
    axis x line*=bottom,
    axis y line=left,
   	y axis line style={-},
    ymajorgrids=true,
    ymin=0,
    bar width=4.5pt,
    symbolic x coords={ms, oc, ps, vc, am},
    xticklabels={None, Motion\\Smoothness, Object\\Consistency, Pixel\\Sharpness, Visual\\Concept, Amount of\\Motion},
    xticklabel style={align=center},
    table/col sep=semicolon,
    label style={font=\scriptsize},
    height=1.6in,
    width=0.8\linewidth,
    legend image code/.code={
        \draw [#1] (0cm,-0.1cm) rectangle (0.2cm,0.25cm); },
    ]
    ]
    \addplot [mustard!50,fill=mustard!50] coordinates {(ms, 44.2) (oc, 24.4) (ps, 77.9) (vc, 31.4) (am, 31.4)};
    \addlegendentry{\OURS vs. \ayol}
    \addplot [pastelorange!50,fill=pastelorange!50] coordinates {(ms, 45.4) (oc, 19.1) (ps, 68.3) (vc, 28.4) (am, 43.2)};
    \addlegendentry{\OURS vs. \pyoco}
    \addplot [pastelgreen!50,fill=pastelgreen!50] coordinates {(ms, 47.8) (oc, 29.3) (ps, 70.6) (vc, 29.3) (am, 31.5)};
    \addlegendentry{\OURS vs. \reusediffuse}
    \addplot [pastelpurple!50,fill=pastelpurple!50] coordinates {(ms, 41.8) (oc, 27.7) (ps, 60.5) (vc, 23.9) (am, 33.8)};
    \addlegendentry{\OURS vs. \gen}
    \addplot [pastelred!50,fill=pastelred!50] coordinates {(ms, 45.4) (oc, 19.2) (ps, 68.2) (vc, 28.4) (am, 43.2)};
    \addlegendentry{\OURS vs. \pika}
    \end{axis}
\end{tikzpicture}

%% file: figures/juice_ours_vs_prior_lose_supp.tex
\begin{figure}
    \centering
    \input{figures/juice_supp_all_lose.tex}
    \caption{\textbf{Percentage of each reason selected for samples where each baseline model wins against \OURS on \quality.} Among the few preferred \mav generations from Figure~\ref{fig:teaser_compare} against \OURS, object consistency has been the primary reason, while for \imagenvideo generations, amount of motion has been an additional considerable reason. \gen generations preferred over \OURS are mainly selected due to their motion smoothness and pixel sharpness.
   }
    \label{fig:juice_ours_vs_prior_supp_lose}
\end{figure}

%% file: figures/juice_supp_all_lose.tex
\begin{tikzpicture}
    \begin{axis}
    [
    ybar,
    legend columns=2,
    legend style={at={(1.1,1.3)},anchor=east,fill=none,draw=none, legend columns=2},
    ylabel={\%age selected},
    legend style={font=\scriptsize },
    label style={font=\scriptsize},
    tick label style={font=\fontsize{8}{9.6}\selectfont},
    axis x line*=bottom,
    axis y line=left,
    y axis line style={-},
    ymajorgrids=true,
    ymin=0,
    bar width=7.0pt,
    symbolic x coords={ms, oc, ps, vc, am},
    xticklabels={None, Motion\\Smoothness, Object\\Consistency, Pixel\\Sharpness, Visual\\Concept, Amount of\\Motion},
    xticklabel style={align=center},
    table/col sep=semicolon,
    label style={font=\scriptsize},
    height=1.5in,
    width=0.8\linewidth,
    legend image code/.code={
        \draw [#1] (0cm,-0.1cm) rectangle (0.2cm,0.25cm); },
    ]
        
    \addplot [snsTealColor!50,fill=snsTealColor!50] coordinates {(ms, 23.5) (oc, 50.0) (ps, 20.1) (vc, 35.3) (am, 23.5)};
    \addlegendentry{\mav vs. \OURS}
    \addplot [snsVioletColor!50,fill=snsVioletColor!50] coordinates {(ms, 23.7) (oc, 39.4) (ps, 10.5) (vc, 15.8) (am, 50.0)};
    \addlegendentry{ \imagenvideo vs. \OURS}
    \addplot [pastelpurple!50,fill=pastelpurple!50] coordinates {(ms, 57.7) (oc, 22.2) (ps, 48.8) (vc, 19.9) (am, 24.4)};
    \addlegendentry{\gen vs. \OURS}
    \end{axis}
\end{tikzpicture}

%% file: tables/appendix_video_dimensions.tex
\begin{table}[!htb]
    \begin{center}
        \centering
        \resizebox{0.7\linewidth}{!}{
            \begin{tabular}{l|c|c|c}
                \bf \multirow{3}{*}{\bf Model} & \multicolumn{3}{c}{\bf Video Dimensions} \\
                 & \multirow{2}{*}{$\mathbf{\vTime \times \zHeight \times \zWidth}$} & \bf Frame & \bf Duration \\
                 &  & \bf Rate & \bf (s) \\
                \midrule
                \quad {\Ours } & $65 \times 512 \times 512$ & 16 & 4.06 \\
                \quad {Pika} & $72 \times 768 \times 768$ & 24 & 3.00 \\
                \quad {Gen2 } & $96 \times 1024 \times 1792$ & 24 & 4.00 \\
                \quad {CogVideo } & $32 \times 480 \times 480$ & 8 & 4.00 \\
                \quad {Reuse \& Diffuse } & $29 \times 512 \times 512$ & 24 & 1.21 \\
                \quad {PYOCO } & $76 \times 1024 \times 1024$ & 16 & 4.75\\
                \quad {Align Your Latents } & $112 \times 1280 \times 2048$ & 30 & 3.73\\
                \quad {Imagen Video } & $128 \times 768 \times 1280$ & 24 & 5.33\\
                \quad {Make-A-Video } & $92 \times 1024 \times 1024$ & 24 & 3.83 \\
                \hline
                \quad {\vidcomp } & $16 \times 256 \times 256$ & 8 & 2 \\

            \end{tabular}
        }
    \end{center}
    \caption{
        \textbf{Video Dimensions.}
        The dimensions of the generated videos from \OURS and each of the prior work.
        The top and bottom part of the table shows the specifications of \TextToV and \ImageToV models respectively.
        Each of the prior works generates videos at different dimensions, making unbiased human evaluation a challenge.
    }
    \label{tab:app_video_dimensions}
\end{table}

%% file: tables/appendix_downsampling.tex
\begin{table}[!htb]
    \begin{center}
        \centering
        \resizebox{0.8\linewidth}{!}{
            \begin{tabular}{l|c|c|c}
                \bf \multirow{3}{*}{\bf Models Compared} & \multicolumn{3}{c}{\bf Dimensions after Postprocessing} \\
                 & \multirow{2}{*}{$\mathbf{\vTime \times \zHeight \times \zWidth}$} & \bf Frame & \bf Duration \\
                 & & \bf Rate & \bf (s) \\
                \midrule
                \quad {\Ours vs. Pika Labs } & $48 \times 512 \times 512$ & 16 & 3.00 \\
                \quad {\Ours vs. Gen2 } & $65 \times 512 \times 512$ & 16 & 4.06 \\
                \quad {\Ours vs. CogVideo } & $32 \times 480 \times 480$ & 8 & 4.00 \\
                \quad {\Ours vs. Reuse \& Diffuse } & $19 \times 512 \times 512$ & 16 & 1.19 \\
                \quad {\Ours vs. PYOCO } & $65 \times 512 \times 512$ & 16 & 4.06\\
                \quad {\Ours vs. Align Your Latents } & $65 \times 512 \times 512$ & 16 & 4.06\\
                \quad {\Ours vs. Imagen Video } & $65 \times 512 \times 512$ & 16 & 4.06\\
                \quad {\Ours vs. Make-A-Video } & $61 \times 512 \times 512$ & 16 & 3.81 \\
                \hline
                \quad {\Ours vs. \vidcomp } & $16 \times 256 \times 256$ & 8 & 2 \\

            \end{tabular}
        }
    \end{center}
    \caption{
        \textbf{Video Dimensions after postprocessing for human evaluations.}.
        To ensure fairness in the human evaluations in in~\cref{fig:teaser_compare} in the main paper, we postprocess the videos for each comparison so that they have equal dimensions and hence are indistinguishable aside from their generated content.
        The top and bottom part of the table shows the specifications of \TextToV and \ImageToV models respectively.
    }
    \label{tab:app_downsampling_settings}
\end{table}

%% file: tables/appendix_sota_human_eval.tex
\begin{table*}[!t]
    \setlength{\tabcolsep}{4pt}
    \centering
    \resizebox{\textwidth}{!}{
    \begin{tabular}{l| cccccccc}
  & \bf \mav & \bf \imagenvideo & \bf \ayol & \bf \pyoco & \bf \reusediffuse & \bf \cogvideo & \bf \gen & \bf \pika \\
    \hline
    \#Prompts  & $307$~\cite{singer2023makeavideo} & $55$~\cite{ho2022imagen} & $65$~\cite{Blattmann2023AlignYL} & $74$~\cite{ge2023preserve} & $23$~\cite{gu2023reuse}  & $65$~\cite{Blattmann2023AlignYL} & $65$~\cite{Blattmann2023AlignYL} & $65$~\cite{Blattmann2023AlignYL}\\
    \hline
    \quality &
    96.8 & 90.9 & 96.9 & 93.2 & 95.7 & 100.0 & 83.1 & 93.9 %
    \\
    \faithfulness  &
    86.0 & 69.1 & 90.8 & 89.2 & 100.0 & 100.0 & 98.5 & 100.0 %
    \end{tabular}}
    \caption{\textbf{\OURS \vs prior work where videos are not postprocessed.}
    We evaluate \textToV generation in terms of video quality and text faithfulness win-rates evaluated by the majority votes of human evaluators for \OURS vs. Prior work methods.
    We compare methods here with their original dimensions (aspect ratio, duration, frame rate).
    \OURS significantly outperforms all prior work across all settings and metrics.
    }
    \label{tab:appendix_sota_human_eval}
\end{table*}

%% file: figures/Appendix_Qual_Compare_Figures/Our_generations_T2V/our_gen_figure.tex
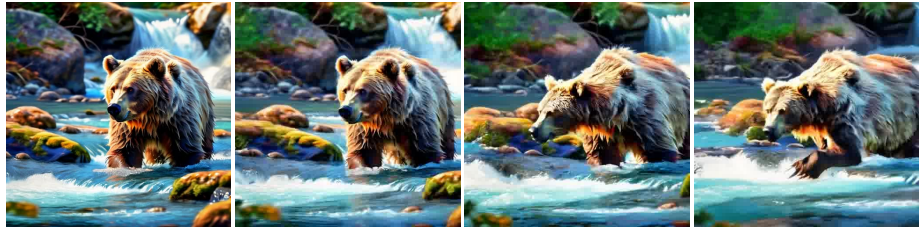
\begin{figure}
    \centering
    \captionsetup{type=figure}
    \vspace{-1cm}
    \input{figures/Appendix_Qual_Compare_Figures/Our_generations_T2V/figure.tex}
    \captionof{figure}{
    Example \textToVShort generations from \OURS for a selection of diverse prompts (shown above each row of frames).
    \OURS generates natural-looking videos which are faithful to the text and high in visual quality.
    The videos are highly temporally consistent, with smooth motion.
    \OURS is able to generate high quality videos for both natural prompts (rows 2 and 4) depicting scenes from the natural world, and also fantasical prompts including DJing hamsters (row 1) and underwater unicorns (row 3).
    }
\label{fig:our_gens_t2v}
\end{figure}

%% file: figures/Appendix_Qual_Compare_Figures/Our_generations_T2V/figure.tex
\setlength{\tabcolsep}{1pt}
\resizebox{\linewidth}{!}{%

    \begin{tabular}{cccc}
        \multicolumn{4}{l}{ \scriptsize {(Ours - \OURS) \it Prompt:} A hamster wearing virtual reality headsets is a dj in a disco.}  \\
        \includegraphics[width=0.27\linewidth]{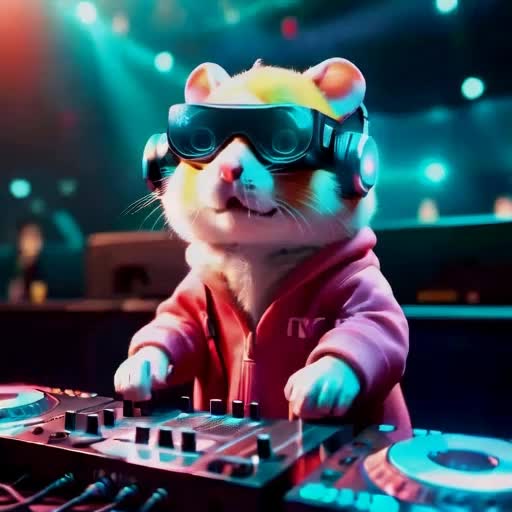} &
        \includegraphics[width=0.27\linewidth]{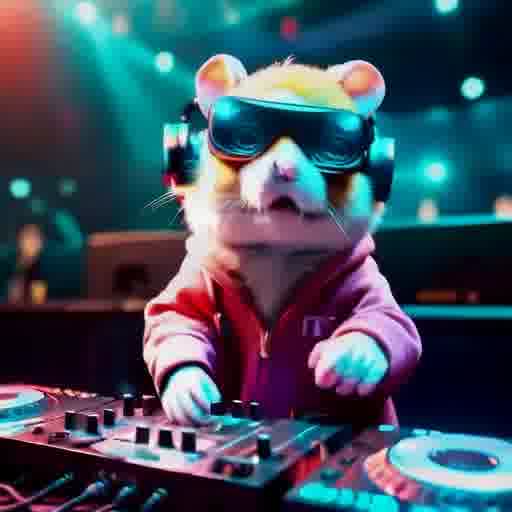} &
        \includegraphics[width=0.27\linewidth]{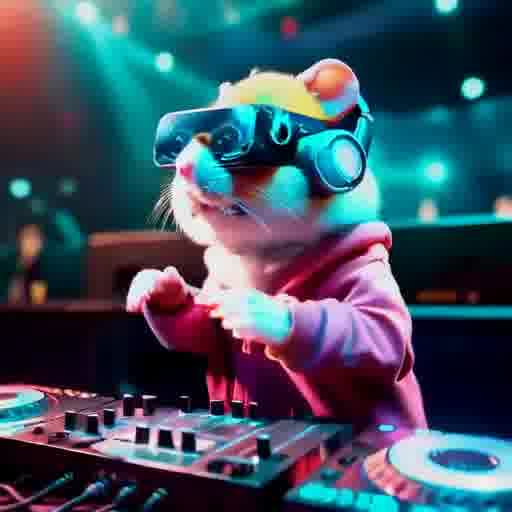} &
        \includegraphics[width=0.27\linewidth]{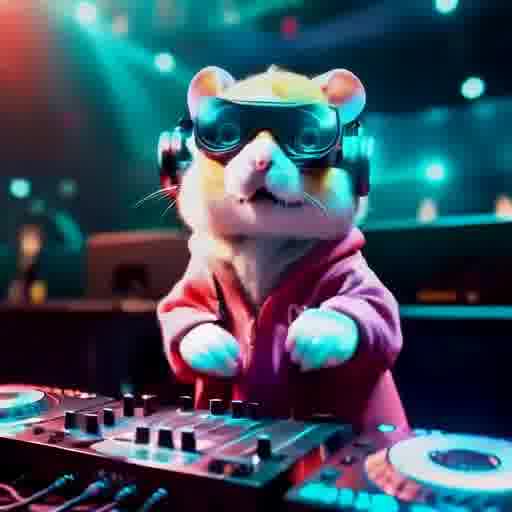} \\
        \multicolumn{4}{l}{ \scriptsize {(\OURS) \it Prompt:} A massive tidal wave crashes dramatically against a rugged coastline.}  \\
        \includegraphics[width=0.27\linewidth]{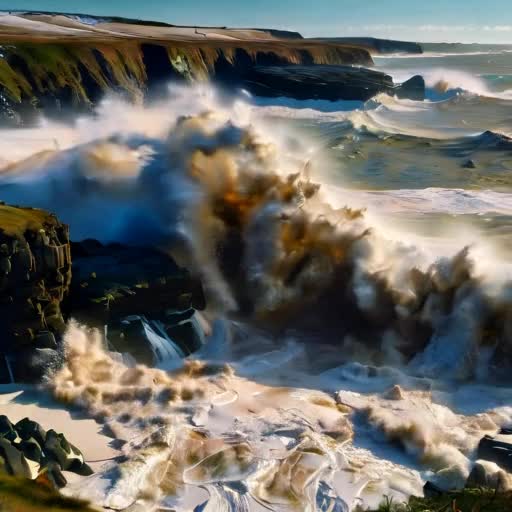} &
        \includegraphics[width=0.27\linewidth]{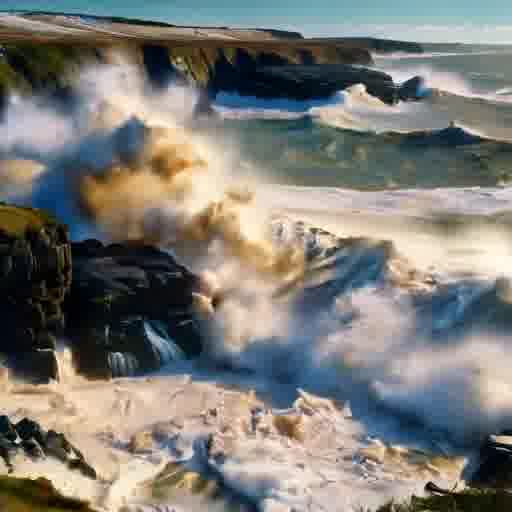} &
        \includegraphics[width=0.27\linewidth]{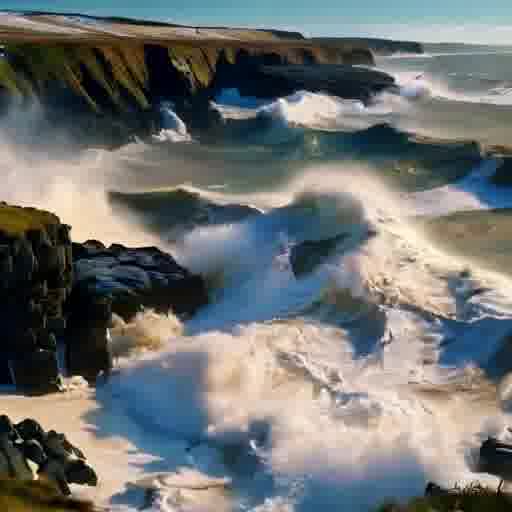} &
        \includegraphics[width=0.27\linewidth]{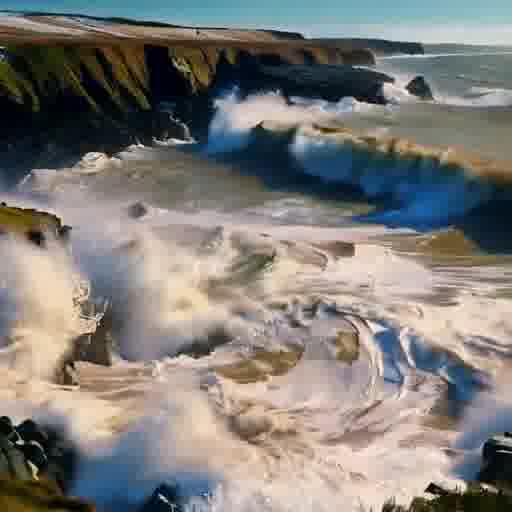} \\
        \multicolumn{4}{l}{ \scriptsize {(\OURS) \it Prompt:} A majestic white unicorn with a golden horn walking in slow-motion under water.}  \\
        \includegraphics[width=0.27\linewidth]{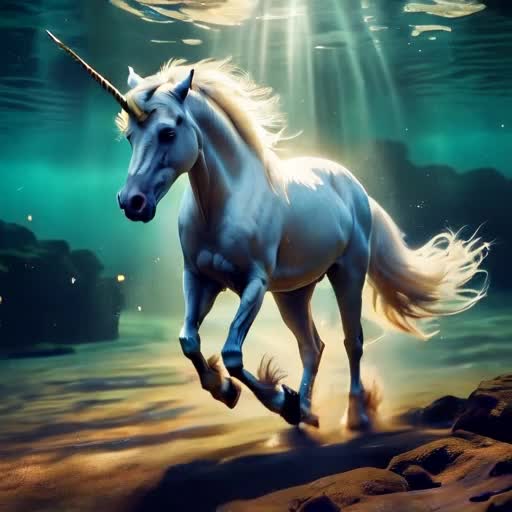} &
        \includegraphics[width=0.27\linewidth]{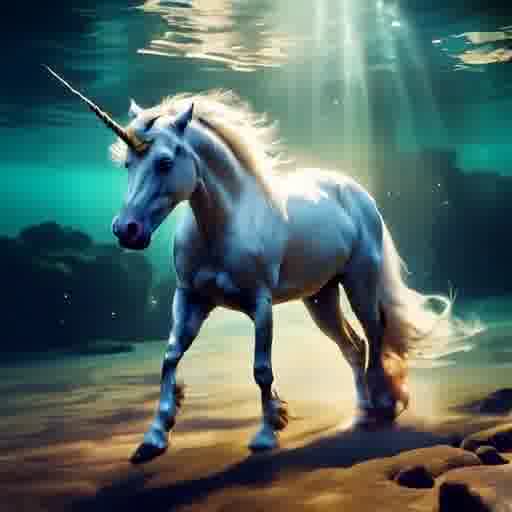} &
        \includegraphics[width=0.27\linewidth]{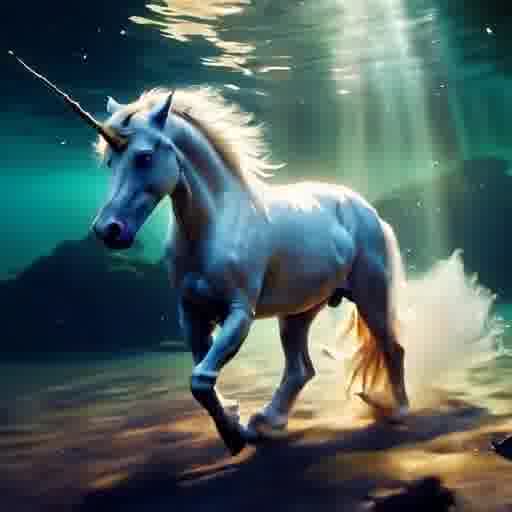} &
        \includegraphics[width=0.27\linewidth]{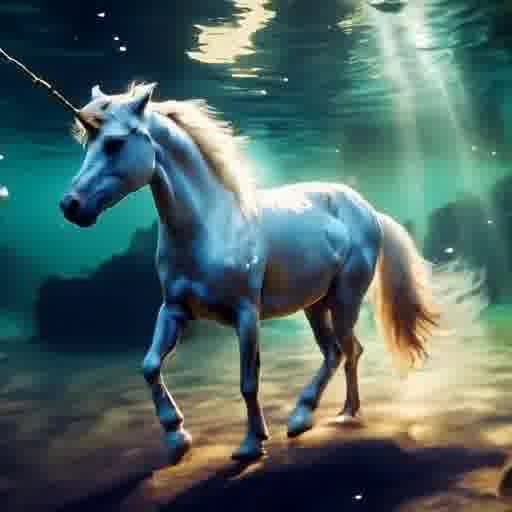} \\
        \multicolumn{4}{l}{ \scriptsize {(\OURS) \it Prompt:}  A grizzly bear hunting for fish in a river at the edge of a waterfall, photorealistic.}  \\
        \includegraphics[width=0.27\linewidth]{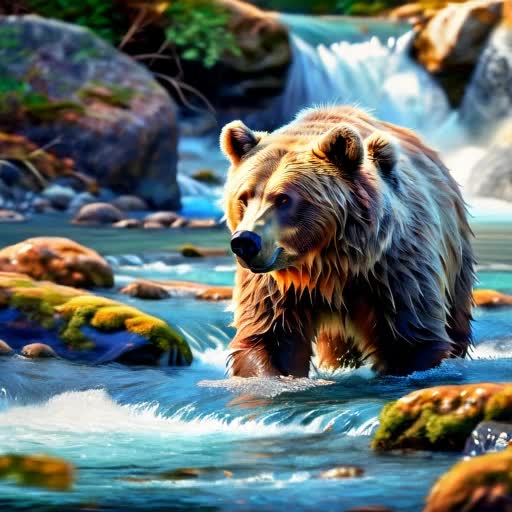} &
        \includegraphics[width=0.27\linewidth]{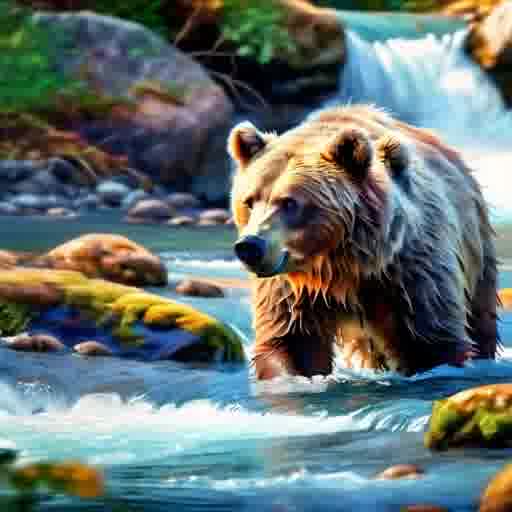} &
        \includegraphics[width=0.27\linewidth]{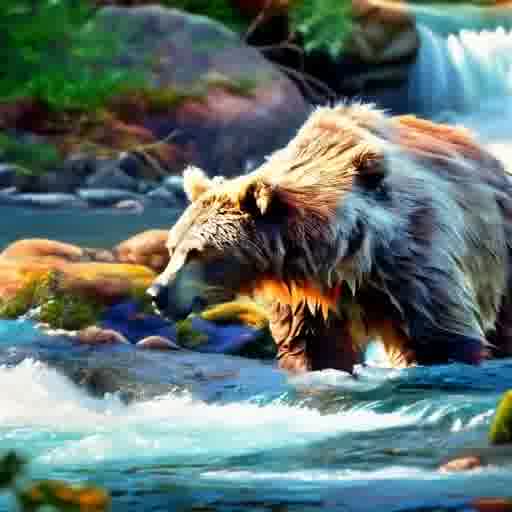} &
        \includegraphics[width=0.27\linewidth]{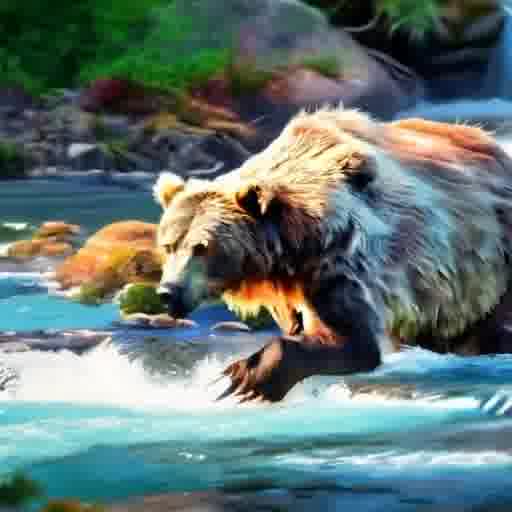} \\

    \end{tabular}%
}

%% file: figures/Appendix_Qual_Compare_Figures/Our_generations_I2V/our_gen_figure.tex
\begin{figure}
    \centering
    \captionsetup{type=figure}
    \vspace{-1cm}
    \input{figures/Appendix_Qual_Compare_Figures/Our_generations_I2V/figure.tex}
    \captionof{figure}{
    Example \imageToVShort generations from \OURS for a selection of diverse prompts (shown above each row of frames).
    \OURS generates natural-loooking videos from the conditioning image (shown in a blue box on the left side of each row of frames) and the text prompt, that have smooth and consistent motion.
    }
\label{fig:our_gens_i2v}
\end{figure}
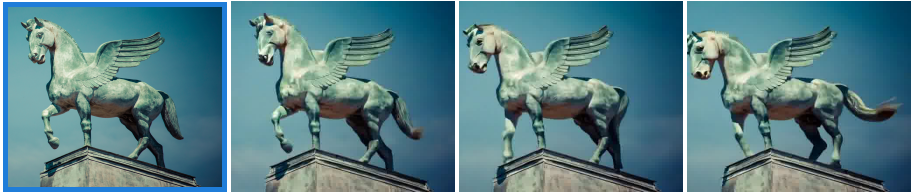%

%% file: figures/Appendix_Qual_Compare_Figures/Our_generations_I2V/figure.tex
\setlength{\tabcolsep}{1pt}
\resizebox{\linewidth}{!}{%

    \begin{tabular}{cccc}
        \multicolumn{4}{l}{ \scriptsize {(Ours - \OURS) \it Prompt:} The American flag waving during the moon landing with the camera panning.}  \\
        \setlength{\fboxrule}{2pt}
        \setlength{\fboxsep}{0pt}
        \hspace{-2\fboxrule-2\fboxsep}\raisebox{\fboxrule}{\fcolorbox{eccvblue}{white}{%
            \includegraphics[width=0.27\linewidth-2\fboxsep-2\fboxrule,height=0.23\linewidth-2\fboxsep-2\fboxrule]{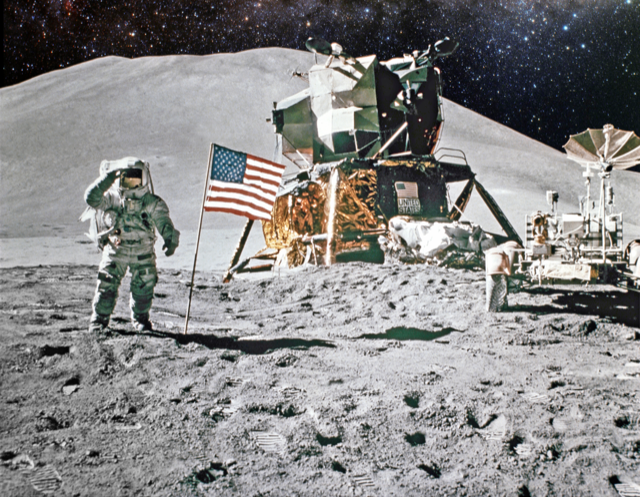}}%
        }
        &
        \includegraphics[width=0.27\linewidth,height=0.23\linewidth]{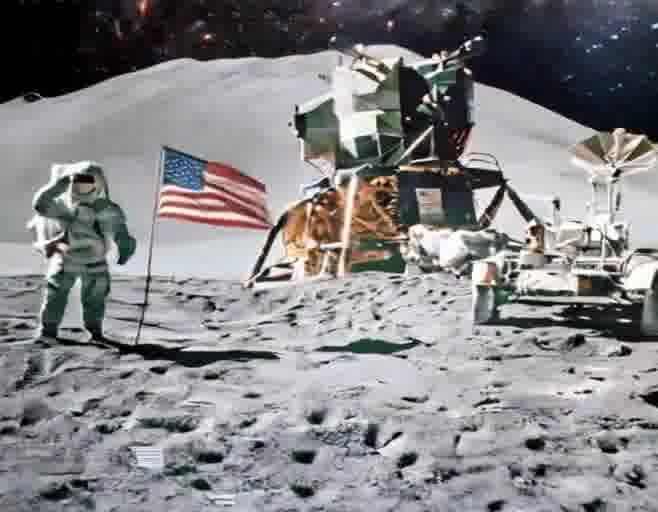} &
        \includegraphics[width=0.27\linewidth,height=0.23\linewidth]{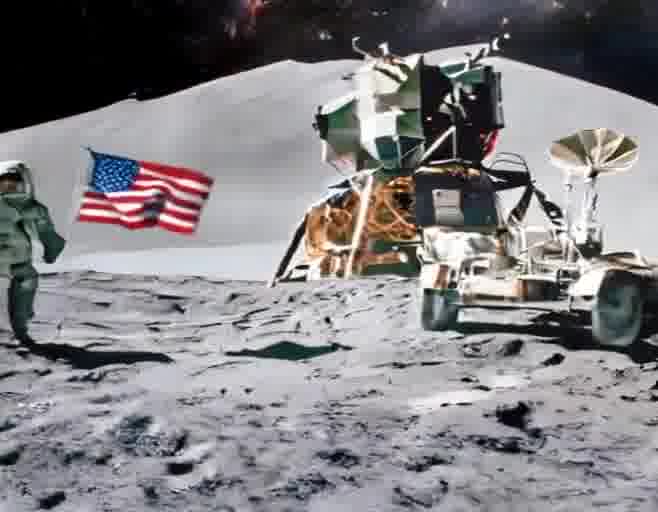} &
        \includegraphics[width=0.27\linewidth,height=0.23\linewidth]{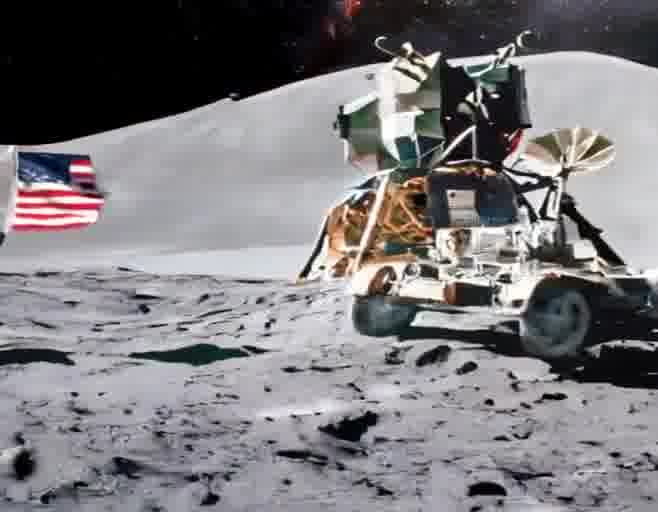} \\
        \multicolumn{4}{l}{ \scriptsize {(\OURS) \it Prompt:} The sun sets and the moon rises.}  \\
        \setlength{\fboxrule}{2pt}
        \setlength{\fboxsep}{0pt}
        \hspace{-2\fboxrule-2\fboxsep}\raisebox{\fboxrule}{\fcolorbox{eccvblue}{white}{%
            \includegraphics[width=0.27\linewidth-2\fboxsep-2\fboxrule,height=0.23\linewidth-2\fboxsep-2\fboxrule]{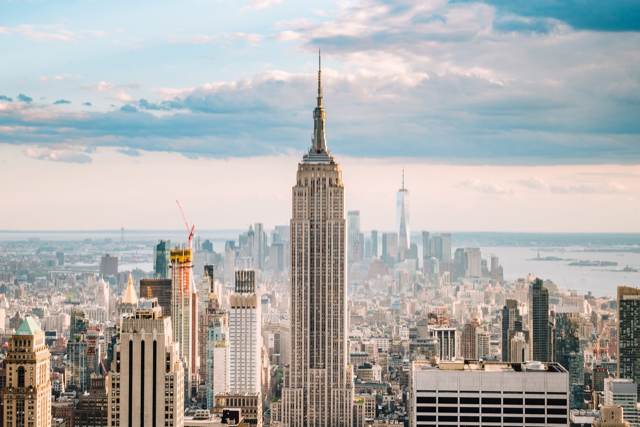}}%
        }
        &
        \includegraphics[width=0.27\linewidth,height=0.23\linewidth]{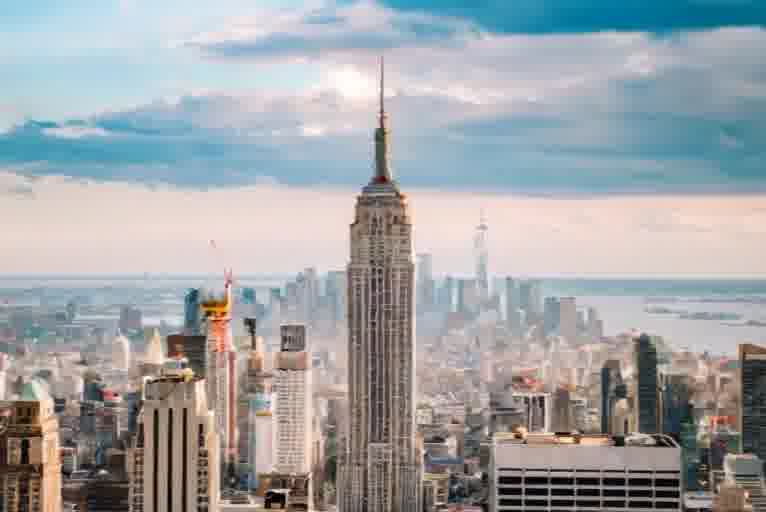} &
        \includegraphics[width=0.27\linewidth,height=0.23\linewidth]{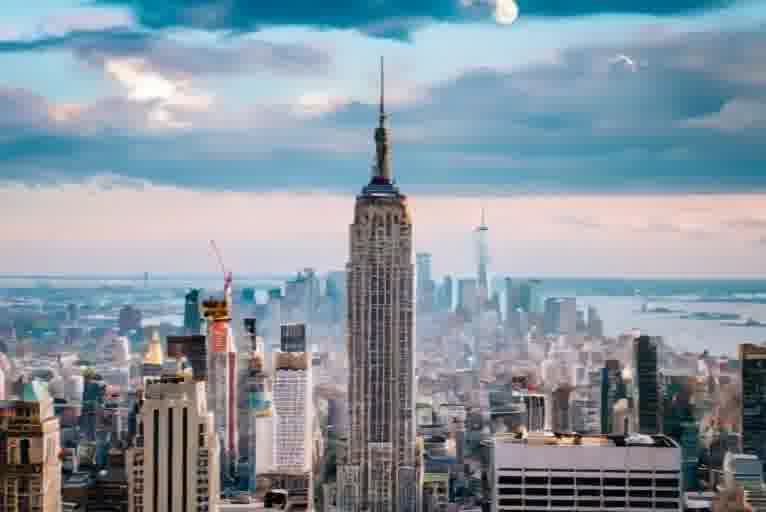} &
        \includegraphics[width=0.27\linewidth,height=0.23\linewidth]{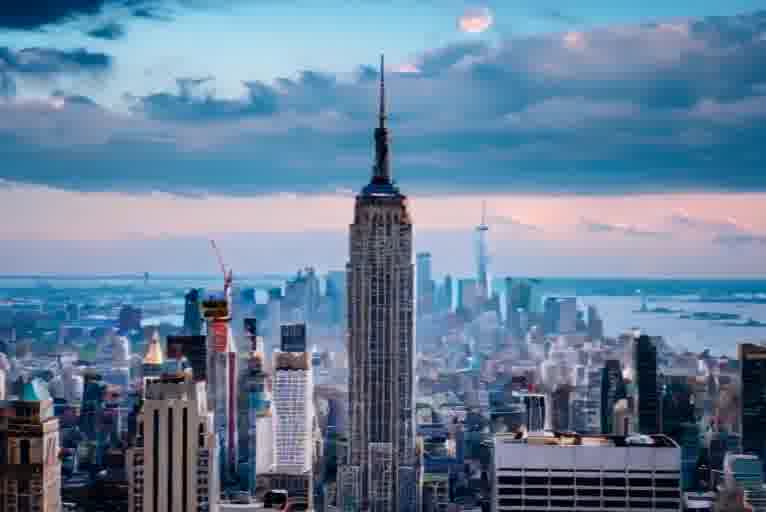} \\
        \multicolumn{4}{l}{ \scriptsize {(\OURS) \it Prompt:} Satellite flies across the globe.}  \\
        \setlength{\fboxrule}{2pt}
        \setlength{\fboxsep}{0pt}
        \hspace{-2\fboxrule-2\fboxsep}\raisebox{\fboxrule}{\fcolorbox{eccvblue}{white}{%
            \includegraphics[width=0.27\linewidth-2\fboxsep-2\fboxrule,height=0.23\linewidth-2\fboxsep-2\fboxrule]{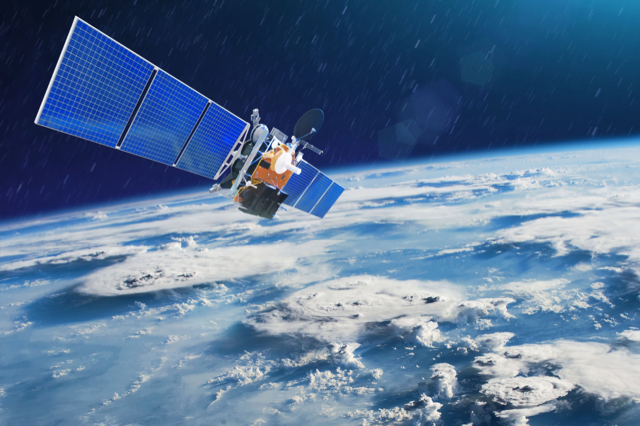}}%
        }
        &
        \includegraphics[width=0.27\linewidth,height=0.23\linewidth]{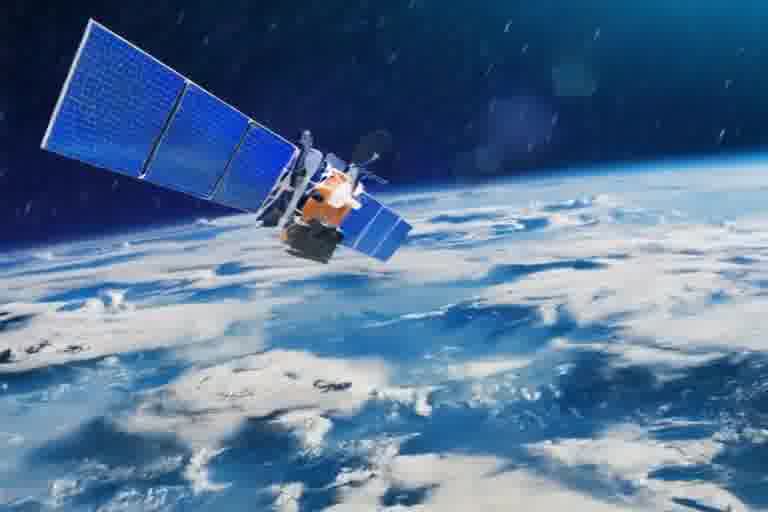} &
        \includegraphics[width=0.27\linewidth,height=0.23\linewidth]{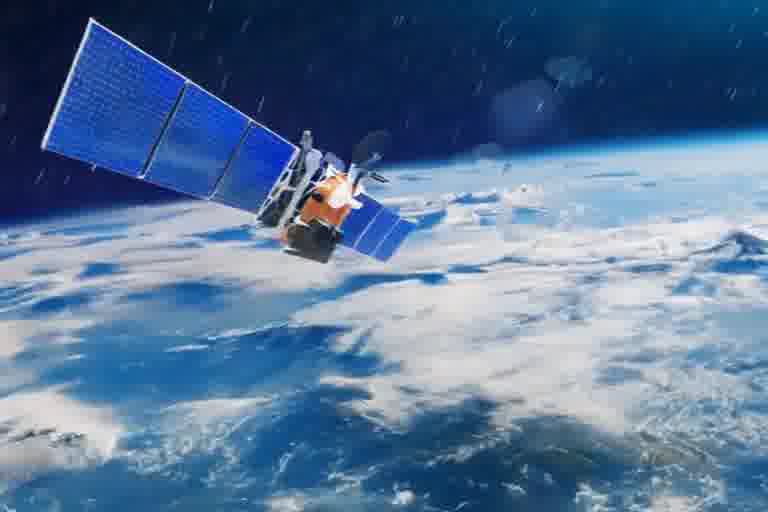} &
        \includegraphics[width=0.27\linewidth,height=0.23\linewidth]{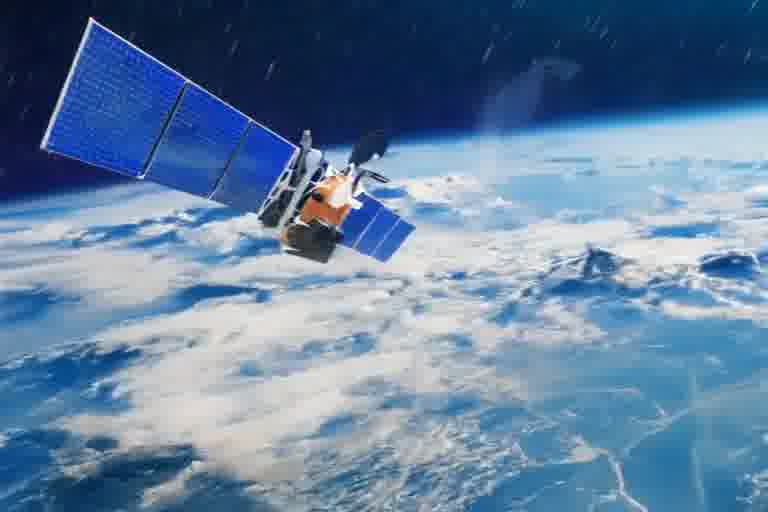} \\
        \multicolumn{4}{l}{ \scriptsize {(\OURS) \it Prompt:}  horse moving its legs.}  \\
        \setlength{\fboxrule}{2pt}
        \setlength{\fboxsep}{0pt}
        \hspace{-2\fboxrule-2\fboxsep}\raisebox{\fboxrule}{\fcolorbox{eccvblue}{white}{%
            \includegraphics[width=0.27\linewidth-2\fboxsep-2\fboxrule,height=0.23\linewidth-2\fboxsep-2\fboxrule]{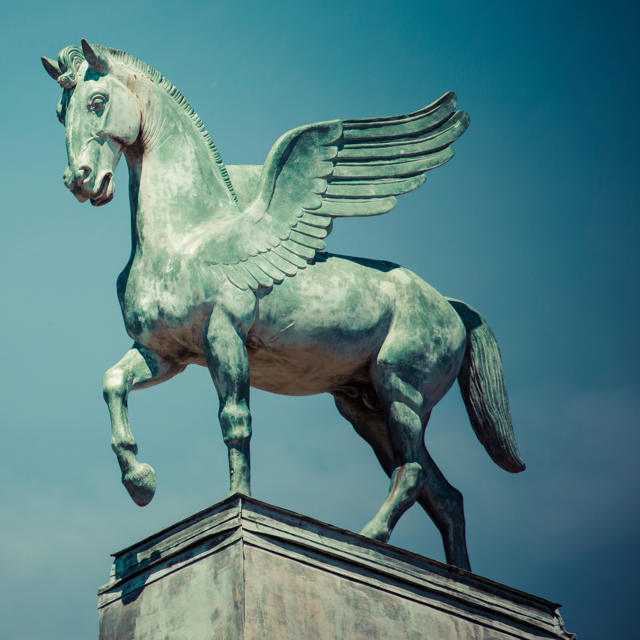}}%
        }
        &
        \includegraphics[width=0.27\linewidth,height=0.23\linewidth]{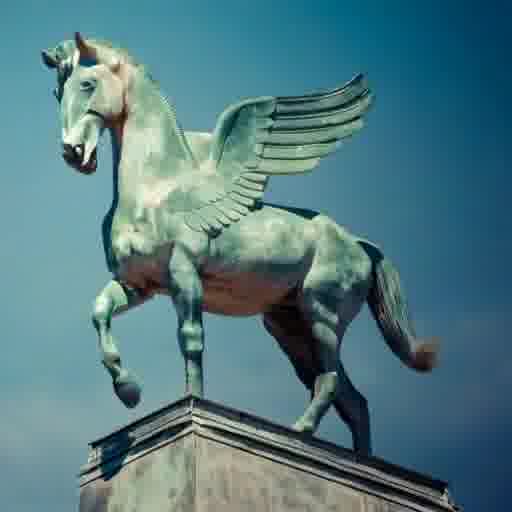} &
        \includegraphics[width=0.27\linewidth,height=0.23\linewidth]{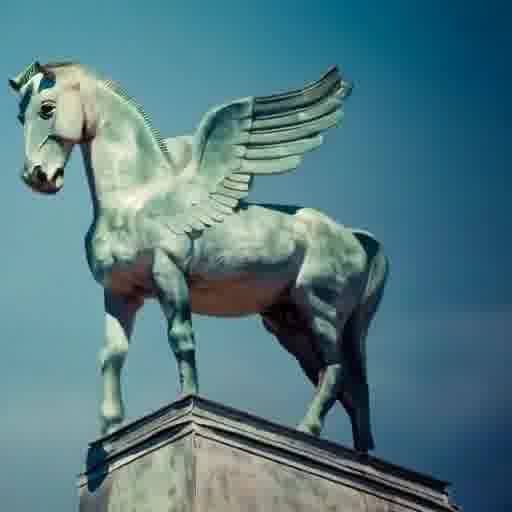} &
        \includegraphics[width=0.27\linewidth,height=0.23\linewidth]{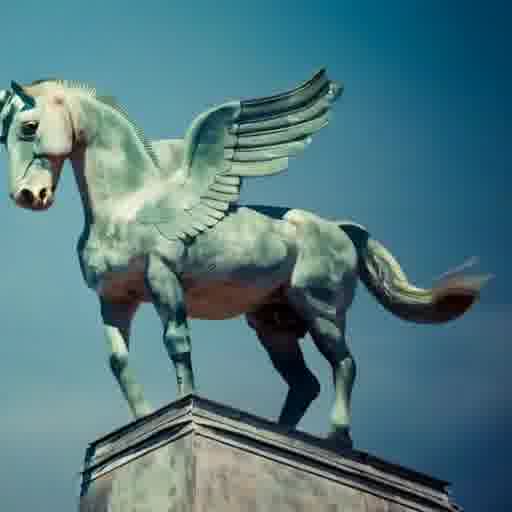} \\

    \end{tabular}%
}

%% file: figures/Appendix_Qual_Compare_Figures/Our_generations_prior_work_compare/our_gen_figure.tex
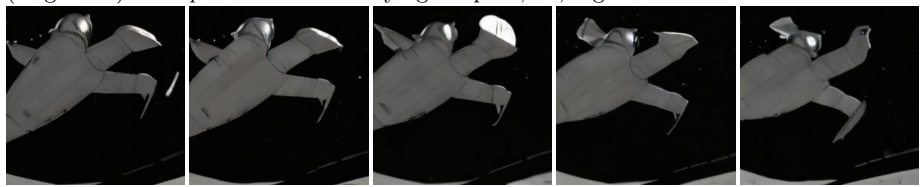
\begin{figure}
    \centering
    \captionsetup{type=figure}
    \vspace{-1cm}
    \input{figures/Appendix_Qual_Compare_Figures/Our_generations_prior_work_compare/figure.tex}
    \captionof{figure}{
    Example \textToVShort generations from \OURS and a selection of prior work methods that we compare to in the main paper for the same prompt, namely Gen2, Pika Labs, Align your latents, and CogVideo.
    \OURS generates higher quality videos that are more faithful to the text, have realistic \& smooth movement, and are visually compelling.
    In this example, \cogvideo cannot generate a natural-looking video (see 5th row).
    \pika is not faithful to the text and does not generate a realistic looking astronaut (see 3rd row), whereas \ayol generates a video with low visual quality.
    \gen's video, although visually superior to other prior work, lacks pixel sharpness and is not as visually compelling as \OURS.
    }
\label{fig:compare_to_prior_1}
\end{figure}%

%% file: figures/Appendix_Qual_Compare_Figures/Our_generations_prior_work_compare/figure.tex
\setlength{\tabcolsep}{1pt}
\resizebox{\linewidth}{!}{%
    \begin{tabular}{ccccc}
        \multicolumn{5}{l}{ \small {(Ours - \OURS) \it Prompt:} An astronaut flying in space, 4k, high resolution.}  \\

        \includegraphics[width=0.2\linewidth]{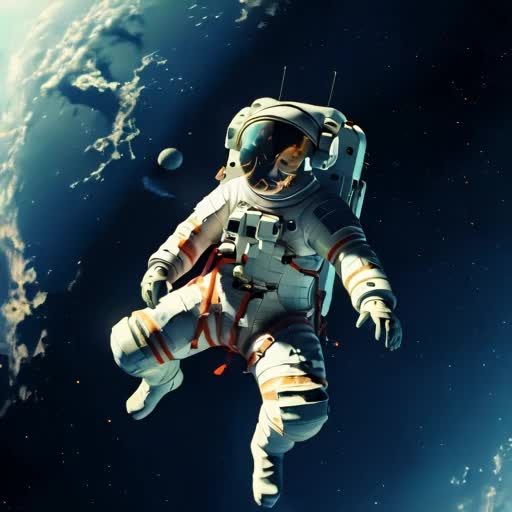} &
        \includegraphics[width=0.2\linewidth]{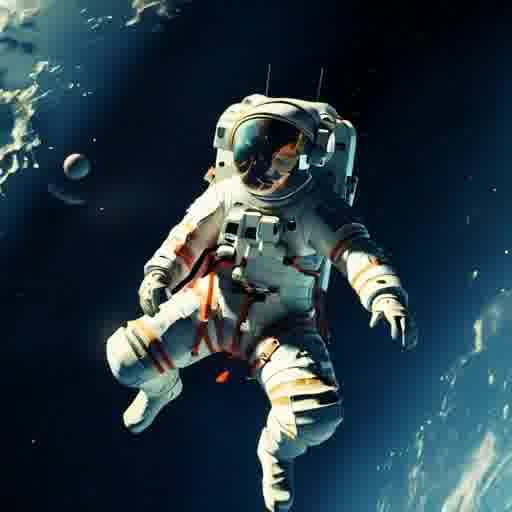} &
        \includegraphics[width=0.2\linewidth]{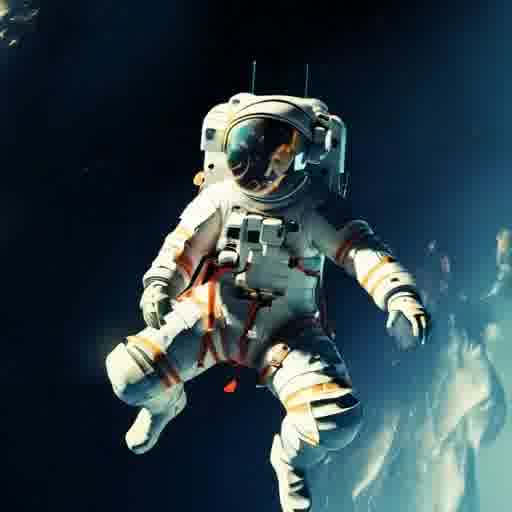} &
        \includegraphics[width=0.2\linewidth]{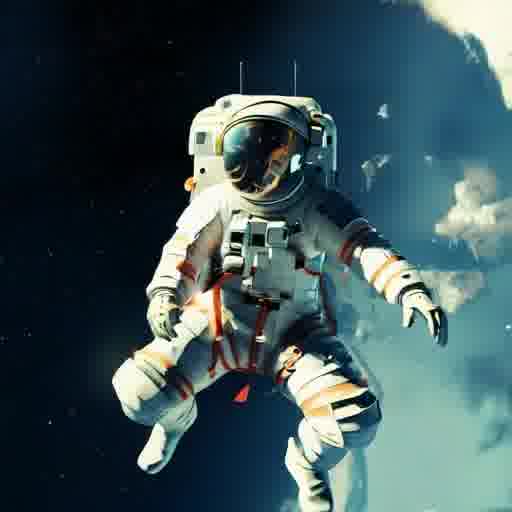} &
        \includegraphics[width=0.2\linewidth]{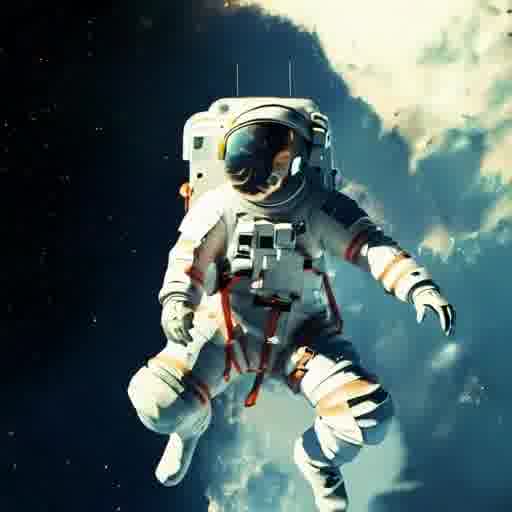} \\

        \multicolumn{5}{l}{ \small {(\gen) \it Prompt:} An astronaut flying in space, 4k, high resolution.}  \\
        \includegraphics[width=0.2\linewidth]{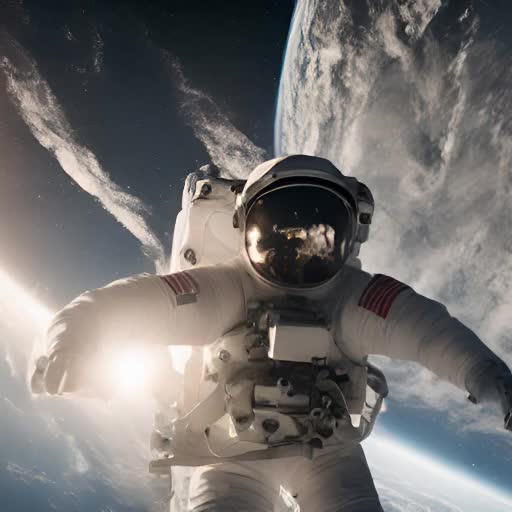} &
        \includegraphics[width=0.2\linewidth]{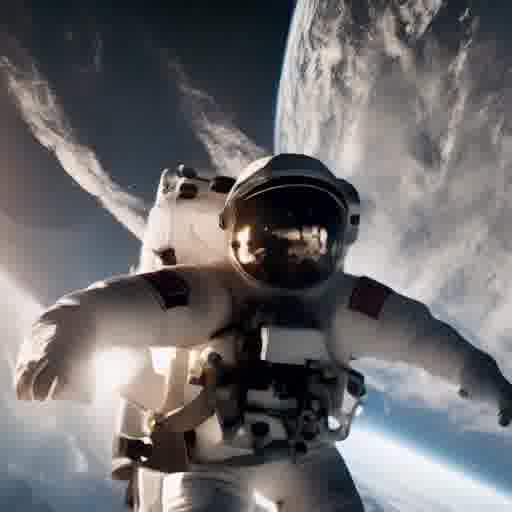} &
        \includegraphics[width=0.2\linewidth]{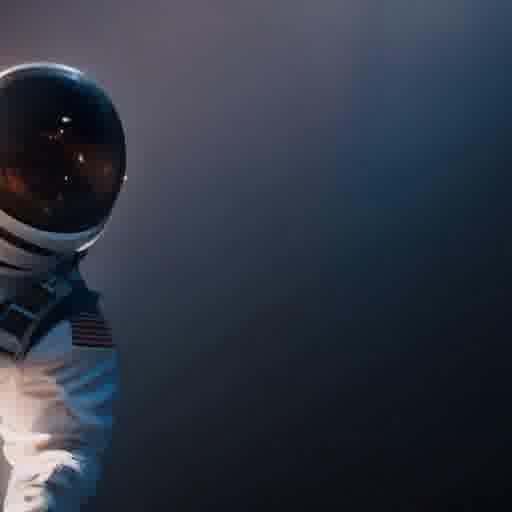} &
        \includegraphics[width=0.2\linewidth]{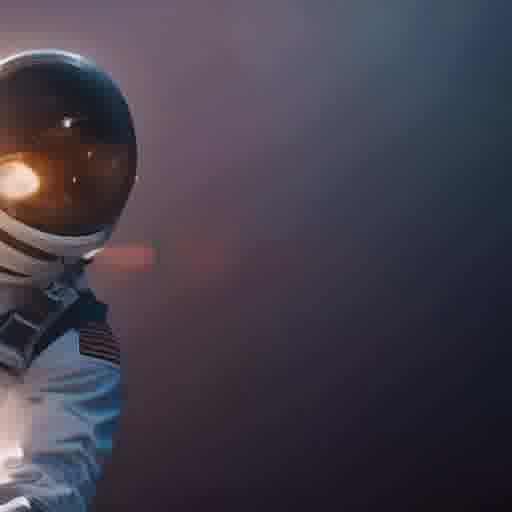} &
        \includegraphics[width=0.2\linewidth]{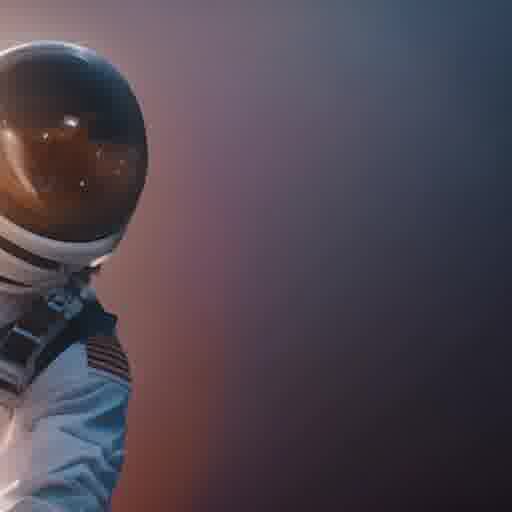} \\

        \multicolumn{5}{l}{ \small {(\pika) \it Prompt:} An astronaut flying in space, 4k, high resolution.}  \\
        \includegraphics[width=0.2\linewidth]{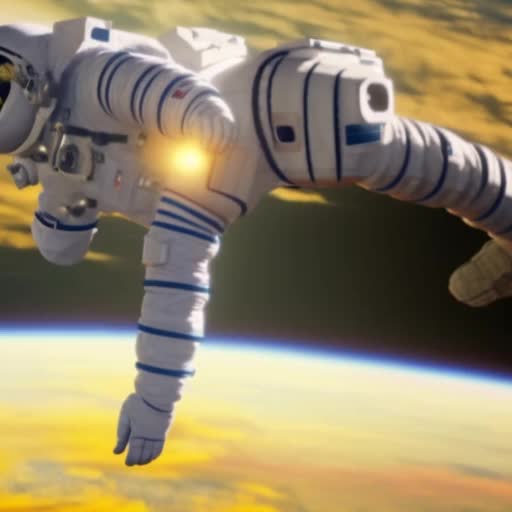} &
        \includegraphics[width=0.2\linewidth]{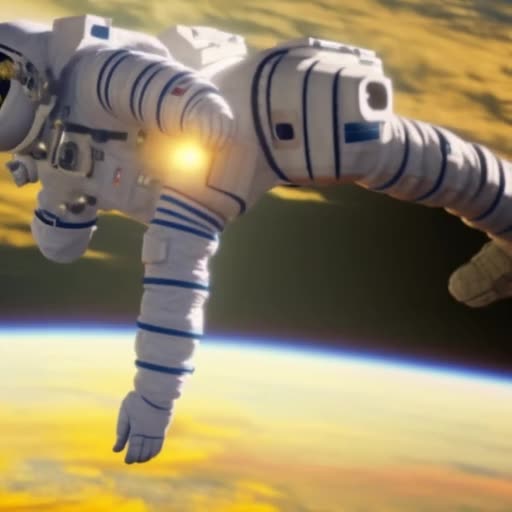} &
        \includegraphics[width=0.2\linewidth]{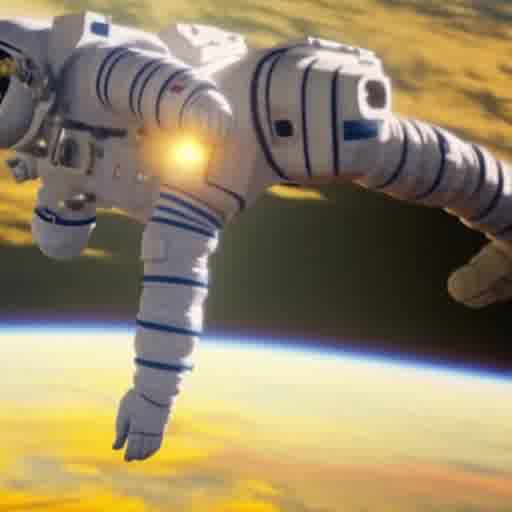} &
        \includegraphics[width=0.2\linewidth]{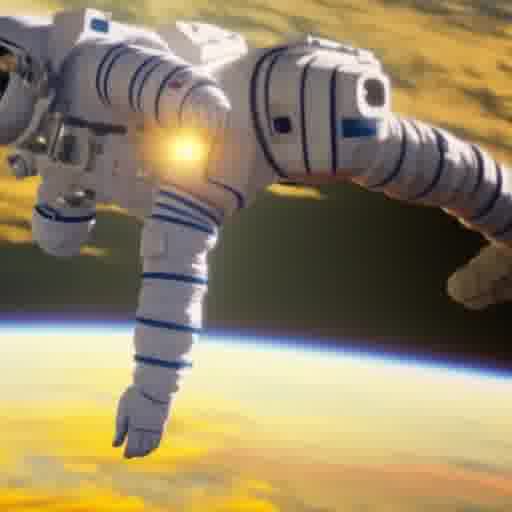} &
        \includegraphics[width=0.2\linewidth]{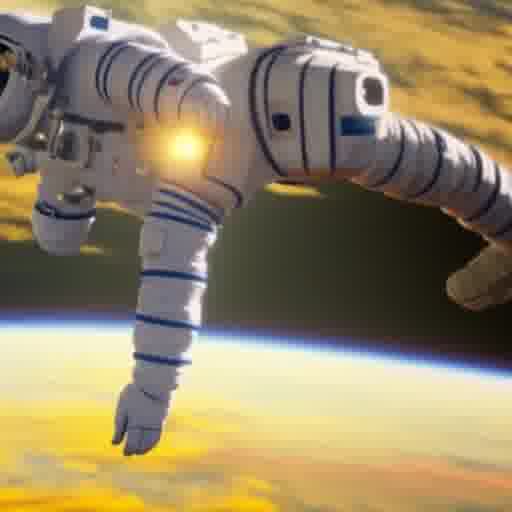} \\

        \multicolumn{5}{l}{ \small {(\ayol) \it Prompt:} An astronaut flying in space, 4k, high resolution.}  \\
        \includegraphics[width=0.2\linewidth]{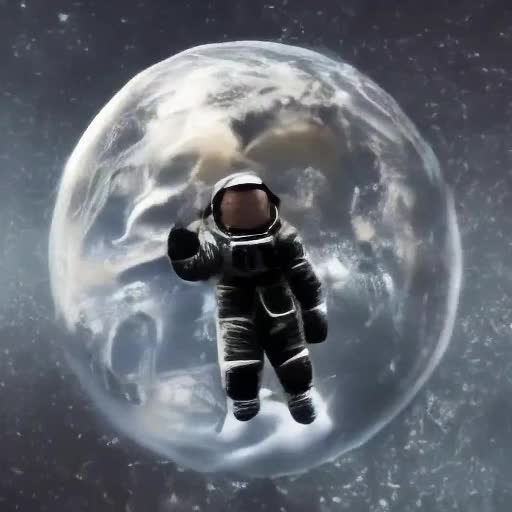} &
        \includegraphics[width=0.2\linewidth]{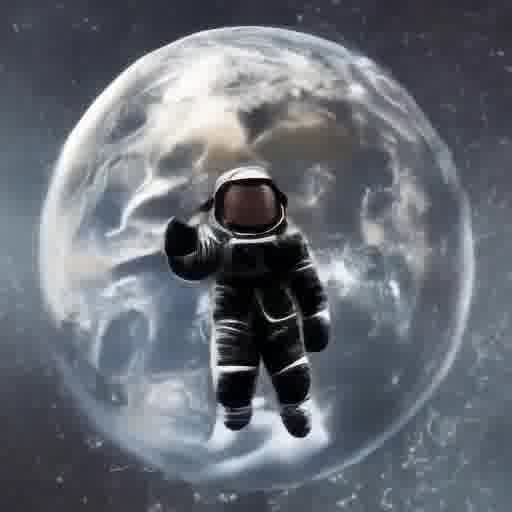} &
        \includegraphics[width=0.2\linewidth]{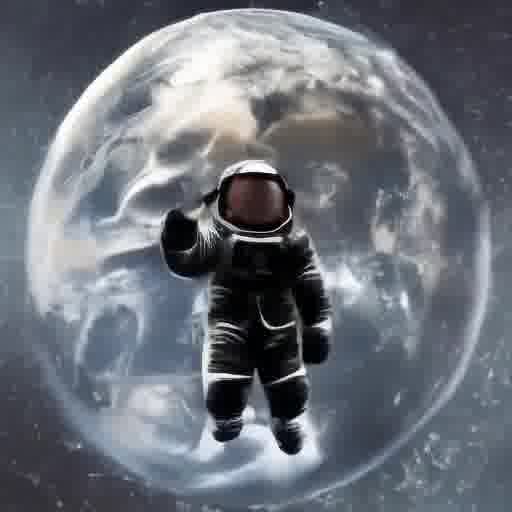} &
        \includegraphics[width=0.2\linewidth]{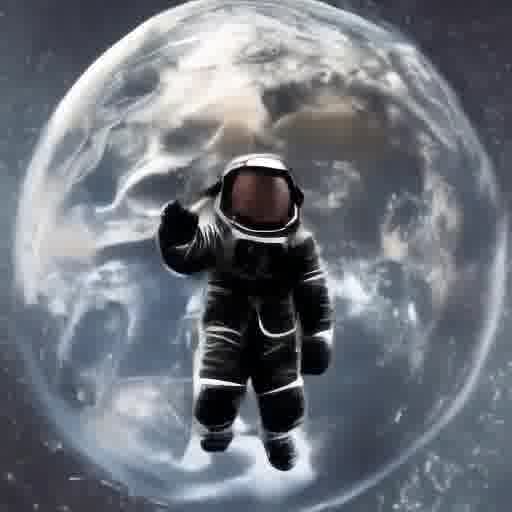} &
        \includegraphics[width=0.2\linewidth]{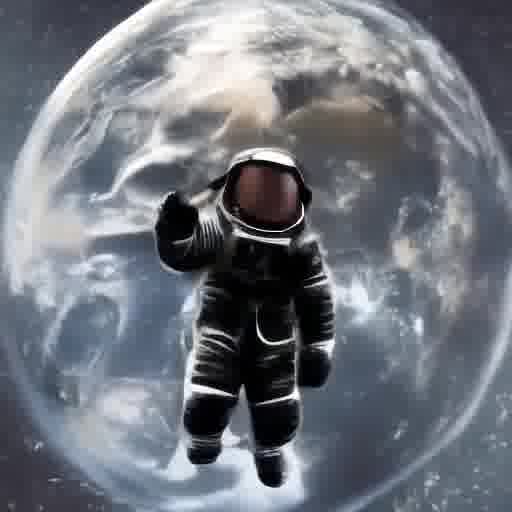} \\

        \multicolumn{5}{l}{ \small {(\cogvideo) \it Prompt:} An astronaut flying in space, 4k, high resolution.}  \\
        \includegraphics[width=0.2\linewidth]{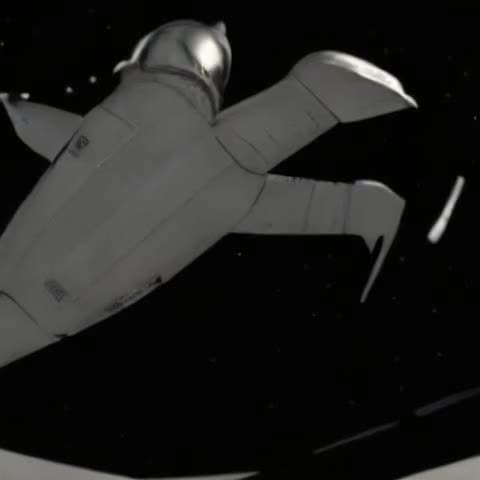} &
        \includegraphics[width=0.2\linewidth]{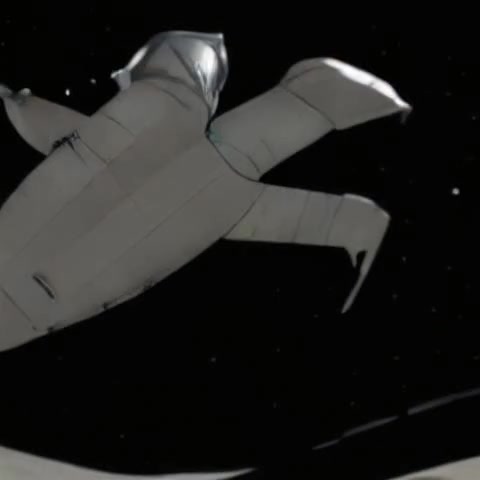} &
        \includegraphics[width=0.2\linewidth]{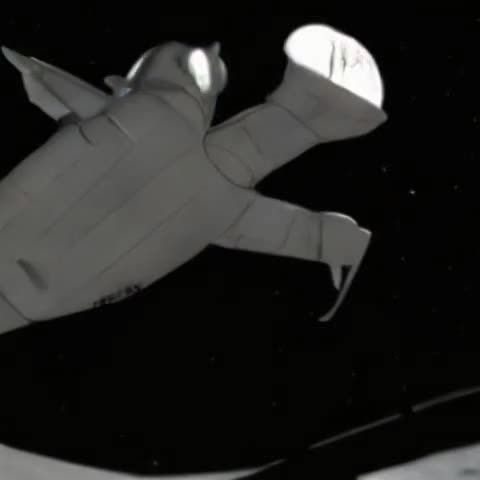} &
        \includegraphics[width=0.2\linewidth]{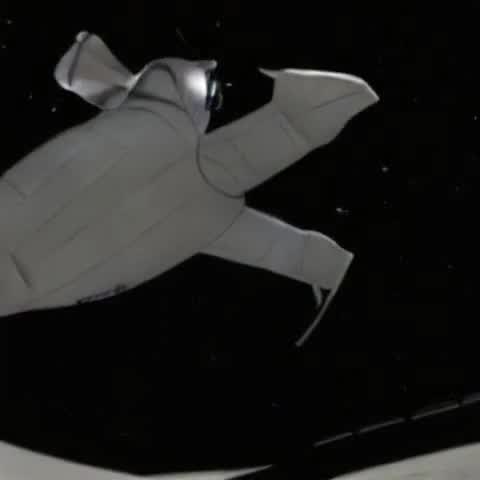} &
        \includegraphics[width=0.2\linewidth]{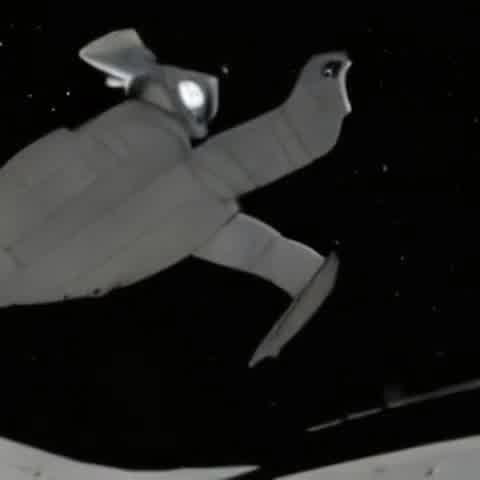} \\

    \end{tabular}%
}

%% file: figures/Appendix_Qual_Compare_Figures/Our_generations_prior_work_compare/our_gen_figure2.tex
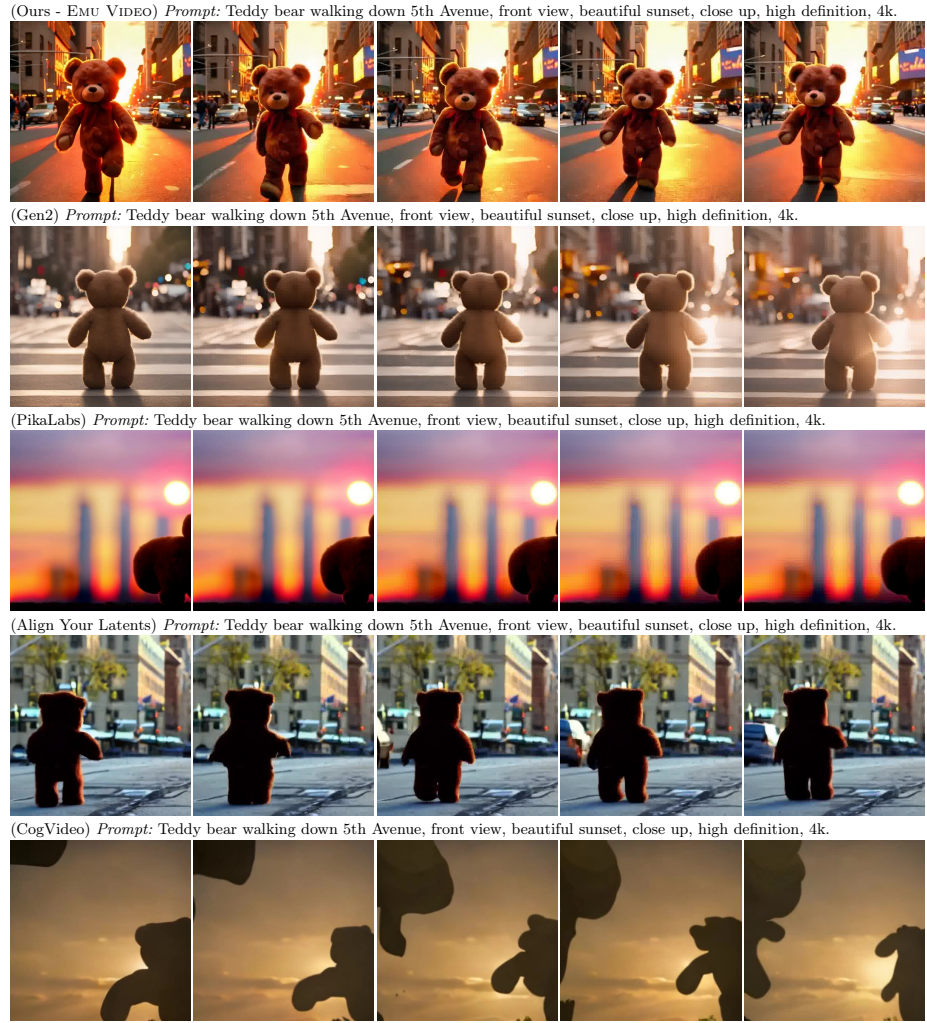
\begin{figure}
    \centering
    \captionsetup{type=figure}
    \vspace{-1cm}
    \input{figures/Appendix_Qual_Compare_Figures/Our_generations_prior_work_compare/figure2.tex}
    \captionof{figure}{
        Example \textToVShort generations from \OURS and a selection of prior work methods that we compare to in the main paper for the same prompt, namely Gen2, Pika Labs, Align your latents, and CogVideo.
        \cogvideo and \pika's videos are not faithful to the text and lack on visual quality.
        \gen correctly generates a video of a bear on a street, but the bear is not moving, and there is limited motion in the vidoeo.
        \ayol's video lacks motion smoothness and pixel sharpness.
        On the other had, \OURS's video has very high visual quality and high text faithfulness, with smooth and consistent high motion.
        }
\label{fig:compare_to_prior_2}
\end{figure}%

%% file: figures/Appendix_Qual_Compare_Figures/Our_generations_prior_work_compare/figure2.tex
\setlength{\tabcolsep}{1pt}
\resizebox{\linewidth}{!}{%
    \begin{tabular}{ccccc}
        \multicolumn{5}{l}{ \footnotesize {(Ours - \OURS) \it Prompt:} Teddy bear walking down 5th Avenue, front view, beautiful sunset, close up, high definition, 4k.}  \\
        \includegraphics[width=0.3\linewidth]{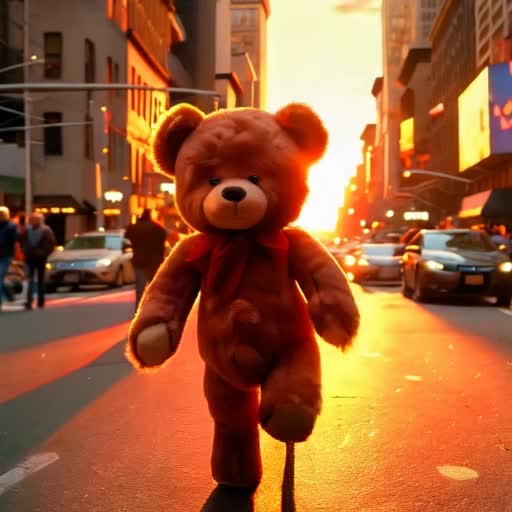} &
        \includegraphics[width=0.3\linewidth]{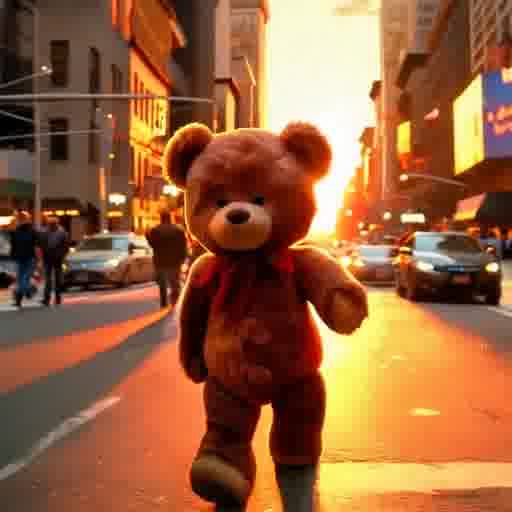} &
        \includegraphics[width=0.3\linewidth]{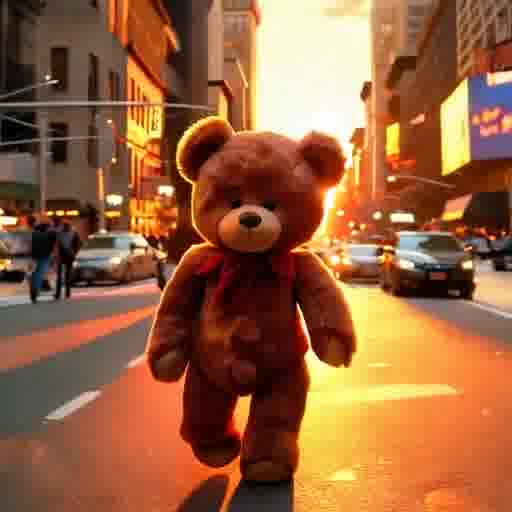} &
        \includegraphics[width=0.3\linewidth]{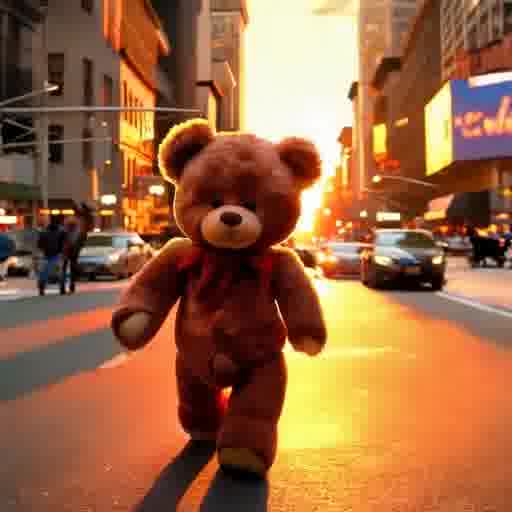} &
        \includegraphics[width=0.3\linewidth]{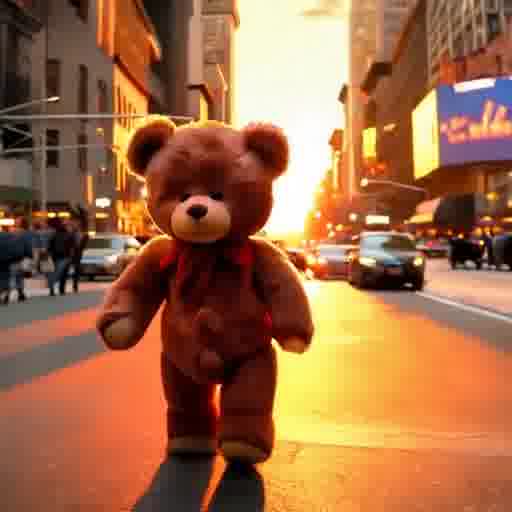} \\

        \multicolumn{5}{l}{ \footnotesize {(\gen) \it Prompt:} Teddy bear walking down 5th Avenue, front view, beautiful sunset, close up, high definition, 4k.}  \\
        \includegraphics[width=0.3\linewidth]{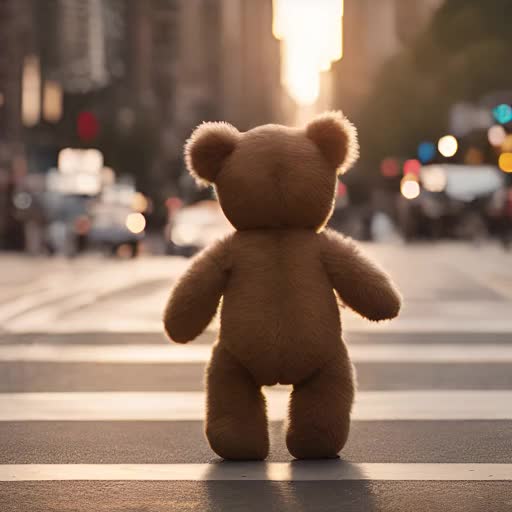} &
        \includegraphics[width=0.3\linewidth]{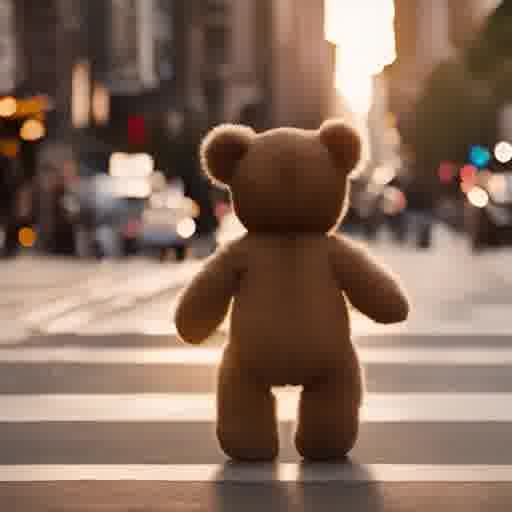} &
        \includegraphics[width=0.3\linewidth]{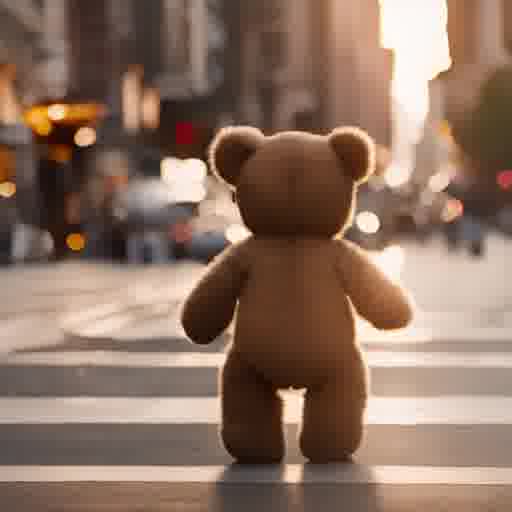} &
        \includegraphics[width=0.3\linewidth]{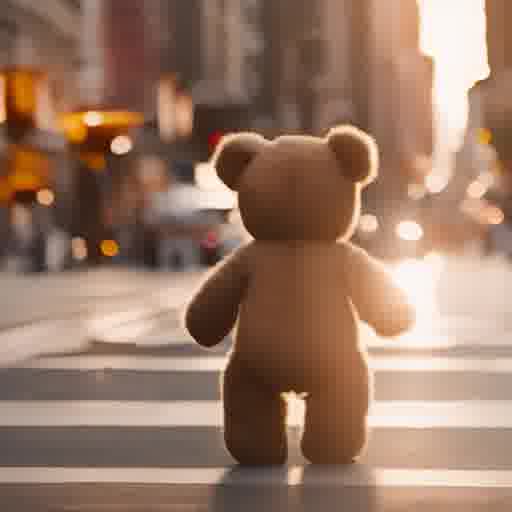} &
        \includegraphics[width=0.3\linewidth]{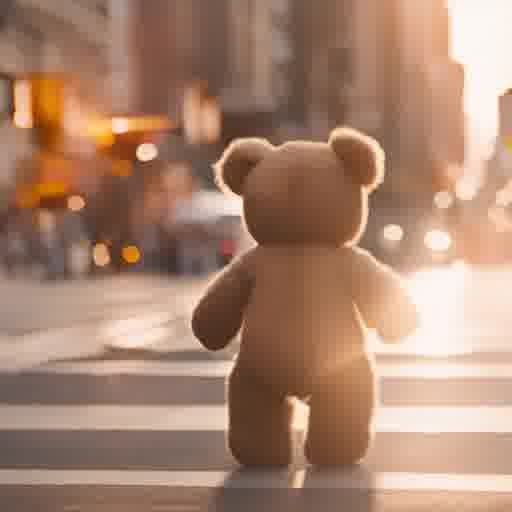} \\

        \multicolumn{5}{l}{ \footnotesize {(\pika) \it Prompt:} Teddy bear walking down 5th Avenue, front view, beautiful sunset, close up, high definition, 4k.}  \\
        \includegraphics[width=0.3\linewidth]{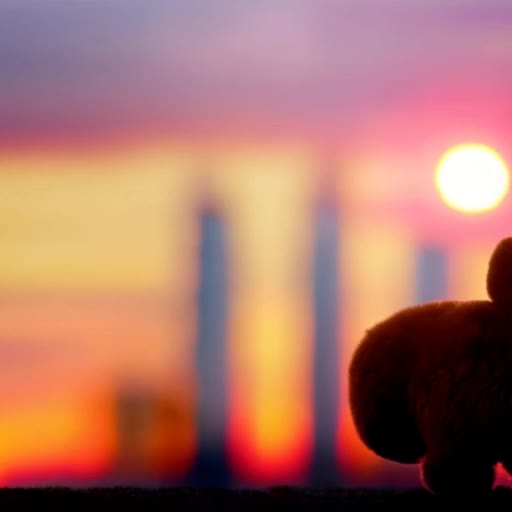} &
        \includegraphics[width=0.3\linewidth]{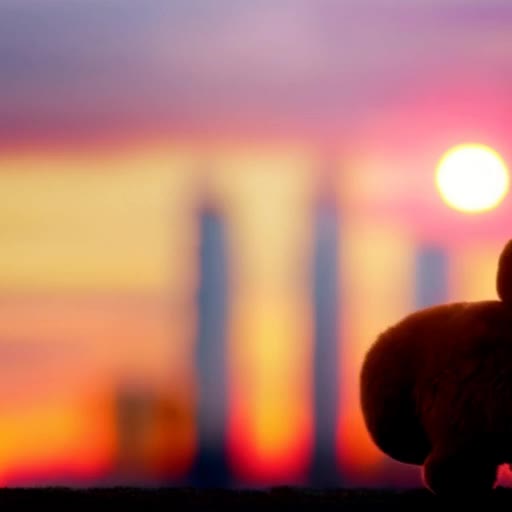} &
        \includegraphics[width=0.3\linewidth]{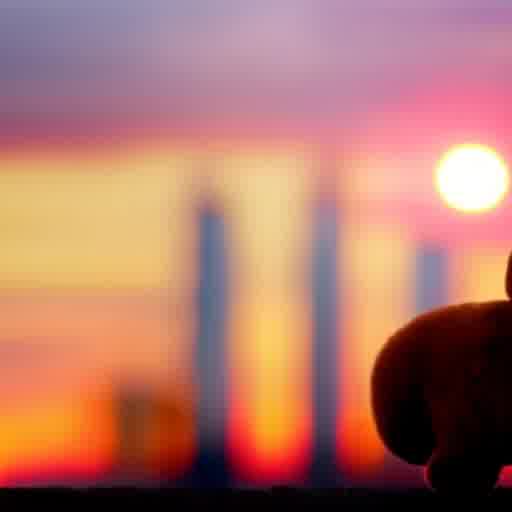} &
        \includegraphics[width=0.3\linewidth]{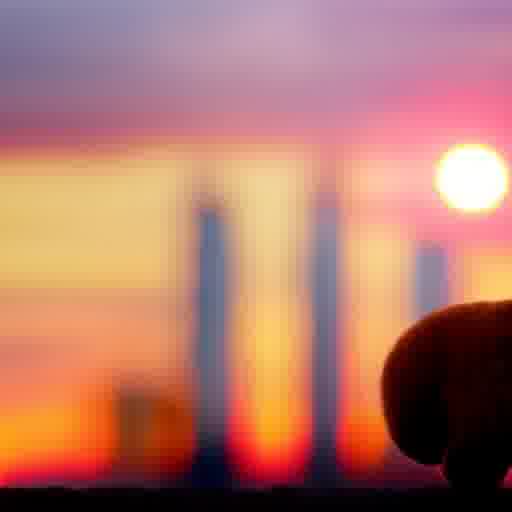} &
        \includegraphics[width=0.3\linewidth]{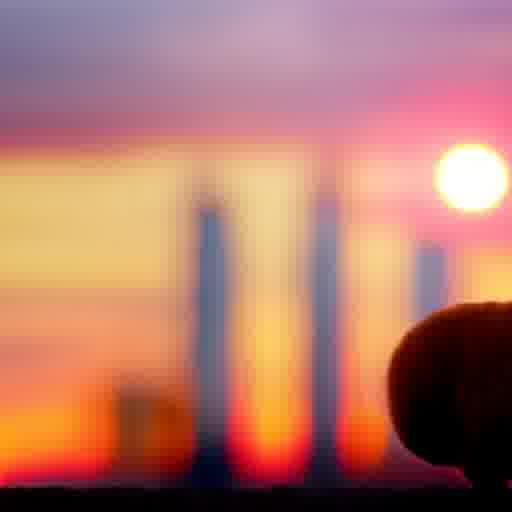} \\

        \multicolumn{5}{l}{ \footnotesize {(\ayol) \it Prompt:} Teddy bear walking down 5th Avenue, front view, beautiful sunset, close up, high definition, 4k.}  \\
        \includegraphics[width=0.3\linewidth]{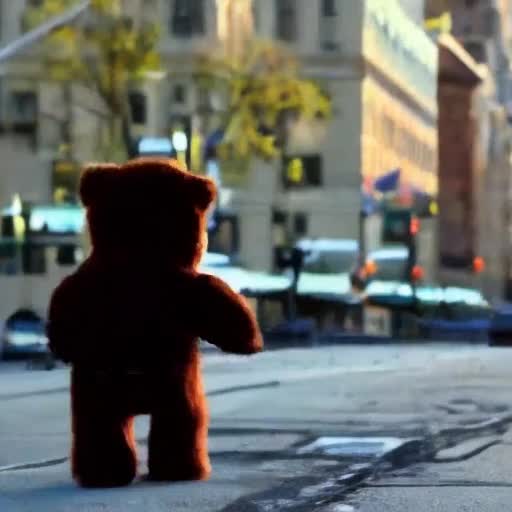} &
        \includegraphics[width=0.3\linewidth]{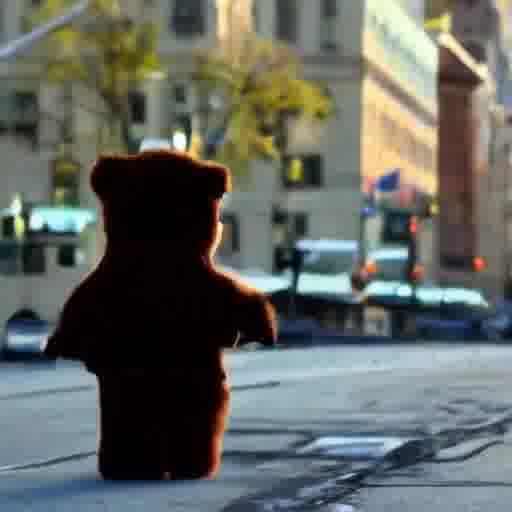} &
        \includegraphics[width=0.3\linewidth]{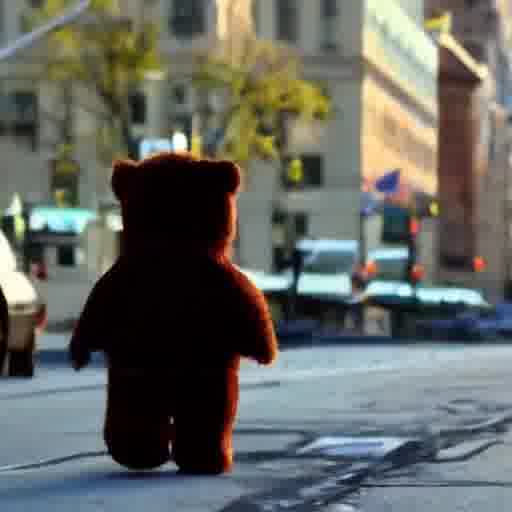} &
        \includegraphics[width=0.3\linewidth]{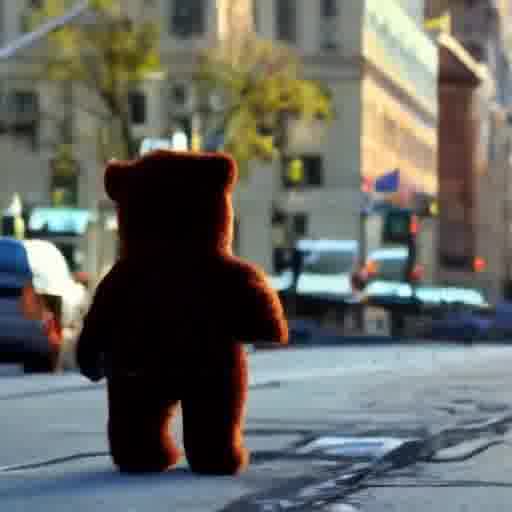} &
        \includegraphics[width=0.3\linewidth]{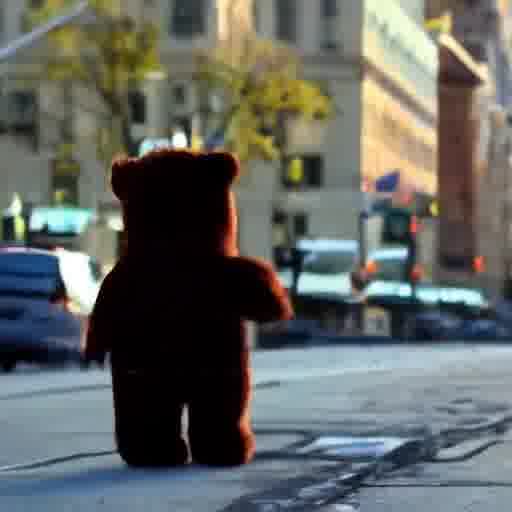} \\

        \multicolumn{5}{l}{ \footnotesize {(\cogvideo) \it Prompt:} Teddy bear walking down 5th Avenue, front view, beautiful sunset, close up, high definition, 4k.}  \\
        \includegraphics[width=0.3\linewidth]{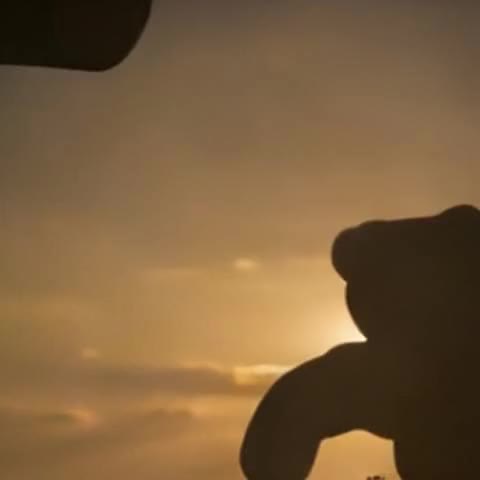} &
        \includegraphics[width=0.3\linewidth]{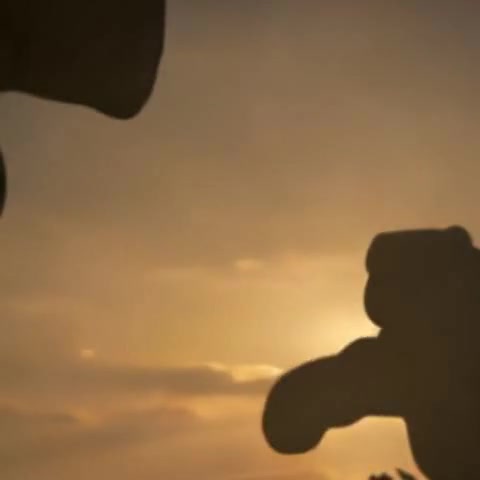} &
        \includegraphics[width=0.3\linewidth]{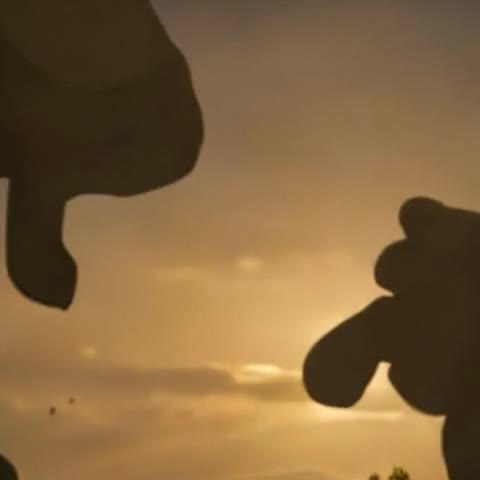} &
        \includegraphics[width=0.3\linewidth]{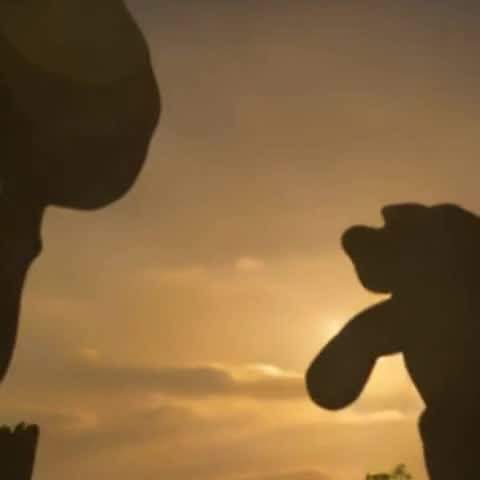} &
        \includegraphics[width=0.3\linewidth]{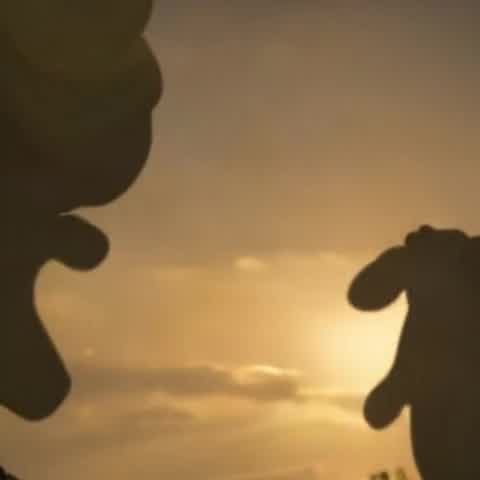} \\

    \end{tabular}%
}

%% file: figures/Appendix_Qual_Compare_Figures/Our_generations_prior_work_compare2/our_gen_figure.tex
\begin{figure}
    \centering
    \captionsetup{type=figure}
    \vspace{-1cm}
    \input{figures/Appendix_Qual_Compare_Figures/Our_generations_prior_work_compare2/figure.tex}
    \captionof{figure}{
    Example \textToVShort generations from \OURS and \imagenvideo  on two prompts (which are shown above each row of frames).
    \imagenvideo generates videos that are faithful to the text, however the videos lack in pixel sharpness and motion smoothness.
    Additionally \imagenvideo's generations lack fine-grained high-quality details such as in the panda's hair (see 4th row) and the water movements (see 2nd row).
    \OURS on the other hand generates high quality videos that are faithful to the text, and with high pixel sharpness and motion smoothness.
    \OURS accurately generates natural looking fine-grained details such as the hair on the panda (see 3rd row) and the water droplets in the waves (see 1st row).
    }
\label{fig:compare_to_prior_3}
\end{figure}
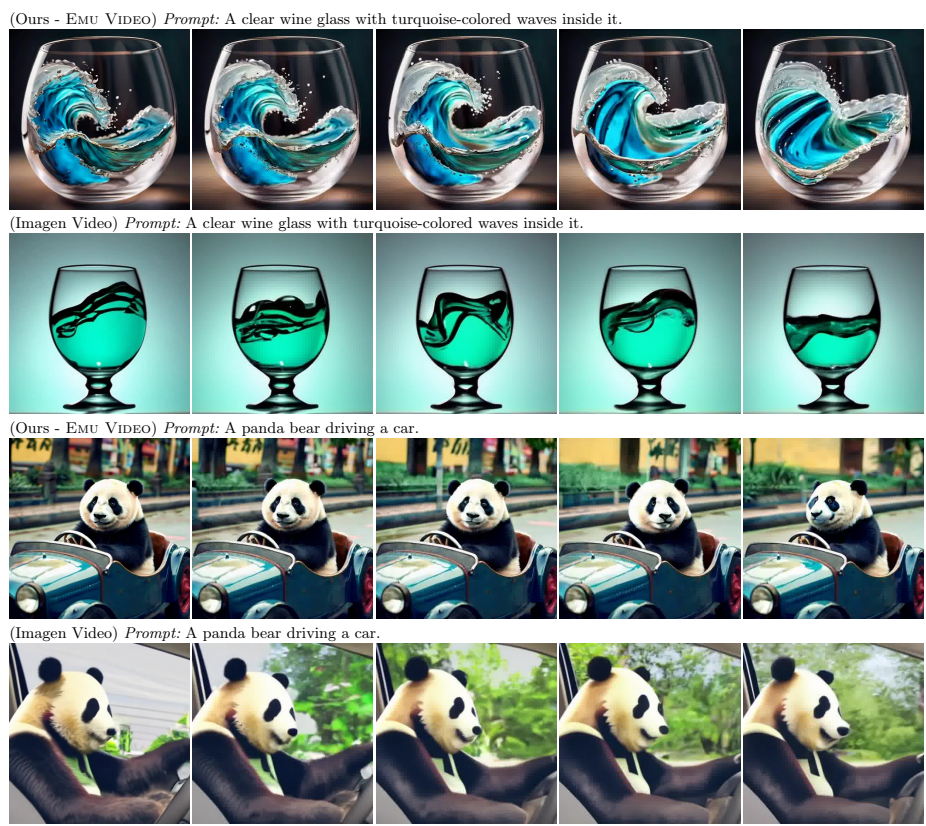%

%% file: figures/Appendix_Qual_Compare_Figures/Our_generations_prior_work_compare2/figure.tex
\setlength{\tabcolsep}{1pt}
\resizebox{\linewidth}{!}{%
    \begin{tabular}{ccccc}
        \multicolumn{5}{l}{ \small {(Ours - \OURS) \it Prompt:} A clear wine glass with turquoise-colored waves inside it.}  \\
        \includegraphics[width=0.3\linewidth]{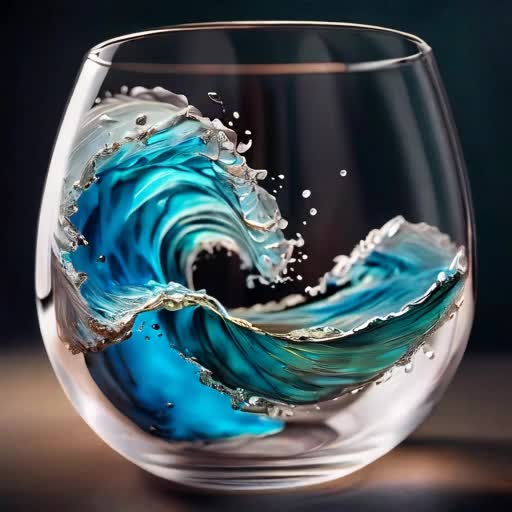} &
        \includegraphics[width=0.3\linewidth]{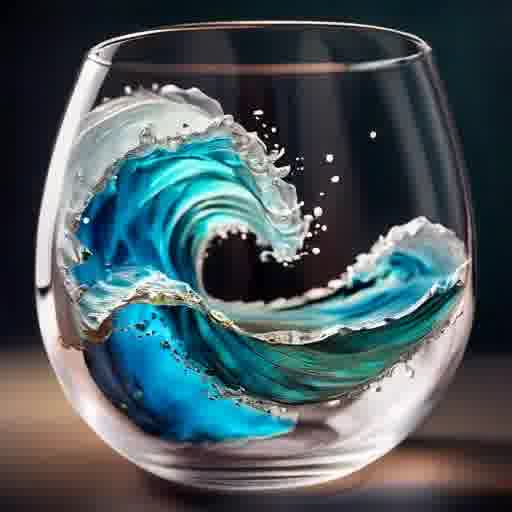} &
        \includegraphics[width=0.3\linewidth]{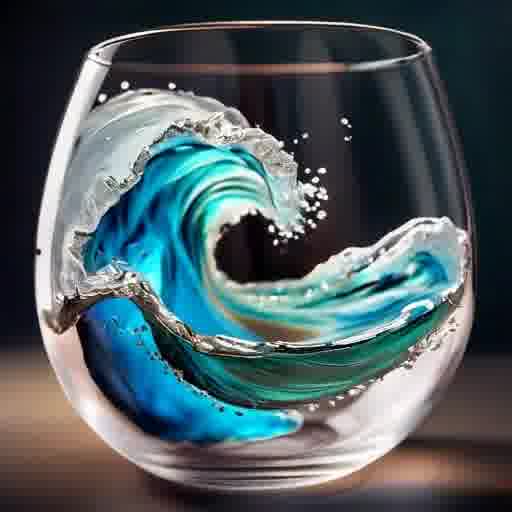} &
        \includegraphics[width=0.3\linewidth]{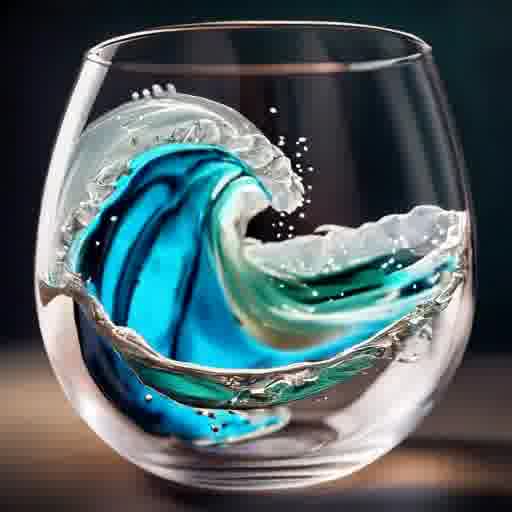} &
        \includegraphics[width=0.3\linewidth]{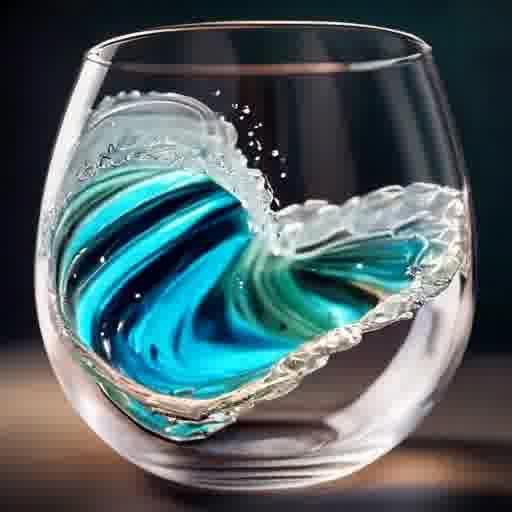} \\
        \multicolumn{5}{l}{ \small {(\imagenvideo) \it Prompt:} A clear wine glass with turquoise-colored waves inside it.}  \\
        \includegraphics[width=0.3\linewidth]{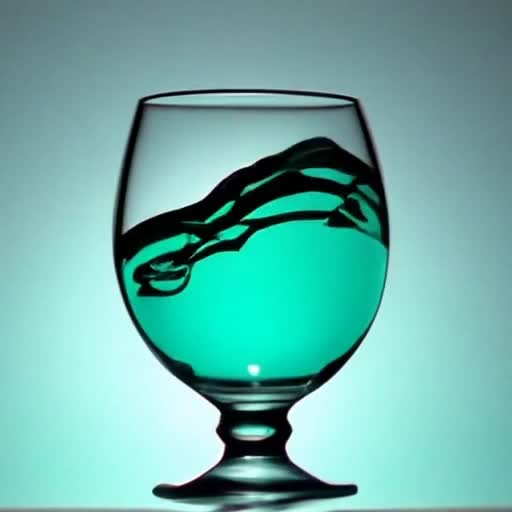} &
        \includegraphics[width=0.3\linewidth]{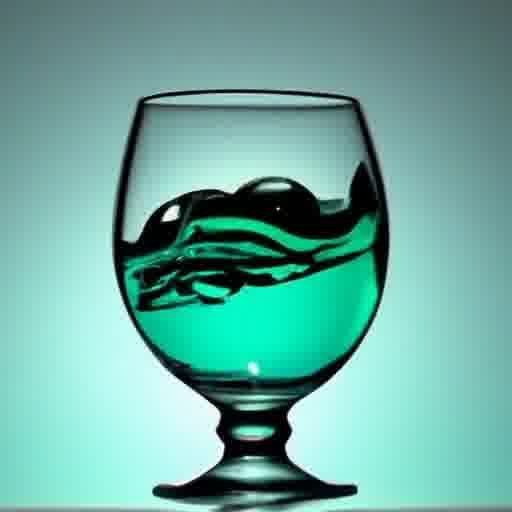} &
        \includegraphics[width=0.3\linewidth]{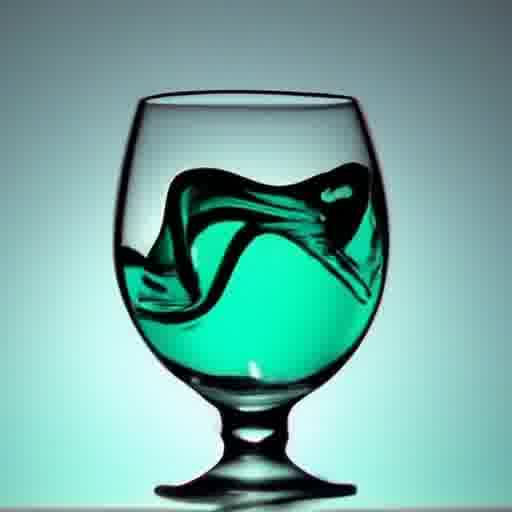} &
        \includegraphics[width=0.3\linewidth]{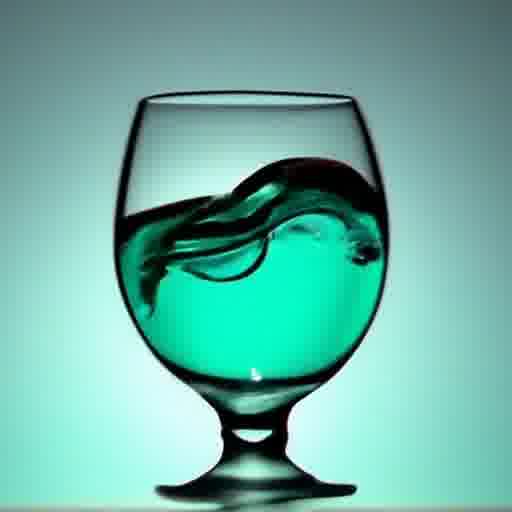} &
        \includegraphics[width=0.3\linewidth]{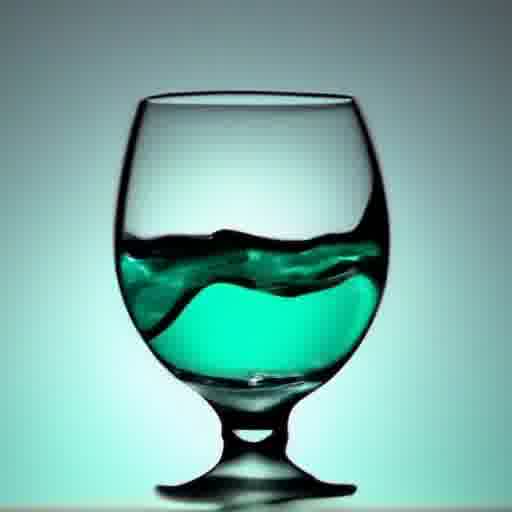} \\

        \multicolumn{5}{l}{ \small {(Ours - \OURS) \it Prompt:} A panda bear driving a car.}  \\
        \includegraphics[width=0.3\linewidth]{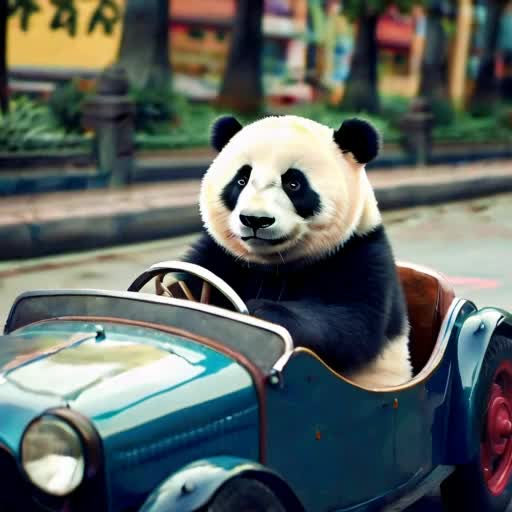} &
        \includegraphics[width=0.3\linewidth]{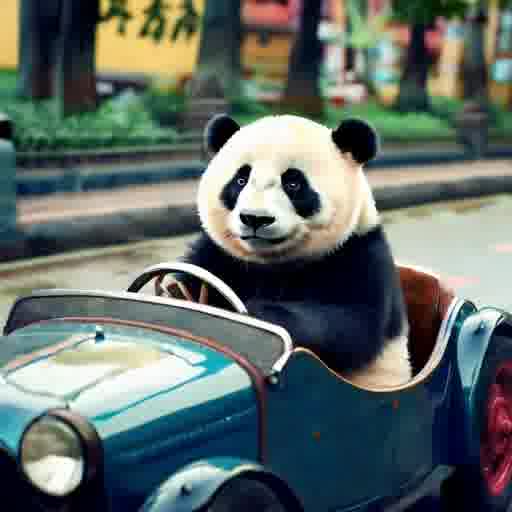} &
        \includegraphics[width=0.3\linewidth]{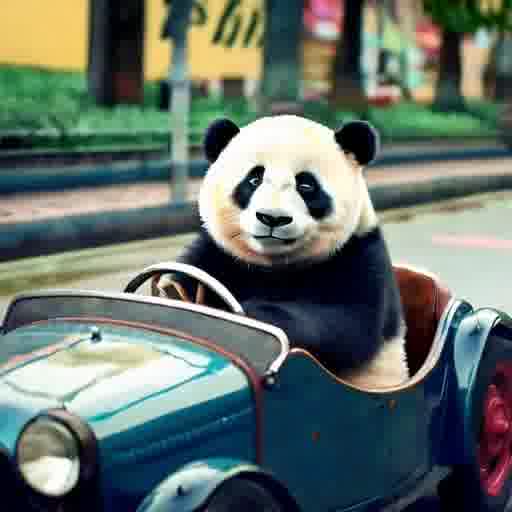} &
        \includegraphics[width=0.3\linewidth]{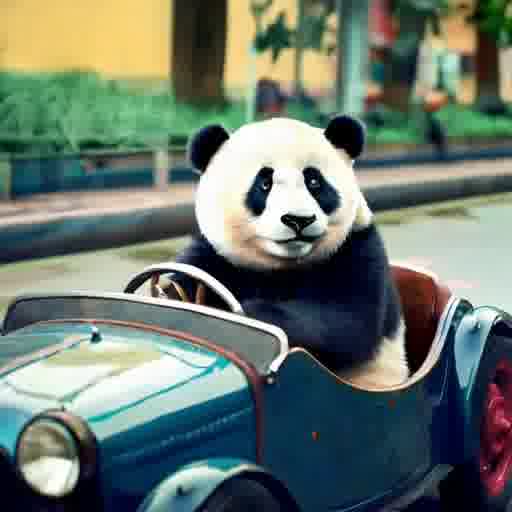} &
        \includegraphics[width=0.3\linewidth]{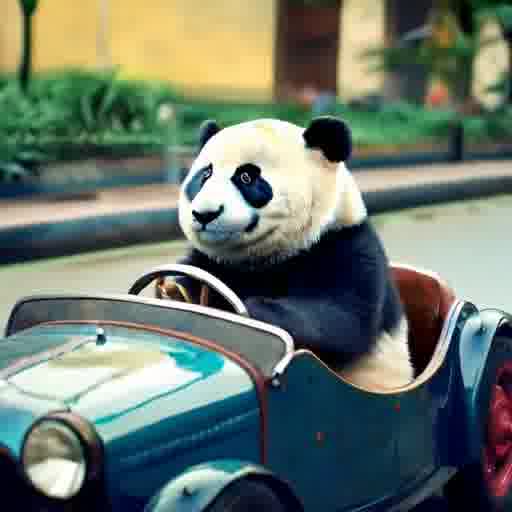} \\
        \multicolumn{5}{l}{ \small {(\imagenvideo) \it Prompt:} A panda bear driving a car.}  \\
        \includegraphics[width=0.3\linewidth]{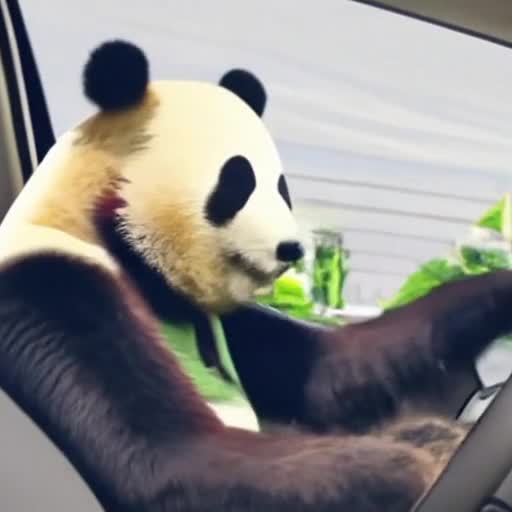} &
        \includegraphics[width=0.3\linewidth]{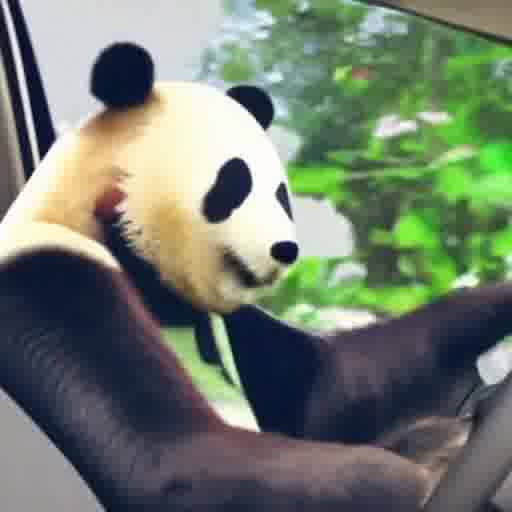} &
        \includegraphics[width=0.3\linewidth]{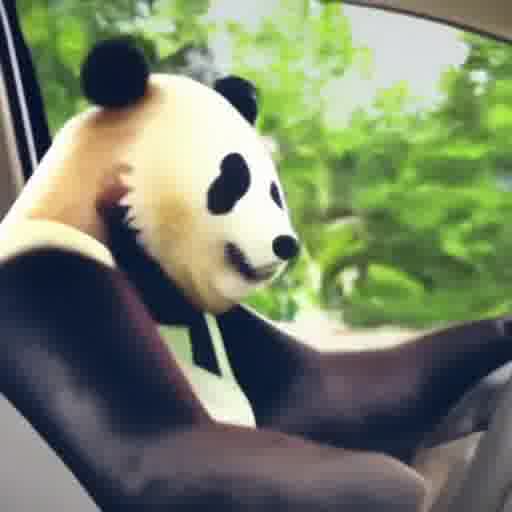} &
        \includegraphics[width=0.3\linewidth]{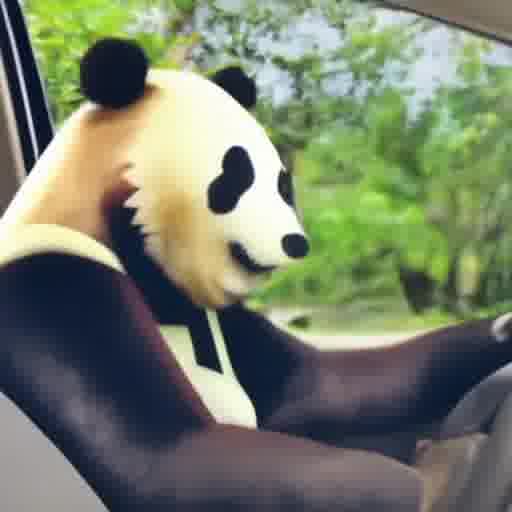} &
        \includegraphics[width=0.3\linewidth]{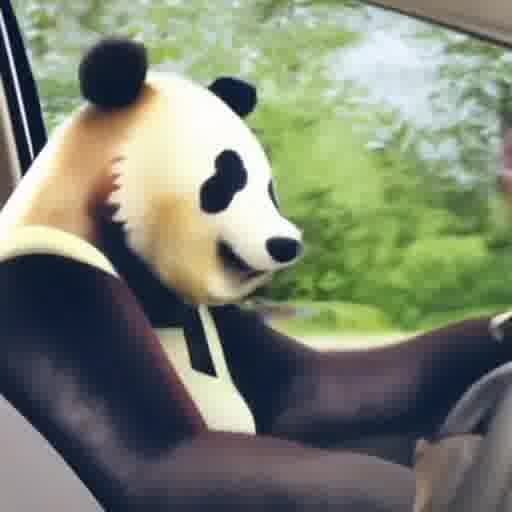} \\

    \end{tabular}%
}

%% file: figures/Appendix_Qual_Compare_Figures/Our_generations_prior_work_compare3/our_gen_figure.tex
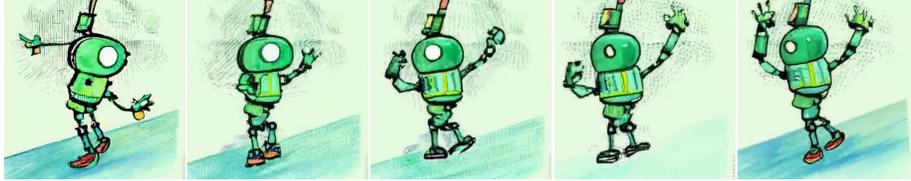
\begin{figure}
    \centering
    \captionsetup{type=figure}
    \vspace{-1cm}
    \input{figures/Appendix_Qual_Compare_Figures/Our_generations_prior_work_compare3/figure.tex}
    \captionof{figure}{
    Example \textToVShort generations from \OURS and \pyoco  on two prompts (which are shown above each row of frames).
    Whereas \pyoco's videos lack motion smoothness or consistency and cannot generate fine-grained details, \OURS instead generates highly realistic videos that are smooth and consistent.
    \OURS can generate high quality videos given fantastical prompts.
    }
\label{fig:compare_to_prior_4}
\end{figure}%

%% file: figures/Appendix_Qual_Compare_Figures/Our_generations_prior_work_compare3/figure.tex
\setlength{\tabcolsep}{1pt}
\resizebox{\linewidth}{!}{%
    \begin{tabular}{ccccc}
        \multicolumn{5}{l}{\footnotesize \multirow{2}{*}{\begin{tabular}[c]{@{}l@{}} {(Ours - \OURS) \it Prompt:} A robot dj is playing the turntable, in heavy raining futuristic tokyo rooftop cyberpunk night, \\ sci-fi, fantasy, intricate, elegant, neon light, highly detailed, concept art, soft light, smooth, sharp focus, illustration.\end{tabular}}} \\ \\
        \includegraphics[width=0.3\linewidth]{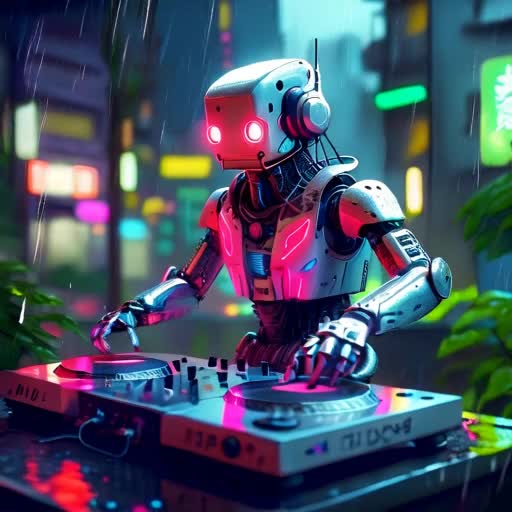} &
        \includegraphics[width=0.3\linewidth]{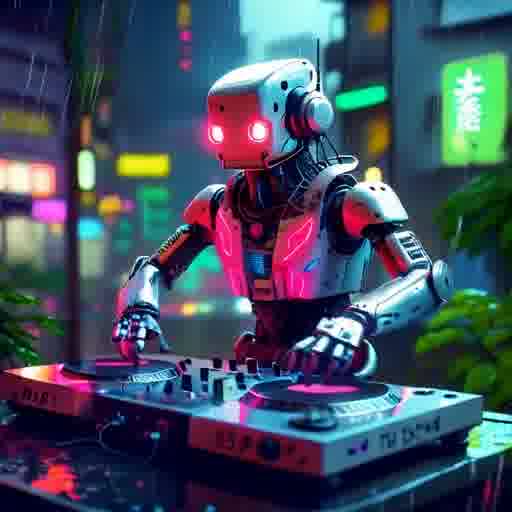} &
        \includegraphics[width=0.3\linewidth]{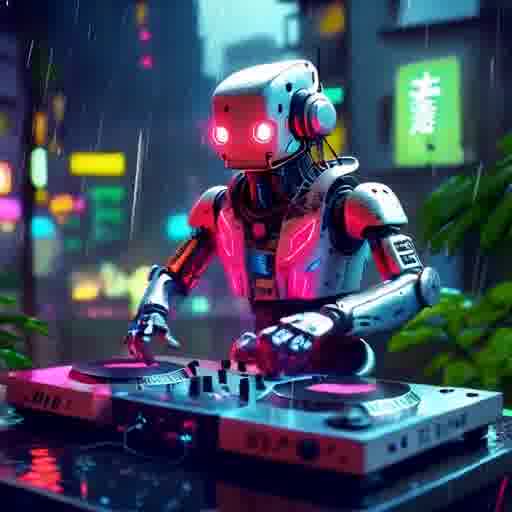} &
        \includegraphics[width=0.3\linewidth]{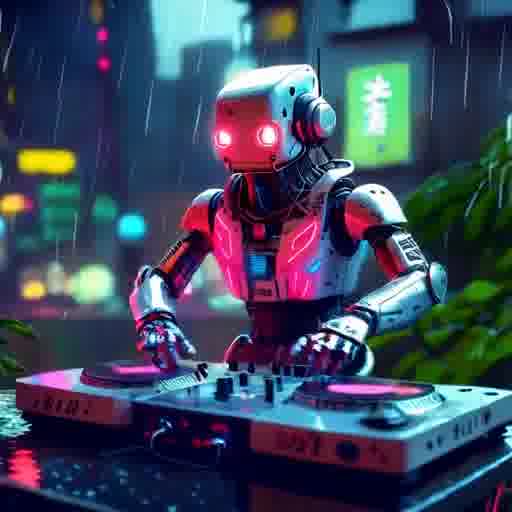} &
        \includegraphics[width=0.3\linewidth]{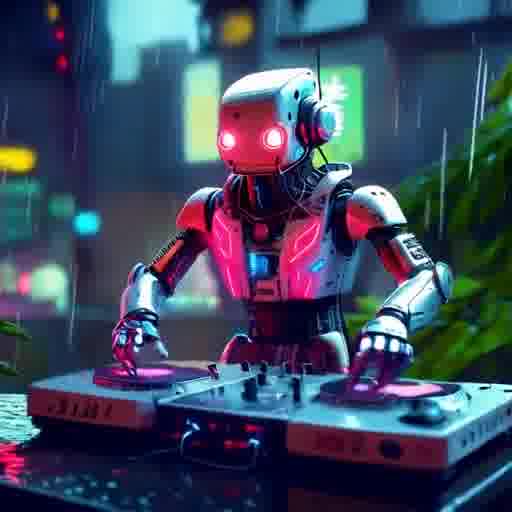} \\
        \multicolumn{5}{l}{\footnotesize \multirow{2}{*}{\begin{tabular}[c]{@{}l@{}} {(\pyoco) \it Prompt:}  A robot dj is playing the turntable, in heavy raining futuristic tokyo rooftop cyberpunk night, \\ sci-fi, fantasy, intricate, elegant, neon light, highly detailed, concept art, soft light, smooth, sharp focus, illustration.\end{tabular}}} \\ \\
        \includegraphics[width=0.3\linewidth]{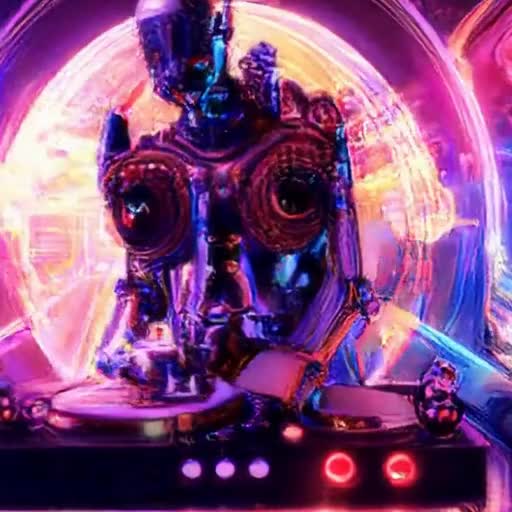} &
        \includegraphics[width=0.3\linewidth]{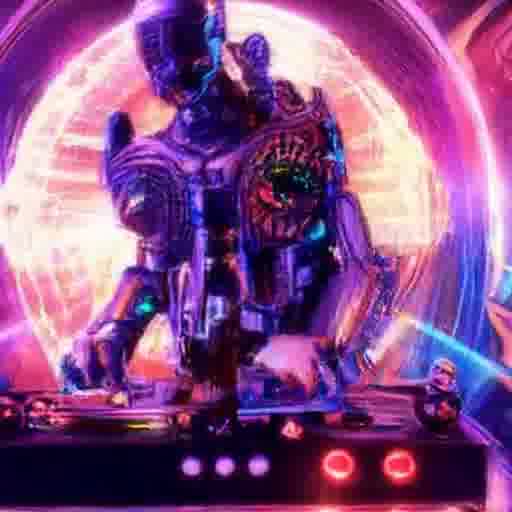} &
        \includegraphics[width=0.3\linewidth]{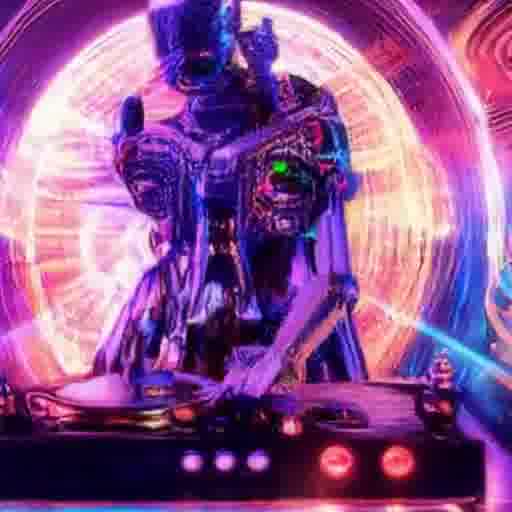} &
        \includegraphics[width=0.3\linewidth]{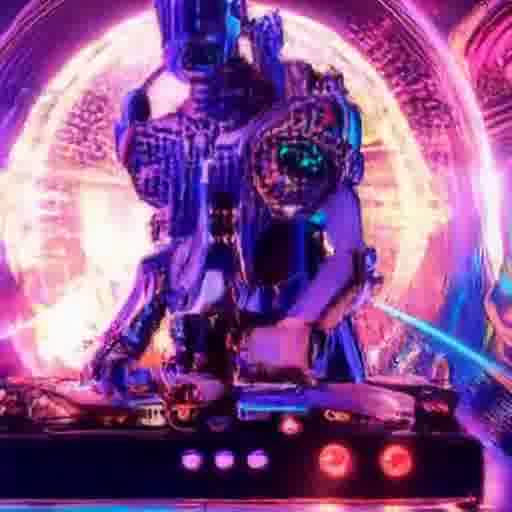} &
        \includegraphics[width=0.3\linewidth]{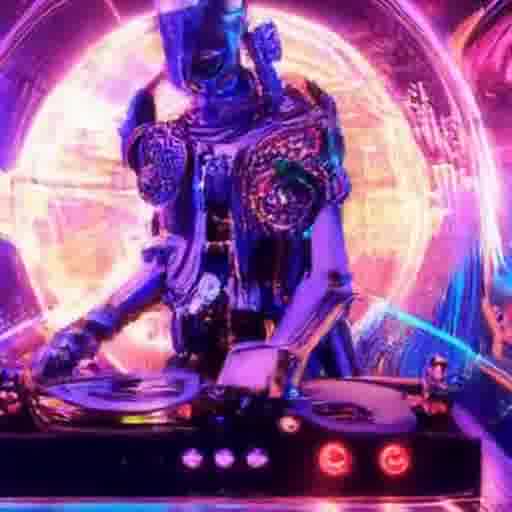} \\

        \multicolumn{5}{l}{\footnotesize \multirow{2}{*}{\begin{tabular}[c]{@{}l@{}} {(Ours - \OURS) \it Prompt:} A cute funny robot dancing, centered, award winning watercolor pen illustration, \\  detailed, isometric illustration, drawing. \end{tabular}}} \\ \\
        \includegraphics[width=0.3\linewidth]{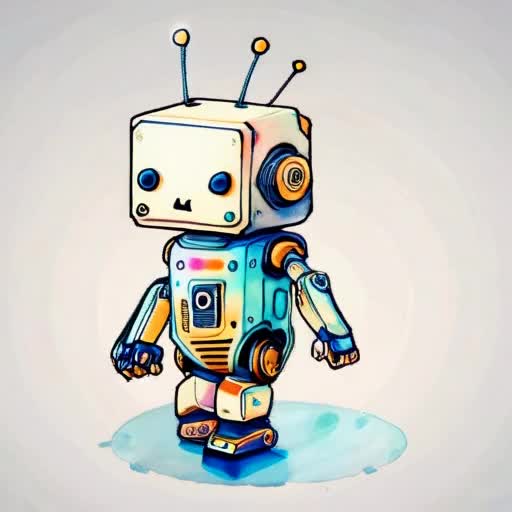} &
        \includegraphics[width=0.3\linewidth]{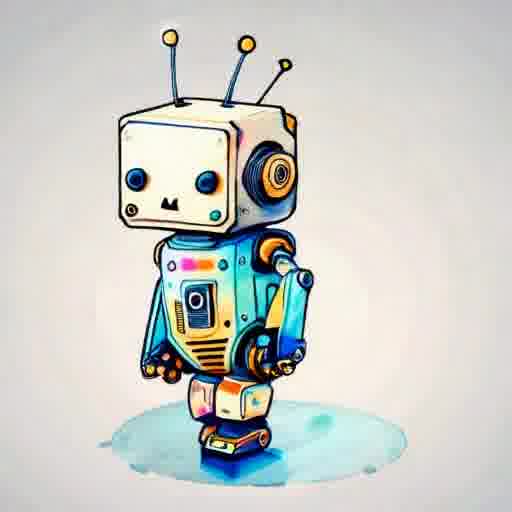} &
        \includegraphics[width=0.3\linewidth]{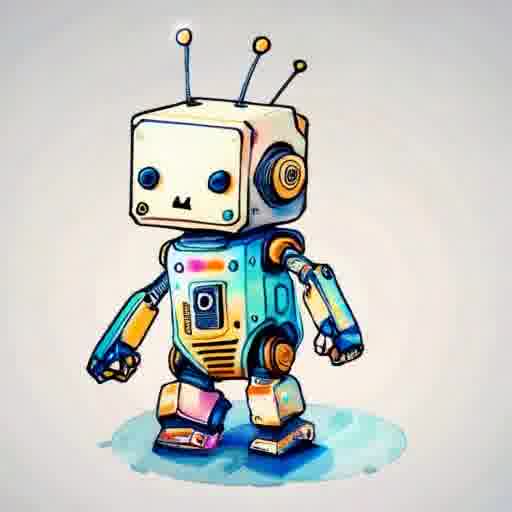} &
        \includegraphics[width=0.3\linewidth]{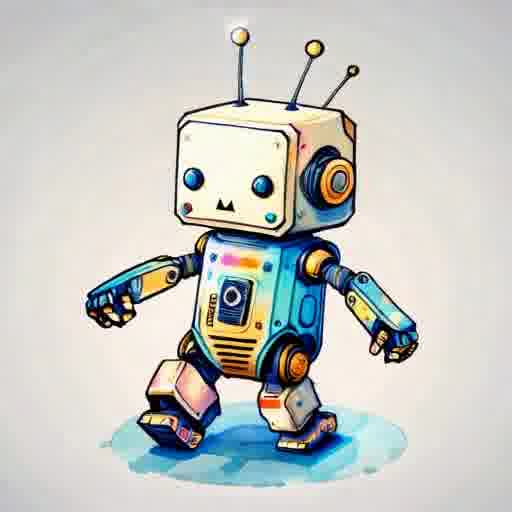} &
        \includegraphics[width=0.3\linewidth]{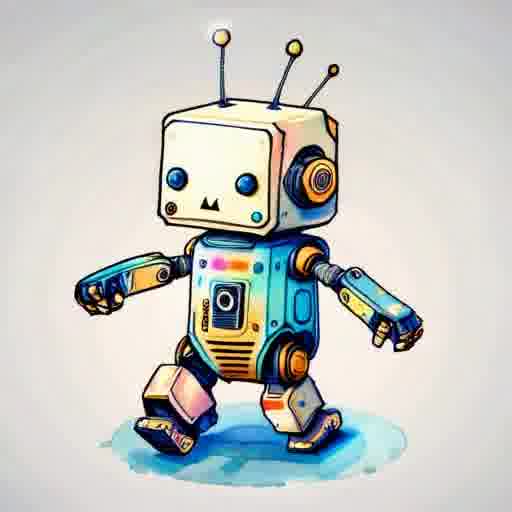} \\
        \multicolumn{5}{l}{\footnotesize \multirow{2}{*}{\begin{tabular}[c]{@{}l@{}} {(\pyoco) \it Prompt:} A cute funny robot dancing, centered, award winning watercolor pen illustration, \\  detailed, isometric illustration, drawing. \end{tabular}}} \\ \\
        \includegraphics[width=0.3\linewidth]{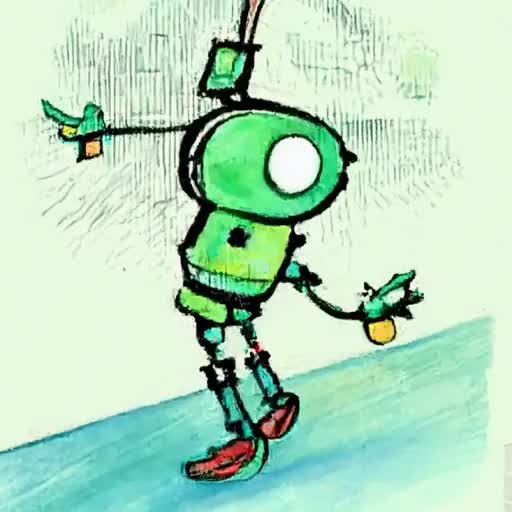} &
        \includegraphics[width=0.3\linewidth]{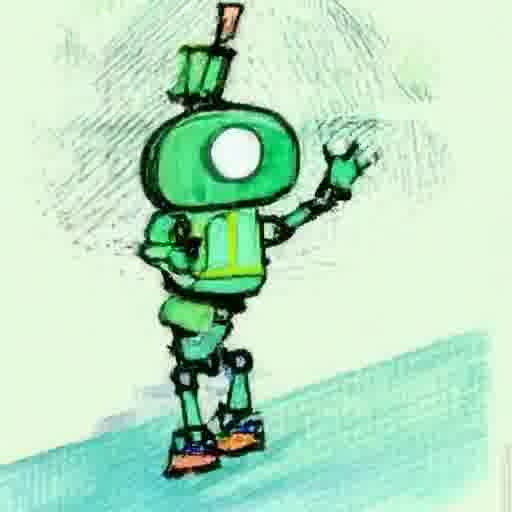} &
        \includegraphics[width=0.3\linewidth]{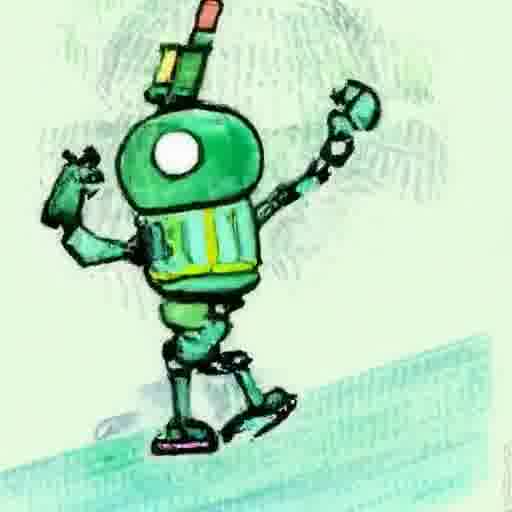} &
        \includegraphics[width=0.3\linewidth]{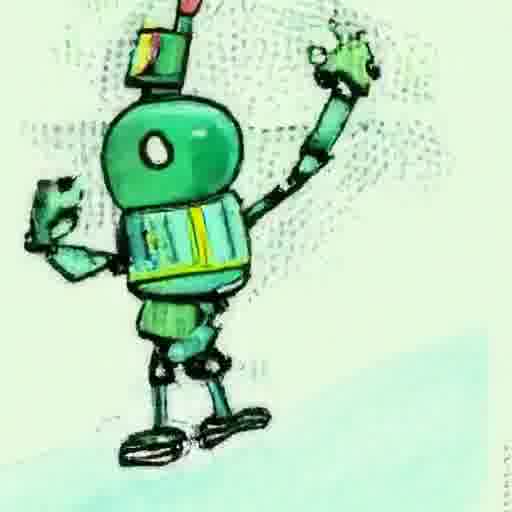} &
        \includegraphics[width=0.3\linewidth]{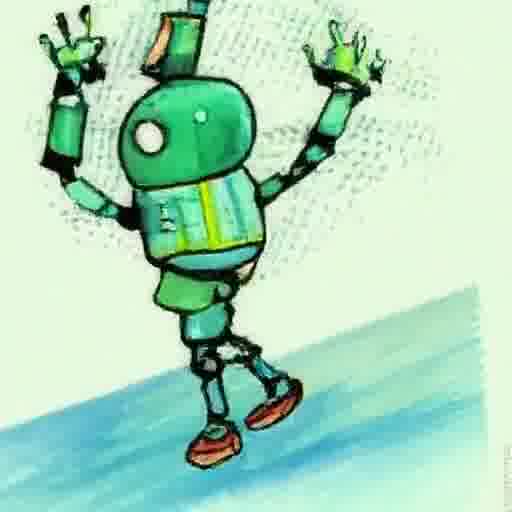} \\

    \end{tabular}%
}

%% file: figures/Appendix_Qual_Compare_Figures/Our_generations_prior_work_compare4/our_gen_figure.tex
\begin{figure}
    \centering
    \captionsetup{type=figure}
    \vspace{-1cm}
    \input{figures/Appendix_Qual_Compare_Figures/Our_generations_prior_work_compare4/figure.tex}
    \captionof{figure}{
    Example \textToVShort generations from \OURS and \mav  on two prompts (which are shown above each row of frames).
    whereas \mav's videos lack pixel sharpness and object consistency, \OURS generates high quality and natural-looking videos.
    \OURS's videos have high motion smoothness and object consistency.
    }
\label{fig:compare_to_prior_5}
\end{figure}
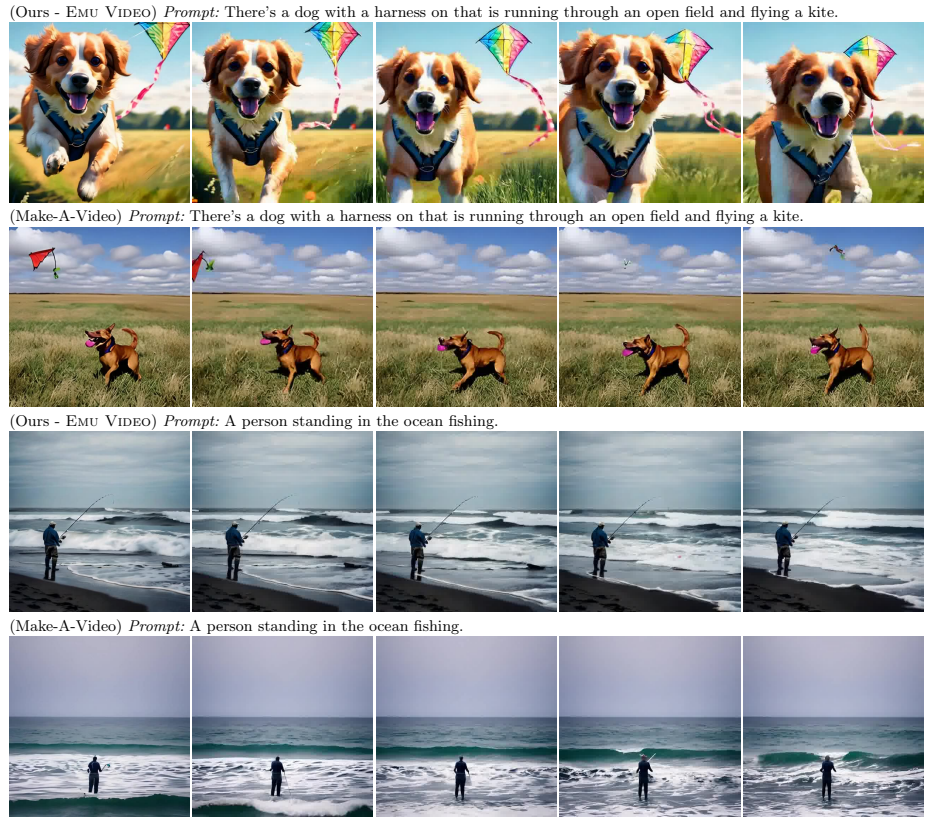%

%% file: figures/Appendix_Qual_Compare_Figures/Our_generations_prior_work_compare4/figure.tex
\setlength{\tabcolsep}{1pt}
\resizebox{\linewidth}{!}{%
    \begin{tabular}{ccccc}
        \multicolumn{5}{l}{ \footnotesize {(Ours - \OURS) \it Prompt:} There's a dog with a harness on that is running through an open field and flying a kite.}  \\

        \includegraphics[width=0.3\linewidth]{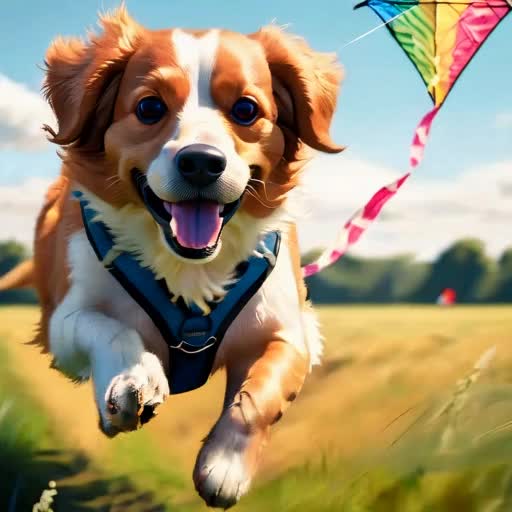} &
        \includegraphics[width=0.3\linewidth]{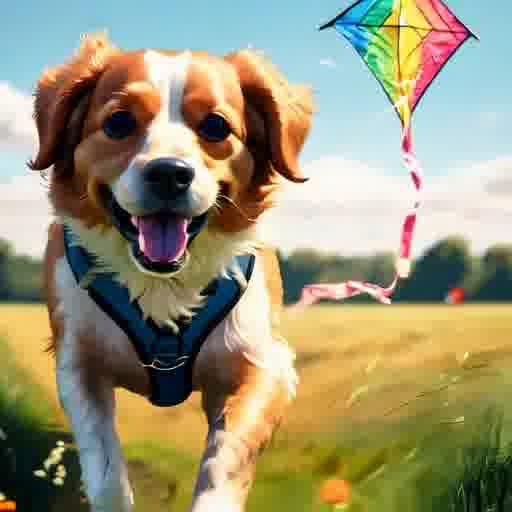} &
        \includegraphics[width=0.3\linewidth]{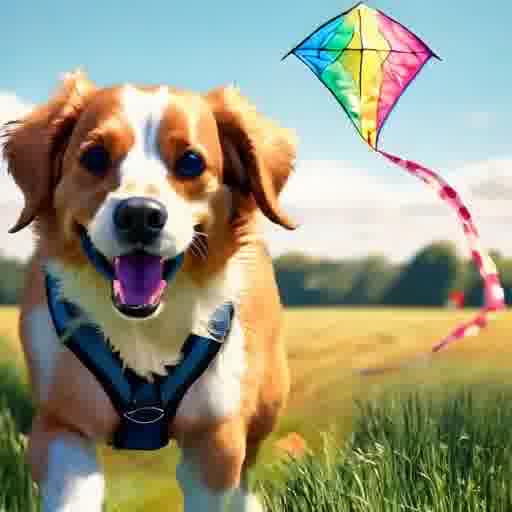} &
        \includegraphics[width=0.3\linewidth]{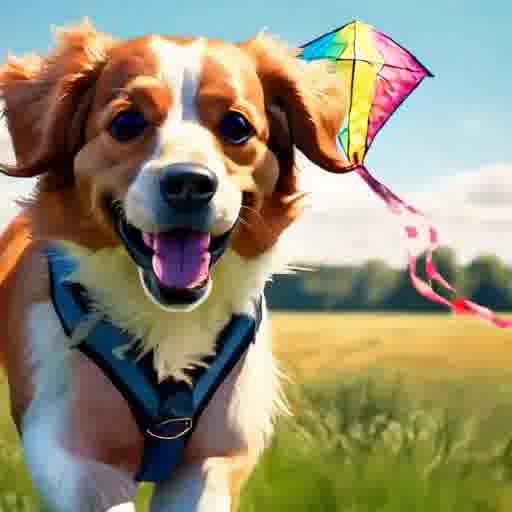} &
        \includegraphics[width=0.3\linewidth]{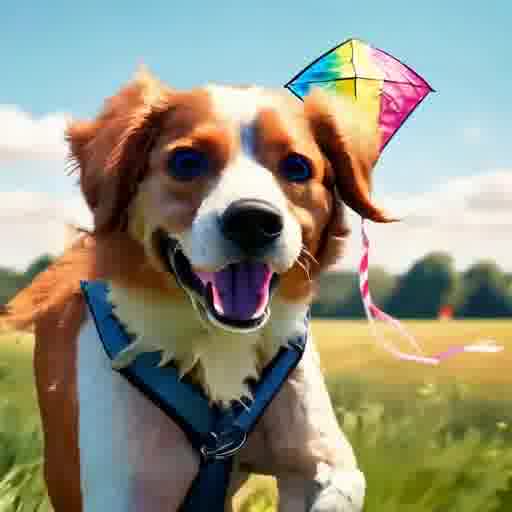} \\
        \multicolumn{5}{l}{ \footnotesize {(\mav) \it Prompt:} There's a dog with a harness on that is running through an open field and flying a kite.}  \\
        \includegraphics[width=0.3\linewidth]{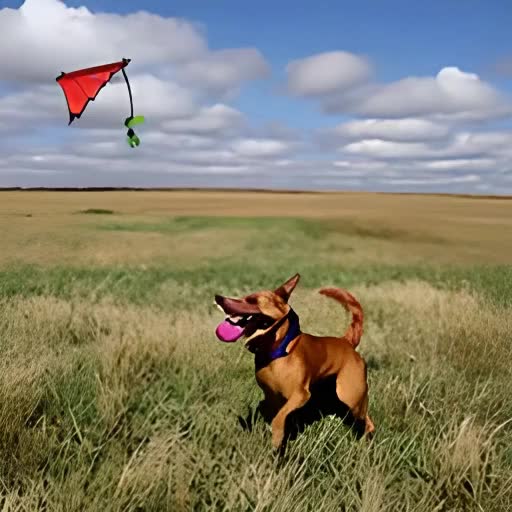} &
        \includegraphics[width=0.3\linewidth]{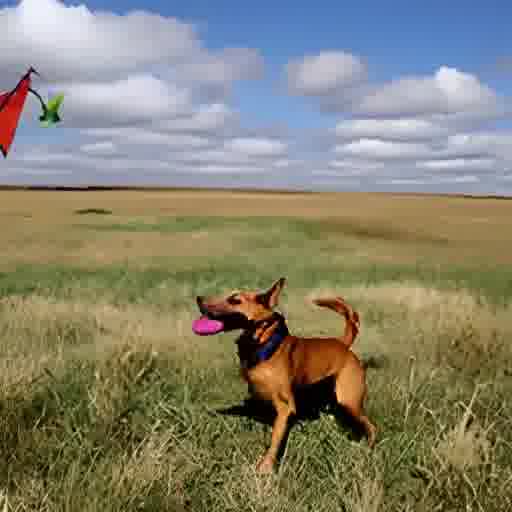} &
        \includegraphics[width=0.3\linewidth]{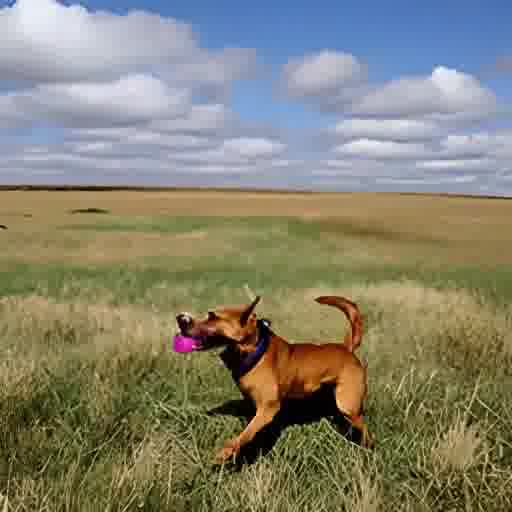} &
        \includegraphics[width=0.3\linewidth]{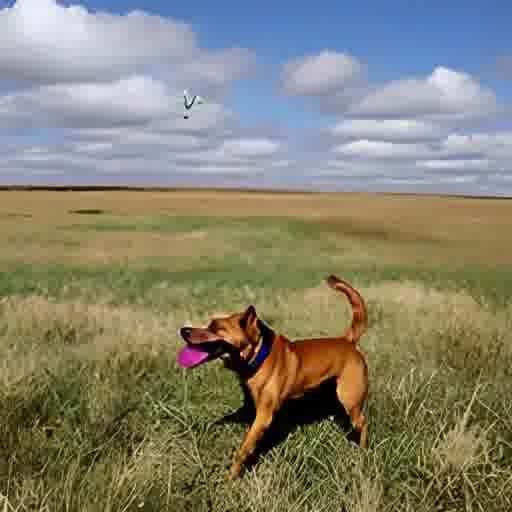} &
        \includegraphics[width=0.3\linewidth]{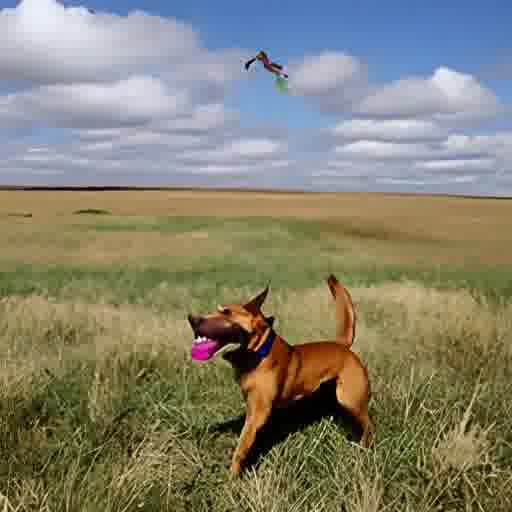} \\

        \multicolumn{5}{l}{ \footnotesize {(Ours - \OURS) \it Prompt:} A person standing in the ocean fishing.}  \\
        \includegraphics[width=0.3\linewidth]{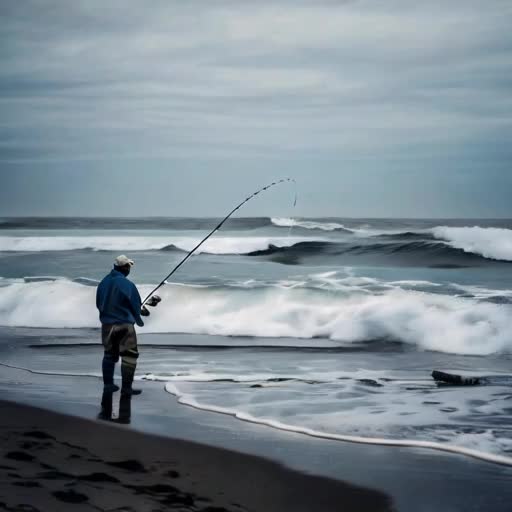} &
        \includegraphics[width=0.3\linewidth]{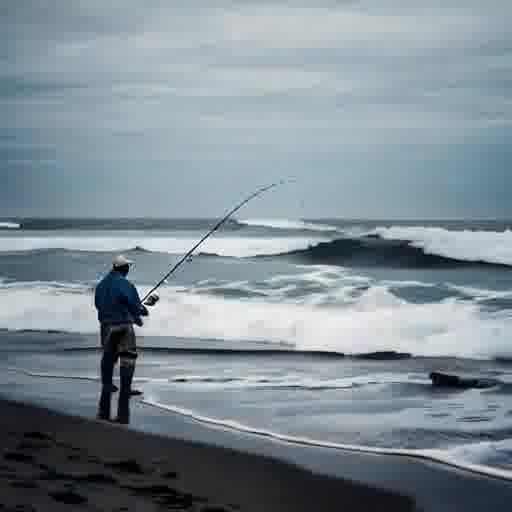} &
        \includegraphics[width=0.3\linewidth]{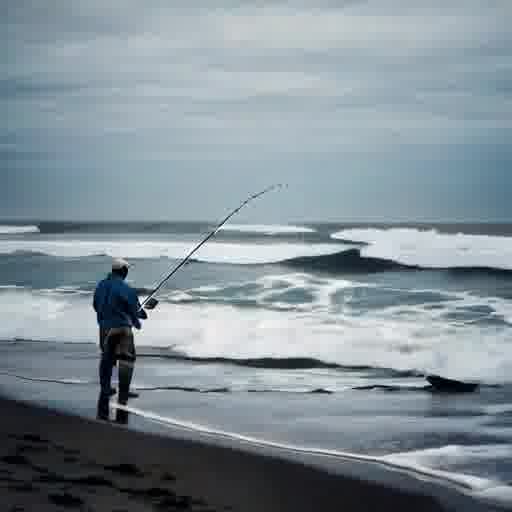} &
        \includegraphics[width=0.3\linewidth]{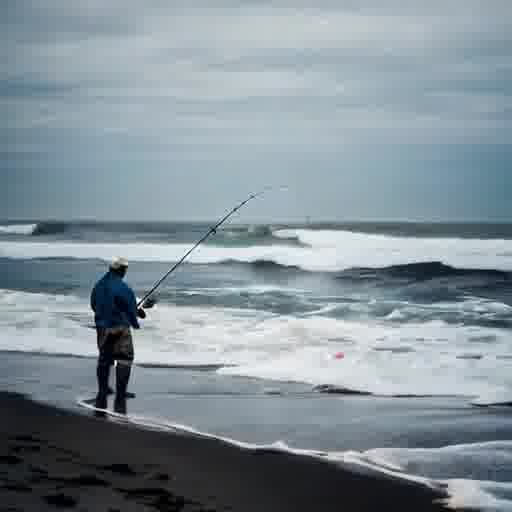} &
        \includegraphics[width=0.3\linewidth]{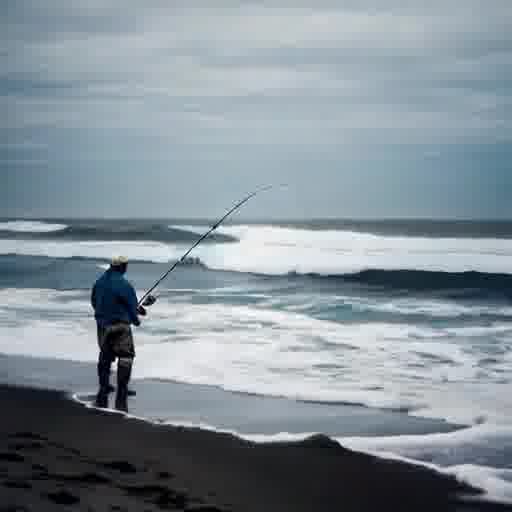} \\
        \multicolumn{5}{l}{ \footnotesize {(\mav) \it Prompt:} A person standing in the ocean fishing.}  \\
        \includegraphics[width=0.3\linewidth]{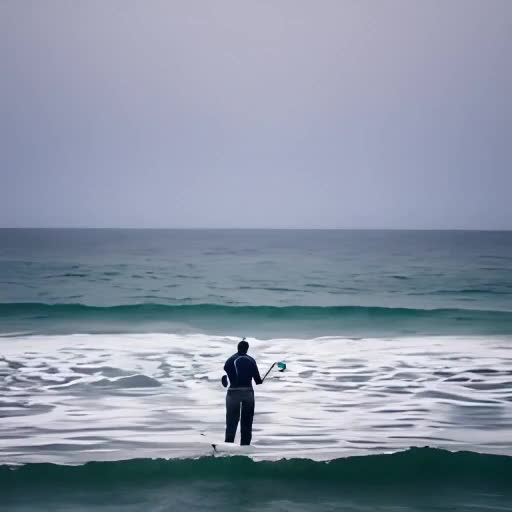} &
        \includegraphics[width=0.3\linewidth]{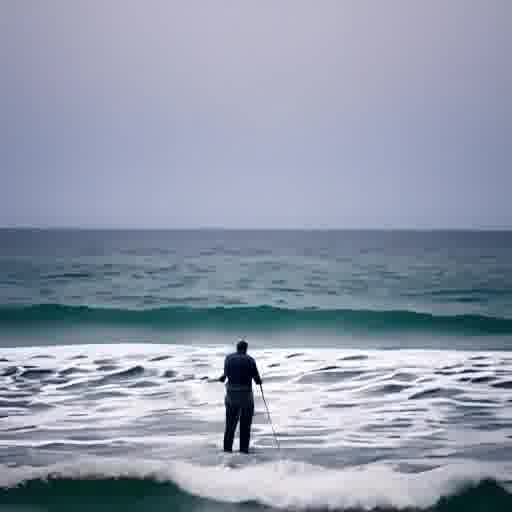} &
        \includegraphics[width=0.3\linewidth]{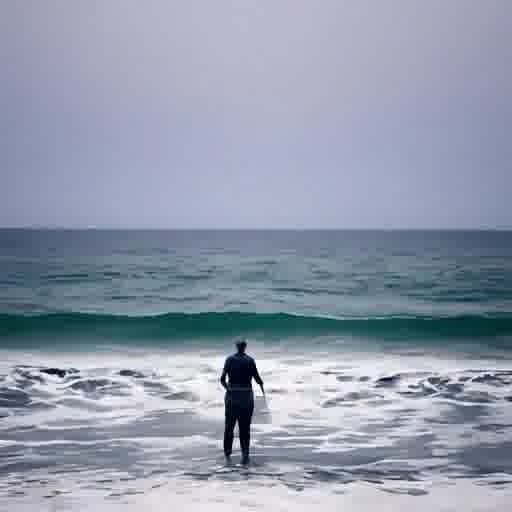} &
        \includegraphics[width=0.3\linewidth]{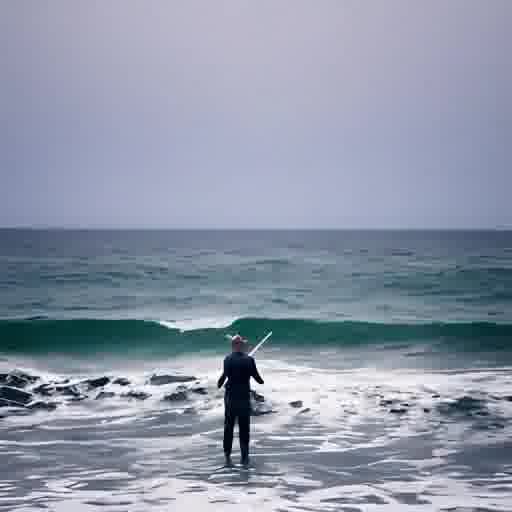} &
        \includegraphics[width=0.3\linewidth]{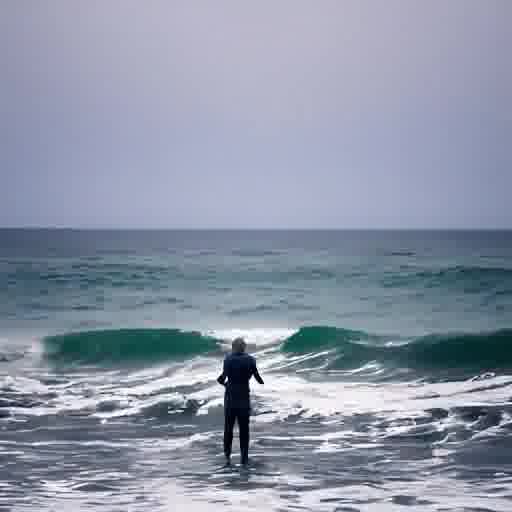} \\

    \end{tabular}%
}

%% file: figures/Appendix_Qual_Compare_Figures/Our_generations_prior_work_compare5/our_gen_figure.tex
\begin{figure}
    \centering
    \captionsetup{type=figure}
    \vspace{-1cm}
    \input{figures/Appendix_Qual_Compare_Figures/Our_generations_prior_work_compare5/figure.tex}
    \captionof{figure}{
    Example \textToVShort generations from \OURS and \reusediffuse  on two prompts (which are shown above each row of frames).
    whereas \reusediffuse's videos lack in visual quality both in terms of pixel sharpness, and temporal consistency, \OURS instead generates visually compelling and natural-looking videos which accurately follow the prompt.
    }
\label{fig:compare_to_prior_6}
\end{figure}
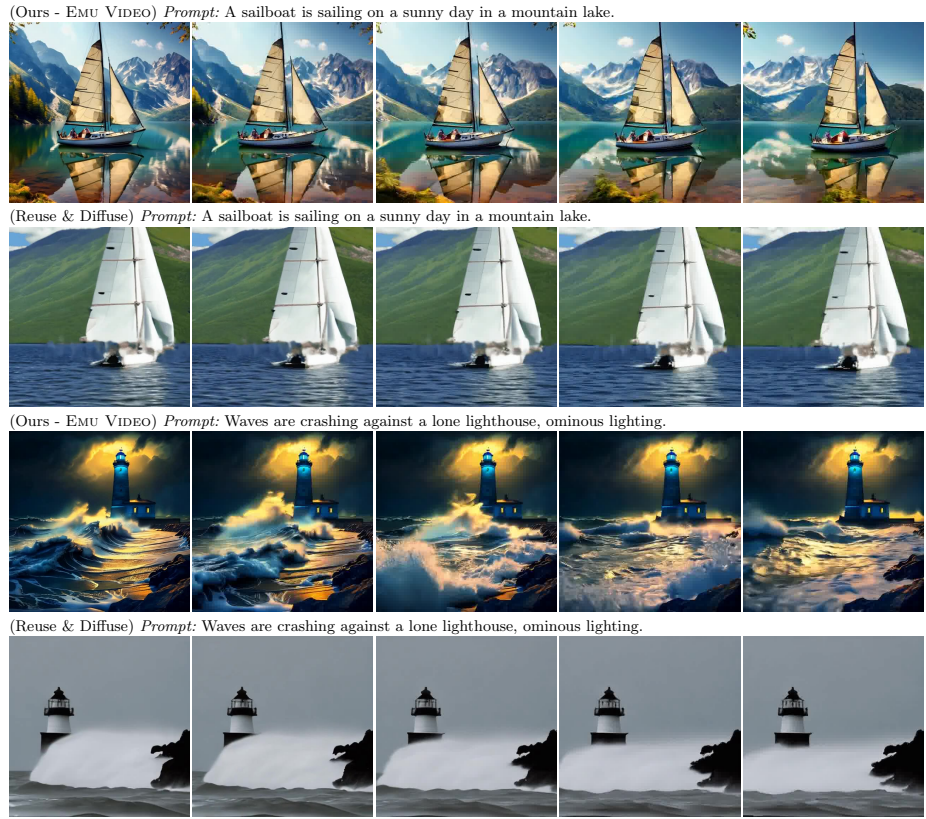%

%% file: figures/Appendix_Qual_Compare_Figures/Our_generations_prior_work_compare5/figure.tex
\setlength{\tabcolsep}{1pt}
\resizebox{\linewidth}{!}{%
    \begin{tabular}{ccccc}
        \multicolumn{5}{l}{ \small {(Ours - \OURS) \it Prompt:} A sailboat is sailing on a sunny day in a mountain lake.}  \\
        \includegraphics[width=0.3\linewidth]{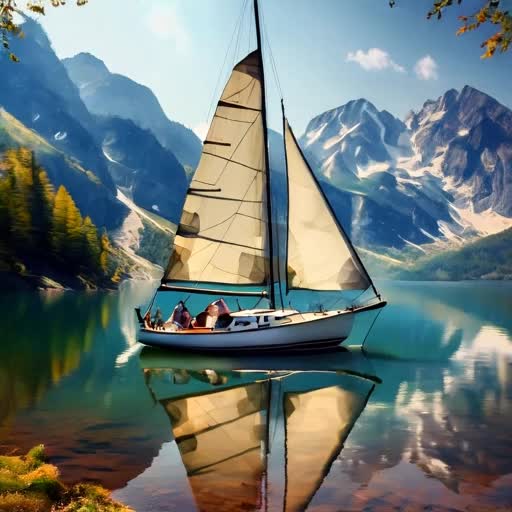} &
        \includegraphics[width=0.3\linewidth]{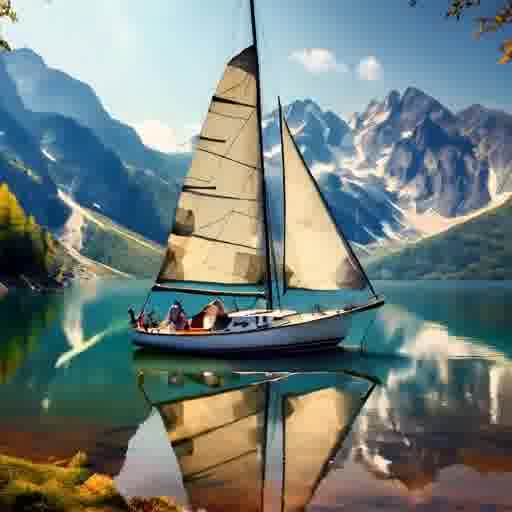} &
        \includegraphics[width=0.3\linewidth]{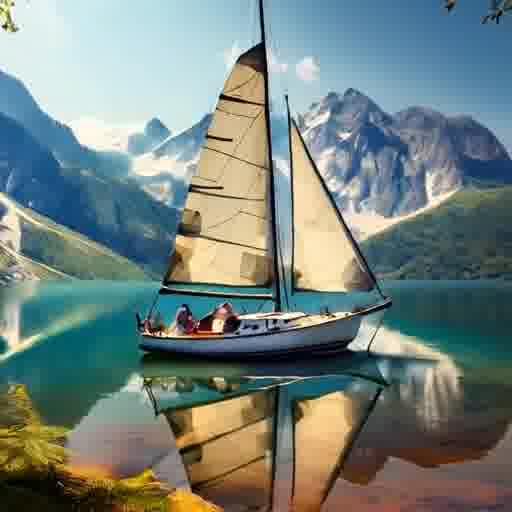} &
        \includegraphics[width=0.3\linewidth]{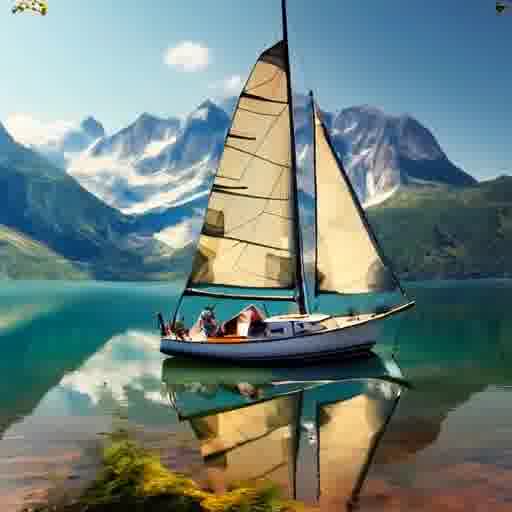} &
        \includegraphics[width=0.3\linewidth]{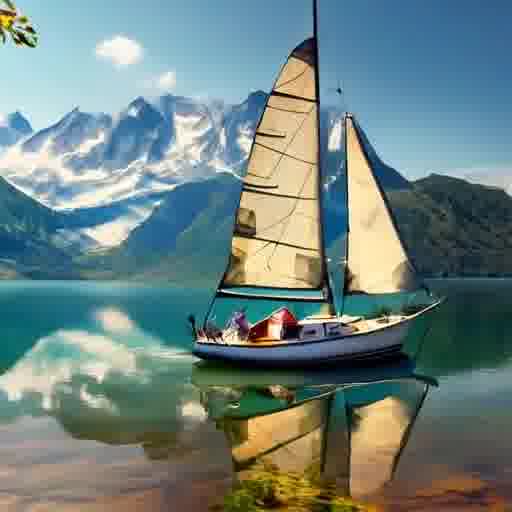} \\
        \multicolumn{5}{l}{ \small {(\reusediffuse) \it Prompt:} A sailboat is sailing on a sunny day in a mountain lake.}  \\
        \includegraphics[width=0.3\linewidth]{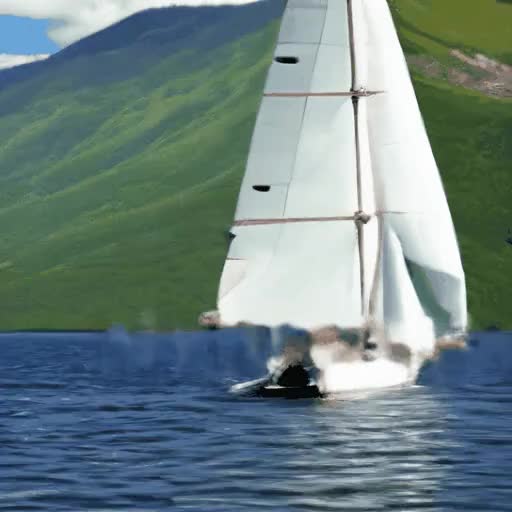} &
        \includegraphics[width=0.3\linewidth]{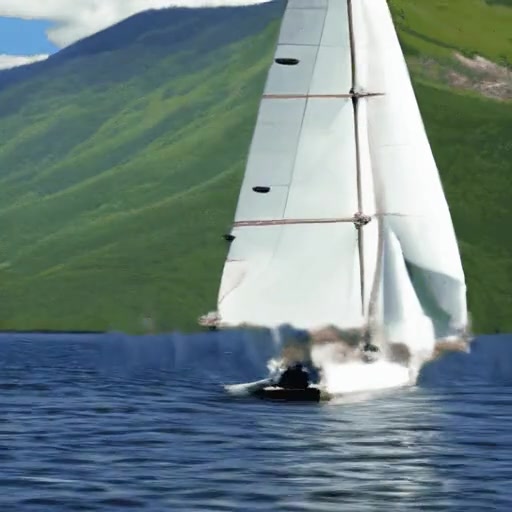} &
        \includegraphics[width=0.3\linewidth]{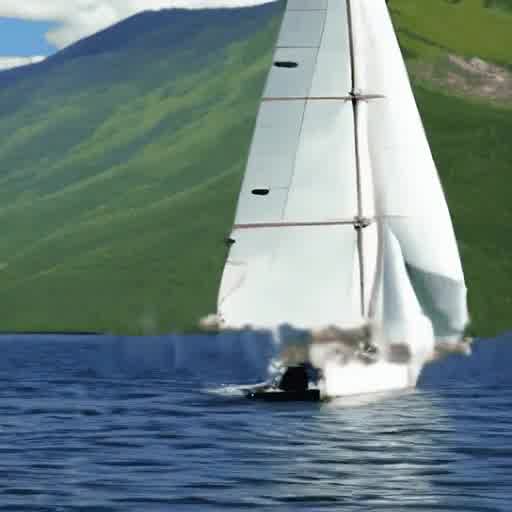} &
        \includegraphics[width=0.3\linewidth]{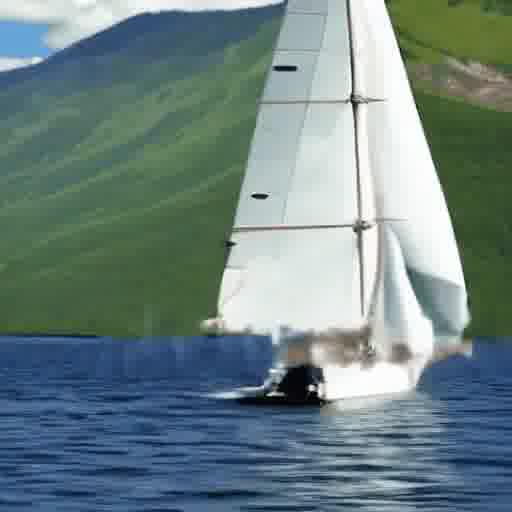} &
        \includegraphics[width=0.3\linewidth]{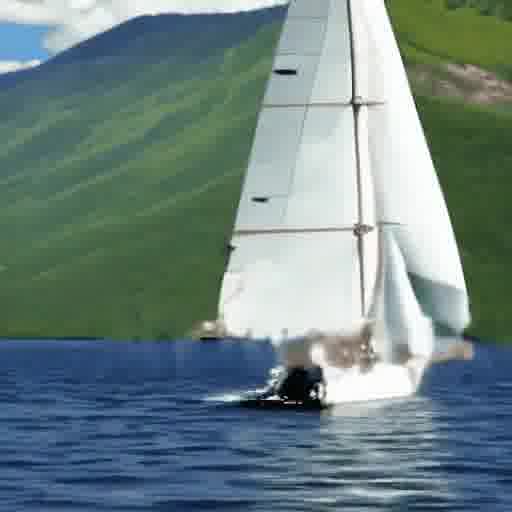} \\

        \multicolumn{5}{l}{ \small {(Ours - \OURS) \it Prompt:} Waves are crashing against a lone lighthouse, ominous lighting.}  \\
        \includegraphics[width=0.3\linewidth]{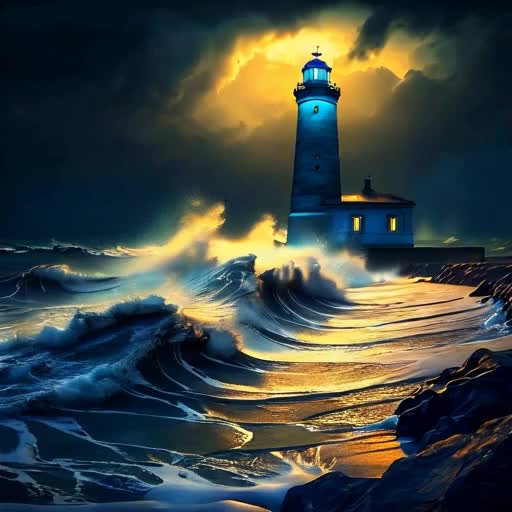} &
        \includegraphics[width=0.3\linewidth]{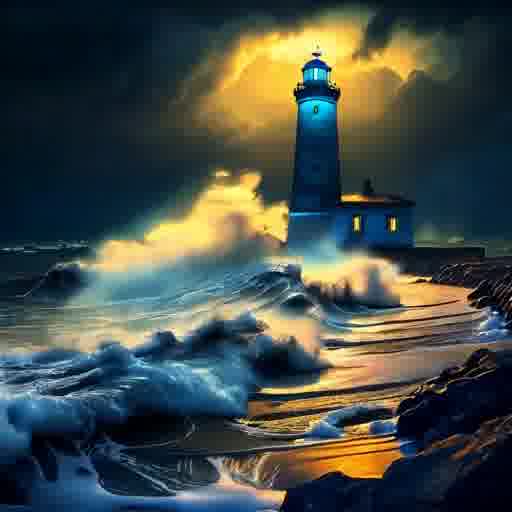} &
        \includegraphics[width=0.3\linewidth]{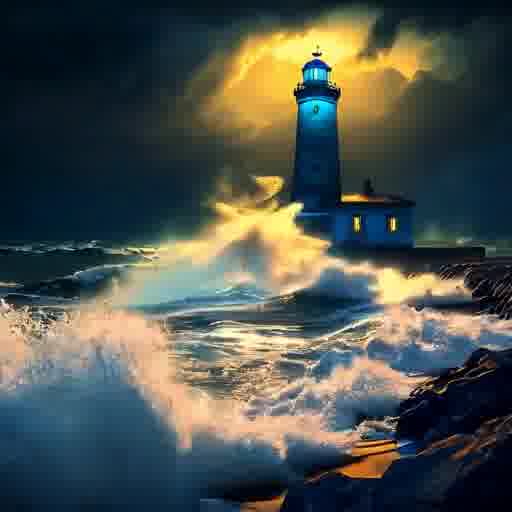} &
        \includegraphics[width=0.3\linewidth]{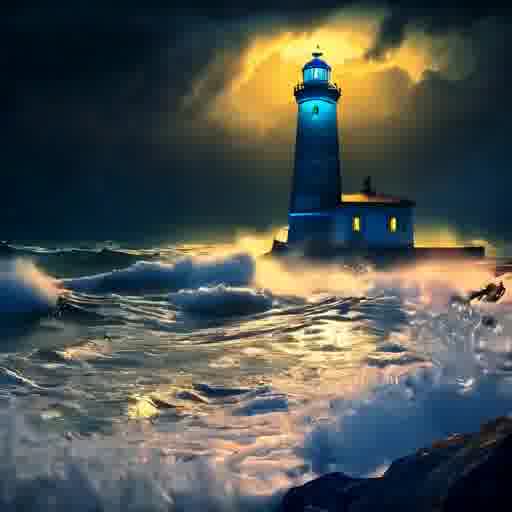} &
        \includegraphics[width=0.3\linewidth]{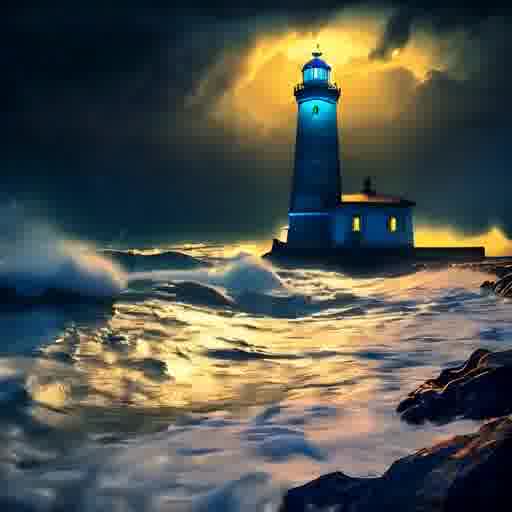} \\
        \multicolumn{5}{l}{ \small {(\reusediffuse) \it Prompt:} Waves are crashing against a lone lighthouse, ominous lighting.}  \\
        \includegraphics[width=0.3\linewidth]{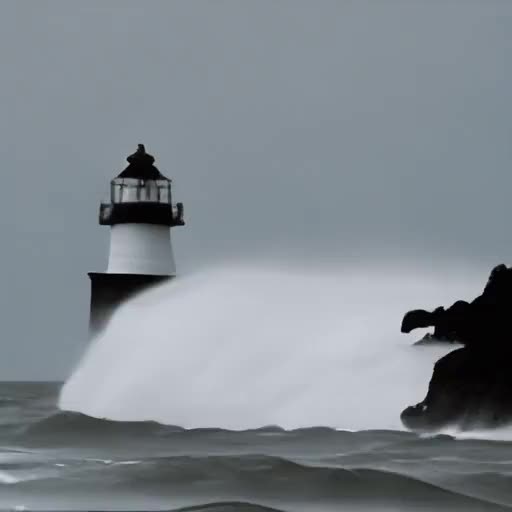} &
        \includegraphics[width=0.3\linewidth]{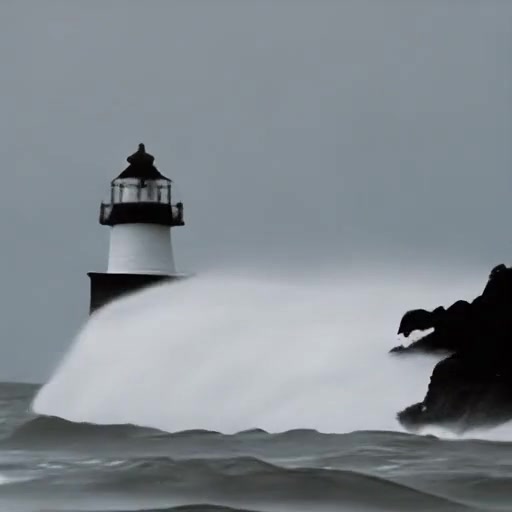} &
        \includegraphics[width=0.3\linewidth]{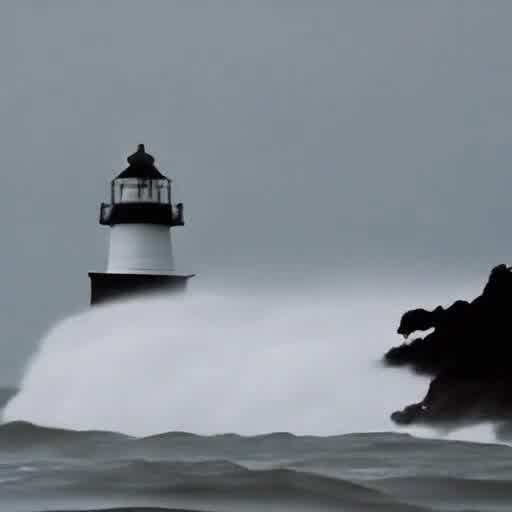} &
        \includegraphics[width=0.3\linewidth]{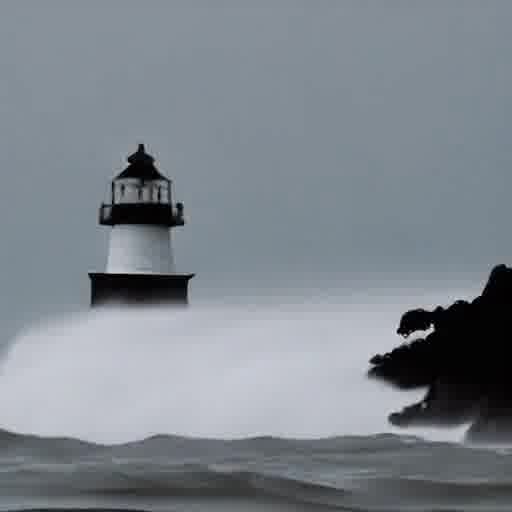} &
        \includegraphics[width=0.3\linewidth]{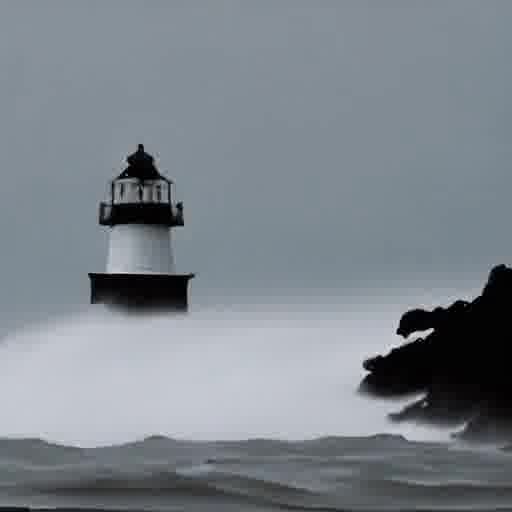} \\

    \end{tabular}%
}

%% file: figures/ucf.tex
\begin{figure*}[!t]
    \centering
    \input{figures/ucf/figure.tex}
    \caption{
    \textbf{Zero-Shot \textToV generation on \ucf.} 
    The classes for these videos from top to bottom are: walking with a dog, biking, handstand pushups, skiing. 
    Our generations are of higher quality and more coherent than those from \mav. 
    }
    \label{fig:ucf}
\end{figure*}
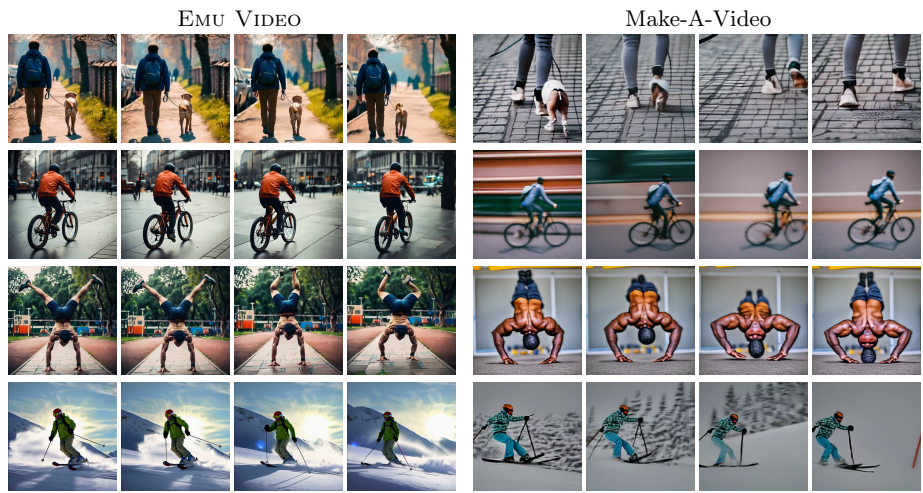

%% file: figures/ucf/figure.tex
\setlength{\tabcolsep}{1pt}
\resizebox{\linewidth}{!}{%
    \begin{tabular}{cccc@{\hskip 0.1in}cccc}
        \multicolumn{4}{c}{\OURS} & \multicolumn{4}{c}{\mav} \\
        \includegraphics[width=0.124\linewidth]{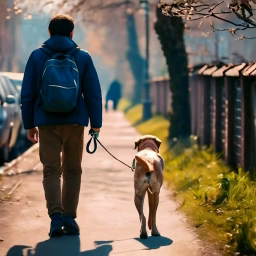} &
        \includegraphics[width=0.124\linewidth]{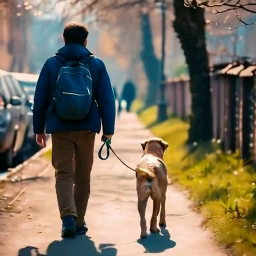} &
        \includegraphics[width=0.124\linewidth]{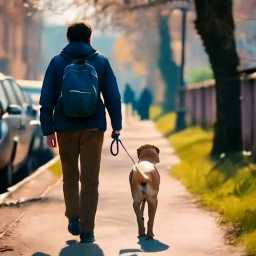} &
        \includegraphics[width=0.124\linewidth]{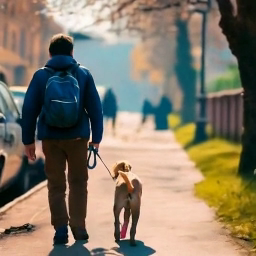} &
        \includegraphics[width=0.124\linewidth]{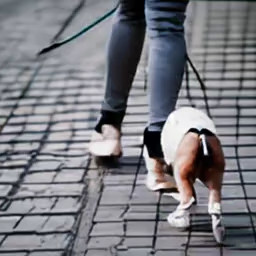} &
        \includegraphics[width=0.124\linewidth]{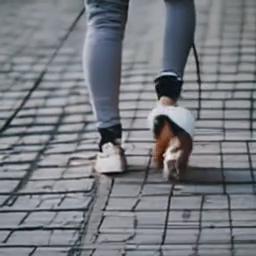} &
        \includegraphics[width=0.124\linewidth]{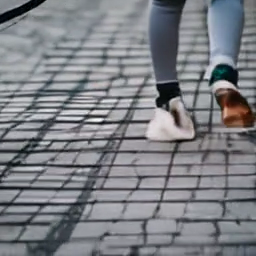} &
        \includegraphics[width=0.124\linewidth]{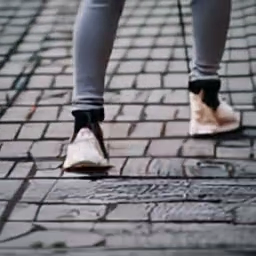} \\
        \includegraphics[width=0.124\linewidth]{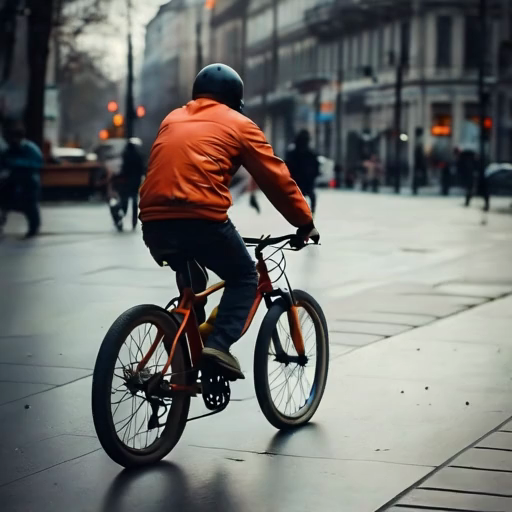} &
        \includegraphics[width=0.124\linewidth]{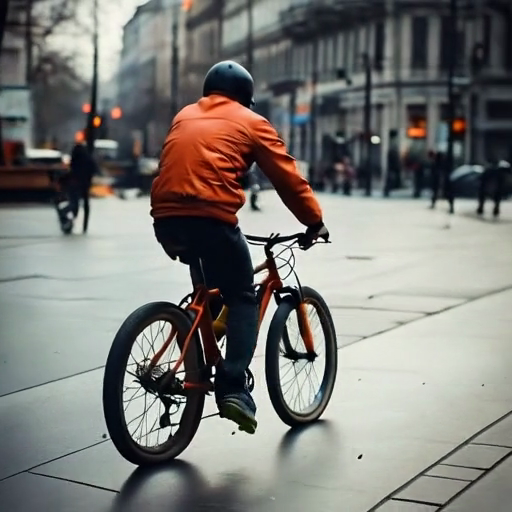} &
        \includegraphics[width=0.124\linewidth]{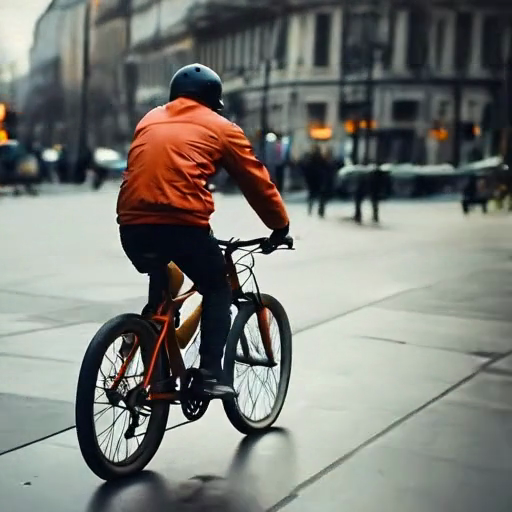} &
        \includegraphics[width=0.124\linewidth]{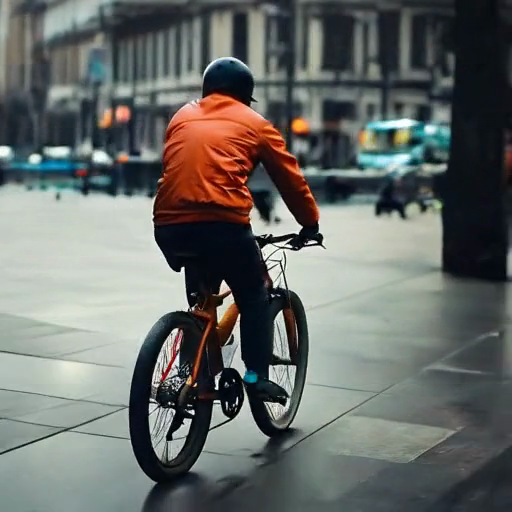} &
        \includegraphics[width=0.124\linewidth]{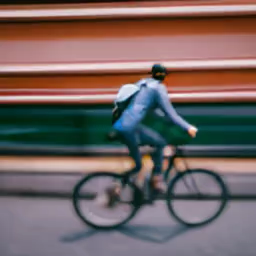} &
        \includegraphics[width=0.124\linewidth]{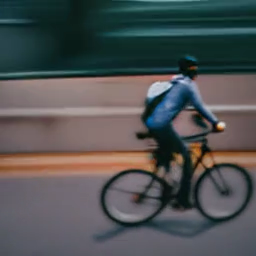} &
        \includegraphics[width=0.124\linewidth]{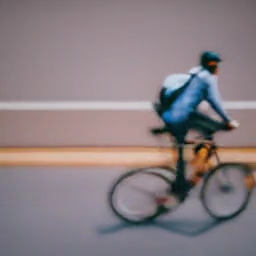} &
        \includegraphics[width=0.124\linewidth]{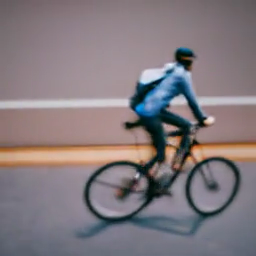} \\
        \includegraphics[width=0.124\linewidth]{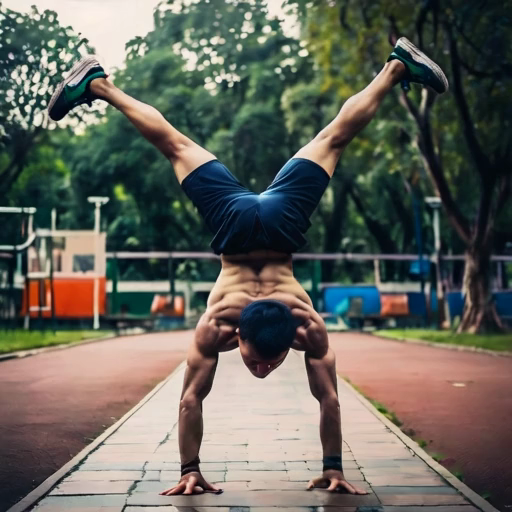} &
        \includegraphics[width=0.124\linewidth]{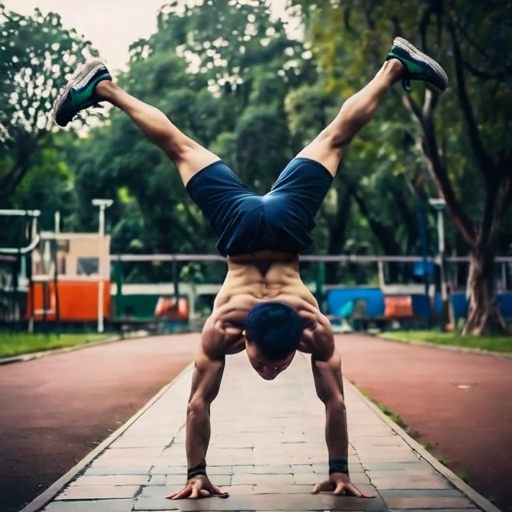} &
        \includegraphics[width=0.124\linewidth]{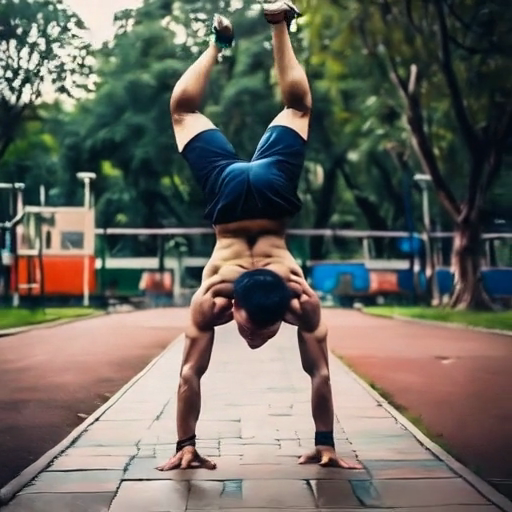} &
        \includegraphics[width=0.124\linewidth]{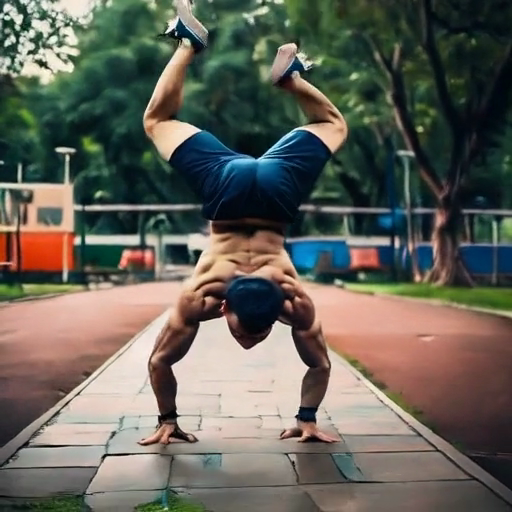} &
        \includegraphics[width=0.124\linewidth]{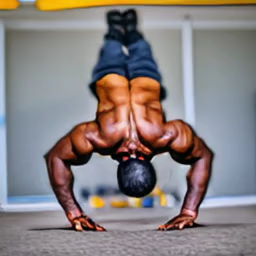} &
        \includegraphics[width=0.124\linewidth]{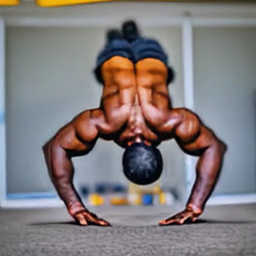} &
        \includegraphics[width=0.124\linewidth]{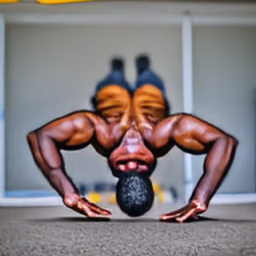} &
        \includegraphics[width=0.124\linewidth]{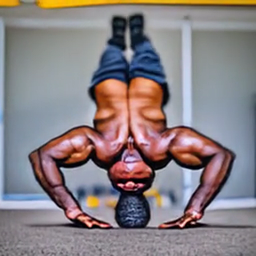} \\
        \includegraphics[width=0.124\linewidth]{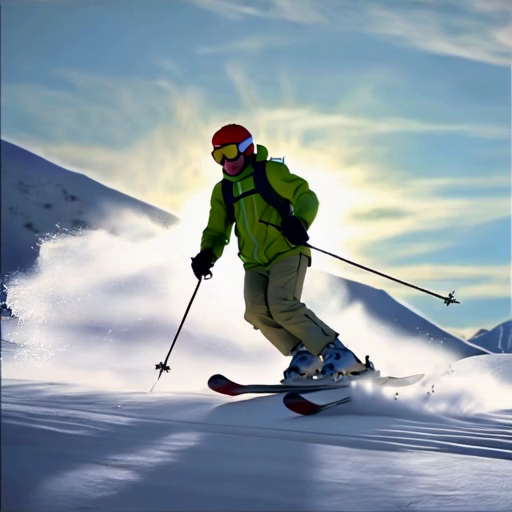} &
        \includegraphics[width=0.124\linewidth]{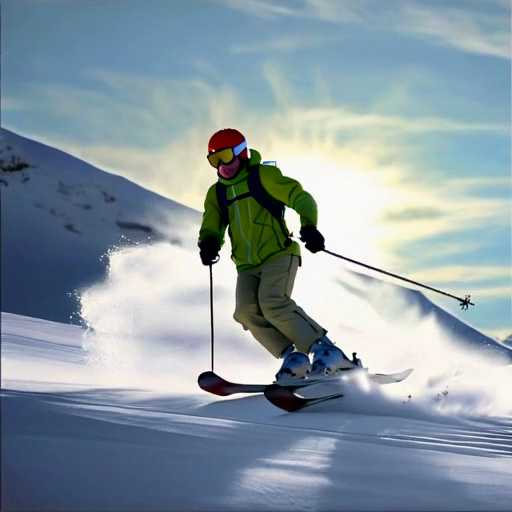} &
        \includegraphics[width=0.124\linewidth]{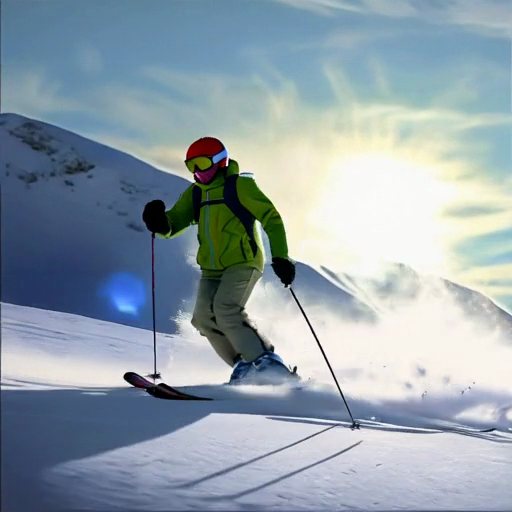} &
        \includegraphics[width=0.124\linewidth]{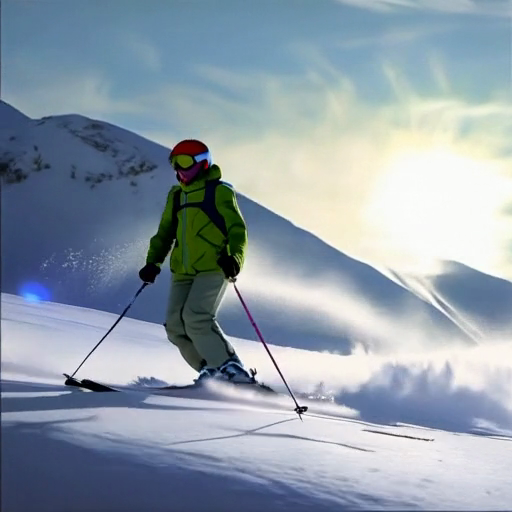} &
        \includegraphics[width=0.124\linewidth]{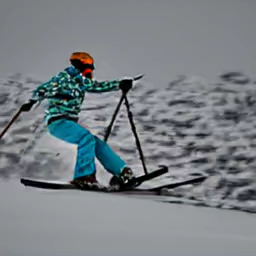} &
        \includegraphics[width=0.124\linewidth]{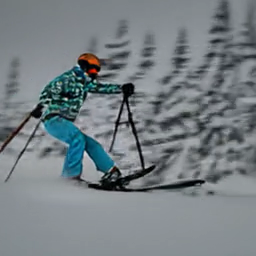} &
        \includegraphics[width=0.124\linewidth]{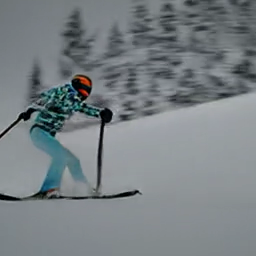} &
        \includegraphics[width=0.124\linewidth]{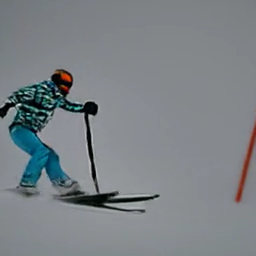} \\
    \end{tabular}%
}